\documentclass{article}

 \usepackage[main, final]{neurips_2026}
\usepackage{amsmath}
\usepackage{amssymb}

\usepackage{algorithm,algpseudocode}
\usepackage{hyperref}       % hyperlinks
\usepackage{url}            % URL typesetting
\usepackage{booktabs}       % professional-quality tables
\usepackage{amsfonts}       % blackboard math symbols
\usepackage{nicefrac}       % compact fractions
\usepackage{microtype}      % microtypography
\usepackage{xspace}
\usepackage{mathabx}
\usepackage{enumitem}
\usepackage{siunitx}        % decimal-aligned columns

\usepackage[table,dvipsnames]{xcolor}
\usepackage{type1cm}
\usepackage{multirow}
\usepackage{multicol}
\usepackage{colortbl}
\usepackage{makecell}
\usepackage{adjustbox}
\usepackage{tabularx}
\usepackage{array}
\usepackage{rotating}       % sidewaystable*

\usepackage{graphicx}
\usepackage{wrapfig}
\usepackage{subcaption}
\usepackage{svg}

\usepackage[utf8]{inputenc}
\usepackage[T1]{fontenc}
\usepackage{pifont}
\usepackage{ragged2e}

\newcommand{\cmark}{\ding{51}}
\newcommand{\xmark}{\ding{55}}
\definecolor{darkgreen}{rgb}{0.0, 0.7, 0.0}

\svgsetup{
  inkscapeopt=--export-type=pdf --export-area-page
}

\newcommand*{\AlgName}{\text{LLM-ACES}\@\xspace}
\newcommand*{\modelname}{\text{LLM-ACES}\@\xspace}

\newcommand*{\qwen}{\texttt{Qwen3-32B}\@\xspace}
\newcommand*{\gpt}{\texttt{GPT-4o-mini}\@\xspace}

\title{\AlgName: Closed-Loop Discovery of Dynamical Systems with LLM-Guided Adaptive Search}
\usepackage{booktabs}

\author{
Nikhil Abhyankar \vspace{-0.1in}\quad
Sha Li \quad
Sanchit Kabra \\\\
\textbf{Naren Ramakrishnan}\quad
\textbf{Yulia Gel}\quad
\textbf{Chandan K. Reddy} \\ \\
\normalsize{Virginia Tech}  \\\small{ \texttt{nikhilsa@vt.edu, sanchit23@vt.edu}}
}

\begin{document}

\maketitle

\begin{abstract}
Recovering governing ordinary differential equations (ODEs) from data is a central challenge in modeling dynamical systems across scientific domains. Existing approaches cast discovery as a static inference problem over fixed datasets, assuming that the observed trajectories are sufficiently informative. However, dynamical systems evolve over large state spaces, and limited data can make multiple equations observationally indistinguishable, leading to identifiability gaps and the recovery of incorrect governing equations. To address this, we introduce \AlgName, or \textbf{LLM}-guided \textbf{A}ctive \textbf{C}losed-loop \textbf{E}quation \textbf{S}earch, a closed-loop framework that jointly optimizes symbolic hypothesis construction and adaptive data acquisition. In \AlgName, a large language model (LLM) proposes operator priors that define constrained subspaces of the symbolic search space, within which candidate equations are fit to the observed data. The disagreement among these candidates guides the acquisition of informative trajectories, creating a feedback loop that iteratively refines both the hypothesis space and the discovered dynamics. On 122 ODE systems spanning ODEBench and ODEBase, \modelname achieves the lowest median NMSE, outperforming state-of-the-art baselines while reaching the highest symbolic accuracy up to 70.8\% on ODEBench and 69.5\% on ODEBase across different LLM backbones. Our analysis further shows that \modelname is sample-efficient, achieving better performance with less data. Under observation noise and sparse sampling, LLM-ACES keeps the lowest reconstruction error without inflating expression complexity, whereas baselines tend to add spurious terms that fit the data locally but obscure the governing relationships.

Code: \small \url{https://github.com/scientific-discovery/LLM-ACES}
\end{abstract}

% \vspace{-0.01in}
\section{Introduction}
\vspace{-0.02in}
Discovering governing equations from experimental data is a fundamental challenge in dynamical-systems modeling and a key driver of scientific progress~\citep{kragh2021cosmology, cranmer2023interpretable}. In scientific domains, ordinary differential equations (ODEs) provide compact and interpretable representations of continuous-time dynamics through vector fields~\citep{breakspear2017dynamic, kleinstreuer2018modern, walker2013dynamical}. Although such equations have traditionally been derived from first principles, the availability of high-fidelity data has made data-driven system identification increasingly viable~\citep{bideh2026llm}, thus motivating automated discovery methods based on genetic programming, sparse regression, symbolic regression, and deep learning~\citep{he2022taylor,brunton2016discovering,sun2023symbolic,qian2022dcode}. Recently, large language models (LLMs) have emerged as powerful tools for scientific hypothesis generation, using pre-trained scientific and mathematical knowledge to propose structured candidate laws~\citep{merler2024context, grayeli2024symbolic, shojaee2025llmsr}. However, dynamical equation discovery is not simply a problem of fitting equations to observed trajectories. It is a closed-loop process that also requires acquiring informative observations across diverse conditions to identify the underlying governing dynamics correctly.

\begin{table}[!htbp]
\centering
\vspace{-0.5em}
\caption{\small Comparison of symbolic ODE discovery methods.}
% \vspace{-0.5em}

\label{tab:method-comparison}
\fontsize{7}{7}\selectfont

\begin{tabular}{lccccc}
\toprule
\multirow{2}{*}{\textbf{Method}}
  & \textbf{Hypothesis}
  & \textbf{Closed Loop}
  & \textbf{LLM-induced}
  & \textbf{Data}
  & \textbf{Search} \\
 & \textbf{Refinement}
 & \textbf{Discovery}
 & \textbf{Prior}
 & \textbf{Acquisition}
 & \textbf{Paradigm} \\
\midrule
SINDy~\citep{brunton2016discovering}
  & \textcolor{red}{\xmark} & \textcolor{red}{\xmark} & \textcolor{red}{\xmark} & Passive & Sparse Regression \\
PySR~\citep{cranmer2023interpretable}
  & \textcolor{darkgreen}{\cmark} & \textcolor{red}{\xmark} & \textcolor{red}{\xmark} & Passive & Evolutionary \\
ODEFormer~\citep{dascoli2024odeformer}
  & \textcolor{red}{\xmark} & \textcolor{red}{\xmark} & \textcolor{red}{\xmark} & Passive & Transformer-based \\
% LLM-SR~\citep{shojaee2025llmsr}
  % & \textcolor{darkgreen}{\cmark} & \textcolor{red}{\xmark} & \textcolor{darkgreen}{\cmark} & Passive & Evolutionary \\
LLM-ODE ~\citep{bideh2026llm}
  & \textcolor{darkgreen}{\cmark} & \textcolor{red}{\xmark} & \textcolor{darkgreen}{\cmark} & Passive & Evolutionary \\
{APPS-ODE}~\citep{jiang2025active}
  & \textcolor{darkgreen}{\cmark} & \textcolor{darkgreen}{\cmark} & \textcolor{red}{\xmark} & \textbf{Active} & Grammar-based RL + Adaptive Sampling \\
{Active-SINDy}~\citep{janeiro2025active}
  & \textcolor{darkgreen}{\cmark} & \textcolor{darkgreen}{\cmark} & \textcolor{red}{\xmark} & \textbf{Active}	& Sparse Regression + Adaptive Sampling \\
\midrule
\rowcolor{blue!10}
\textbf{\modelname (ours)}
  & \textbf{\textcolor{darkgreen}{\cmark}} 
  & \textbf{\textcolor{darkgreen}{\cmark}} 
  & \textbf{\textcolor{darkgreen}{\cmark}} 
  & \textbf{Active} 
  & \textbf{Evolutionary + Adaptive Sampling} \\
\bottomrule
%\bottomrule
\end{tabular}
\vspace{-1.5em}
\end{table}

Trajectories initialized from a limited region of the state space can render multiple structurally distinct equations observationally indistinguishable. A candidate equation may fit the observed data well yet predict entirely different behavior under new initial conditions or over longer time horizons~(see Figure~\ref{fig:identifiability_gap} in Appendix~\ref{app:identifiability}). This risk is especially pronounced in nonlinear systems, where small changes in initial conditions lead to dramatically divergent trajectories~\citep{strogatz2001nonlinear}, producing an equation with low training error but incorrect governing structure. Thus, \textit{the central challenge is not only to search over candidate equations, but also to acquire data that actively distinguishes among competing dynamical hypotheses.} This setting arises naturally in self-driving laboratories and Bayesian experimental design, where an agent can adaptively query experimentally feasible conditions under a limited acquisition budget~\citep{chaloner1995bayesian,tom2024self}. Existing methods address parts of this challenge, but not the full closed-loop discovery problem. Classical regression methods search for interpretable equations from fixed datasets but rely on manually specified, static operator spaces and passive observations~\citep{brunton2016discovering, cranmer2023interpretable}. LLM-guided approaches improve hypothesis generation by injecting scientific priors into symbolic search, treating discovery as a passive regression task~\citep{shojaee2025llmsr, bideh2026llm}. Active symbolic discovery methods adaptively acquire new trajectories, but typically do so without using structured symbolic hypothesis spaces as the objects that guide experimentation~\citep{haut2023active, haut2024active, jiang2025active}. This leaves open a fundamental question: \emph{can symbolic hypotheses themselves guide adaptive data acquisition toward more data-efficient and identifiable dynamical-system discovery?}

\begin{figure}[!htbp]
    \vspace{-0.75em}
    \centering
    \includegraphics[width=\linewidth]{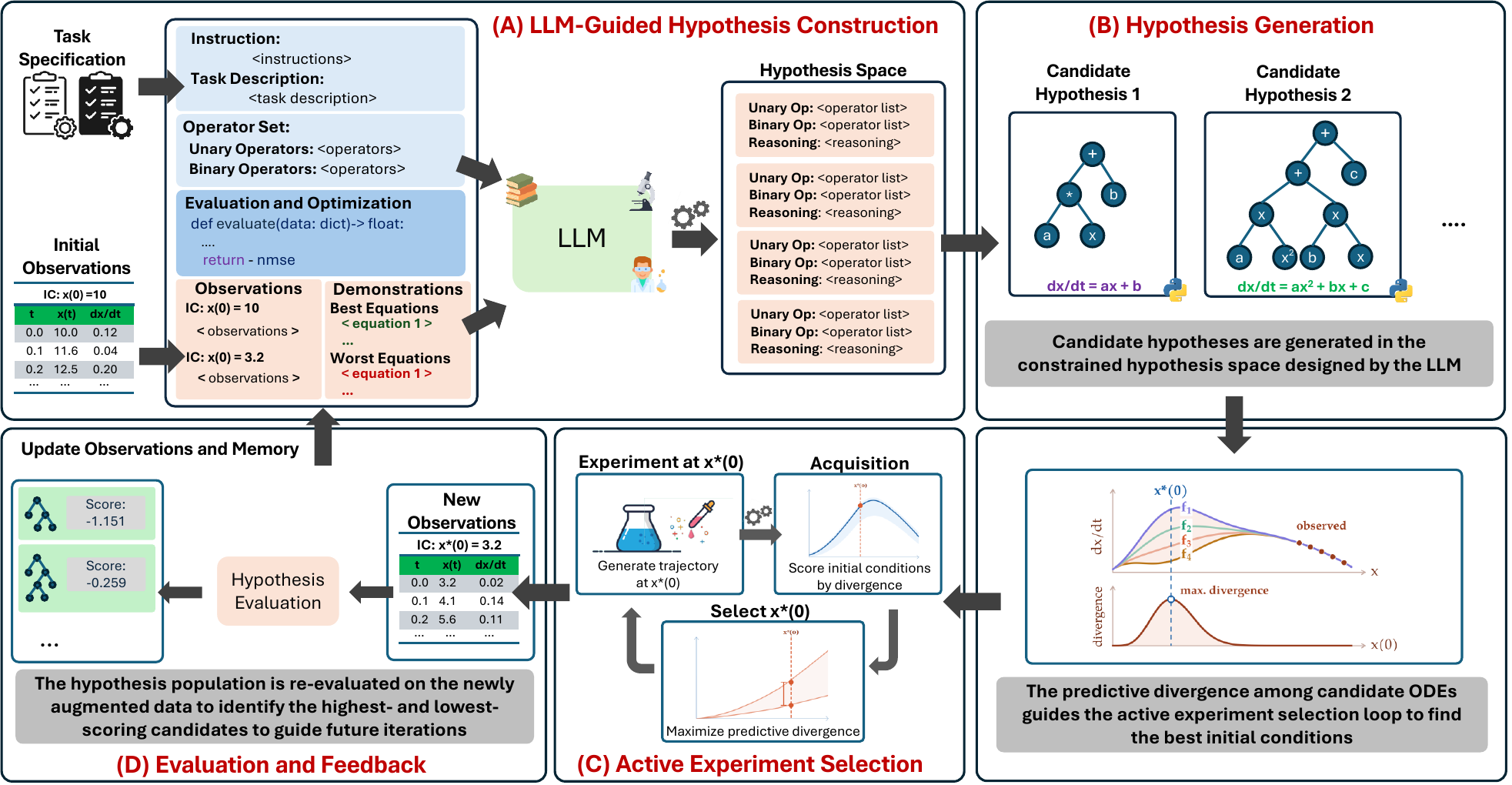}
    \vspace{-1.25em}
 \caption{\small \textbf{Overview of \AlgName}. \AlgName couples LLM-guided hypothesis-space construction with active trajectory acquisition. (A) The LLM induces constrained symbolic search spaces from task information and prior feedback. (B) Candidate equations are generated and optimized within these spaces. (C) New initial conditions are selected by maximizing predictive disagreement among candidate rollouts. (D) Acquired trajectories update the candidate equations and refine the next hypothesis space.}    \vspace{-0.5em}
\label{fig:overview}
\end{figure}

To address this question, we propose \textbf{\AlgName} ({\textbf{LLM}-guided \textbf{A}ctive \textbf{C}losed-loop \textbf{E}quation \textbf{S}earch}), a closed-loop framework that jointly performs symbolic hypothesis construction and adaptive trajectory acquisition. Unlike prior methods that separate equation search from data collection, \AlgName formulates \emph{dynamical-system discovery as an active inference process in which the symbolic hypothesis space and the acquired dataset co-evolve}. As illustrated in Figure~\ref{fig:overview}, \AlgName uses LLMs to induce operator-level priors that constrain the symbolic search spaces, instead of relying on LLMs to output final equations directly~(Figure~\ref{fig:overview}~(A)). Candidate equations to fit the observed data are then instantiated and optimized within these constrained spaces. The resulting population of fitted equations serves as a predictive-disagreement set over the unknown dynamics~(Figure~\ref{fig:overview}~(B)). \AlgName then selects experimentally feasible initial conditions by identifying regions of the state space where candidate equations produce maximally divergent rollouts~(Figure~\ref{fig:overview}~(C)). The newly acquired trajectories are fed back into the discovery loop, thereby refining both the candidate equations and the subsequent LLM-guided construction of the hypothesis space~(Figure~\ref{fig:overview}~(D)). In this way, \AlgName uses candidate hypotheses to resolve identifiability gaps, rather than treating them as passive regression outputs. Table~\ref{tab:method-comparison} compares dynamical-system discovery methods across hypothesis refinement, closed-loop discovery, LLM-induced priors, data acquisition, and search paradigm. Existing methods largely address these dimensions in isolation. Classical approaches~\citep{brunton2016discovering, fasel2022ensemble} rely on fixed symbolic spaces and passive datasets, limiting generalization across dynamical regimes. \texttt{PySR}~\citep{cranmer2023interpretable}, \texttt{LLM-SR}~\citep{shojaee2025llmsr}, and \texttt{LLM-ODE}~\citep{bideh2026llm} improve symbolic search and hypothesis exploration, but still operate on static datasets. \texttt{APPS-ODE}~\citep{jiang2025active} introduces active acquisition for ODE discovery, but does not couple acquisition with LLM-induced symbolic hypothesis-space refinement. In contrast, \AlgName unifies LLM-guided operator-prior induction, iterative symbolic refinement, and predictive-divergence-driven trajectory acquisition within a single closed-loop framework. We instantiate \AlgName with  LLM backbones \texttt{GPT-4o-mini} and \texttt{Qwen3-32B} and evaluate it on ODEBench and ODEBase datasets. Our results show that \AlgName improves equation recovery and sample efficiency compared to prior methods. Further analyses demonstrate the importance of LLM-induced hypothesis spaces, feedback-driven refinement, and disagreement-based data acquisition in resolving identifiability gaps. We summarize our contributions as follows: \vspace{-0.25em}
\begin{itemize}[leftmargin=*]
    \item \textbf{Closed-loop equation discovery.} We propose \AlgName, a unified framework that tightly couples LLM-guided symbolic hypothesis construction with adaptive trajectory acquisition, enabling the hypothesis space and observed dataset to co-evolve through iterative feedback.

    \item \textbf{LLM-induced symbolic search spaces.} We use LLMs to construct domain-informed operator priors that constrain symbolic regression, decoupling hypothesis-space design from equation fitting.

    \item  \textbf{Divergence-driven data acquisition.} We introduce a trajectory acquisition strategy that selects initial conditions by maximizing predictive divergence among candidate equations, directly targeting regions where competing dynamical hypotheses disagree most.

    \item \textbf{Empirical improvements.}  On ODEBench and ODEBase, \AlgName outperforms strong passive and active baselines across reconstruction, generalization, and extrapolation settings, achieving the highest symbolic accuracy while remaining sample-efficient, noise-robust, and interpretable.
\end{itemize}
\vspace{-0.1em}

\vspace{-1.em}
\section{LLM-ACES}
\vspace{-0.5em}
\label{sec:methodology}
\subsection{Problem Formulation}
\label{sec:problem}
\vspace{-.25em}
\paragraph{Dynamical symbolic discovery.}
We consider an autonomous dynamical system
$\dot{\mathbf{u}}(t) = \mathbf{f}^{\star}(\mathbf{u}(t)),$
% \vspace{-0.5em}
where $\mathbf{u}(t)\in\mathbb{R}^d$ is the state and $\mathbf{f}^{\star}:\mathbb{R}^d\rightarrow\mathbb{R}^d$ is the unknown governing vector field. The goal is to recover an interpretable symbolic approximation $\hat{\mathbf{f}}$ from a hypothesis space $\mathcal{H}$ constructed from an operator vocabulary $\mathcal{O}$. Given $\mathcal{D}=\{(\mathbf{u}_i,\dot{\mathbf{u}}_i)\}_{i=1}^N$ with $N$ observations, or derivatives estimated from sampled trajectories, discovery seeks a parsimonious model that best fits the data:
\vspace{-0.35em}
\begin{equation}
\vspace{-0.35em}
    \hat{\mathbf{f}}
    =
    \arg\min_{\mathbf{f}\in\mathcal{H}}
    \mathcal{L}(\mathbf{f};\mathcal{D}),
    \label{eq:sr_objective}
\vspace{-.5em}
\end{equation}

where $\mathcal{L}$ measures prediction error; parsimony is handled by the symbolic-regression backend’s complexity-aware model selection (Appendix~\ref{app:implementation}). Two equations $\mathbf{f}_1, \mathbf{f}_2 \in \mathcal{H}$ are \emph{observationally indistinguishable} on dataset $\mathcal{D}$ if $\mathcal{L}(\mathbf{f}_1;\mathcal{D}) \approx \mathcal{L}(\mathbf{f}_2;\mathcal{D})$ yet $\mathbf{f}_1 \neq \mathbf{f}_2$ structurally. We refer to this as the \emph{identifiability gap} in which distinct candidate dynamics fit the available observations comparably well and cannot be reliably distinguished at the available observation precision.

% We refer to this ambiguity as an \emph{identifiability gap}. When trajectories explore only a limited region of space, structurally distinct equations can remain indistinguishable from the observed data alone, failing to resolve this ambiguity.

\vspace{-0.75em}

\paragraph{Active data acquisition.} The key insight is that dynamical systems, unlike static datasets, can be \emph{queried}
given access to a simulator or experimental oracle $\Omega$. Ideally, the next initial
condition $\mathbf{u}_0$ yields a trajectory $\tau(\mathbf{u}_0)$ that contradicts at least one
competing candidate. Because $\tau(\mathbf{u}_0)$ is unknown before the query, we use the
candidates' predicted disagreement as a proxy. Let $\hat\tau_{\mathbf{f}}(\mathbf{u}_0)$ denote
the rollout of candidate $\mathbf{f}$ from $\mathbf{u}_0$. For any trajectory $\tau^\star$, the
triangle inequality gives
\begin{equation*}
\max\bigl(\|\hat\tau_{\mathbf{f}_1}(\mathbf{u}_0)-\tau^\star\|,\;
          \|\hat\tau_{\mathbf{f}_2}(\mathbf{u}_0)-\tau^\star\|\bigr)
\ge \tfrac12\,\|\hat\tau_{\mathbf{f}_1}(\mathbf{u}_0)-\hat\tau_{\mathbf{f}_2}(\mathbf{u}_0)\|.
\label{eq:active_query}
\end{equation*}
Thus, a large separation between candidate rollouts implies that at least one candidate must incur a substantial trajectory error, motivating predictive disagreement as a heuristic for selecting informative queries.

\vspace{-0.25em}
\subsection{LLM-guided Hypothesis Generation}
\label{sec:llm_generation}
\vspace{-0.5em}
\modelname begins with an LLM-guided hypothesis generation stage. Instead of asking the LLM to directly output final equations, we use an LLM $\pi_\theta$ to generate operator priors that define structured subspaces of the symbolic hypothesis space. Candidate equations are then fitted within these constrained subspaces using a symbolic regression backend. This design separates LLM-based hypothesis-space construction from numerical equation fitting and evaluation.
\vspace{-.5em}
\paragraph{Hypothesis-space exploration.}
At each iteration $t$, the LLM receives a prompt $p_t$ containing (i) task-specific information, (ii) the available operator vocabulary $\mathcal{O}$, (iii) the evaluation objective, and (iv) in-context demonstrations derived from prior iterations. It generates a set of $K$ operator priors
\vspace{-0.25em}
\begin{equation}
\vspace{-0.5em}
    \mathcal{C}_t = \{c_1^{(t)},\ldots,c_K^{(t)}\},
    \qquad
    c_i^{(t)} \sim \pi_\theta(\cdot \mid p_t).
\end{equation}
Each prior $c_i^{(t)}$ specifies a subset of unary operators, such as $\sin$, $\cos$, $\exp$, and $\log$, and binary operators, such as $+$, $-$, $\times$, and $\div$. These priors define constrained symbolic subspaces $\mathcal{H}(c_i^{(t)}) \subseteq \mathcal{H}$ that encode dynamically plausible functional forms. One prior is generated to exploit high-performing operator patterns stored in the experience buffer $\mathcal{E}_{t}$, while the remaining $K-1$ priors encourage structurally distinct operator compositions conditioned on previously explored operators. This induces a structured exploration-exploitation tradeoff over symbolic hypothesis spaces.

\vspace{-.5em}
\paragraph{Candidate equation generation.}
Given the operator prior set $\mathcal{C}_t$, \modelname fits candidate equations within the corresponding constrained symbolic subspaces. For each prior $c_i^{(t)}$, a symbolic regression backend solves $    \mathbf{f}_i^{(t)}
    =
    \arg\min_{\mathbf{f}\in\mathcal{H}(c_i^{(t)})}
    \mathcal{L}(\mathbf{f};\mathcal{D}_t^{\rm tr}).$
% \vspace{-0.5em}
This produces a candidate population $    \mathcal{P}_t^{\rm new}=\{\mathbf{f}_1^{(t)},\ldots,\mathbf{f}_K^{(t)}\},$ spanning multiple LLM-induced symbolic subspaces. Each candidate is evaluated on a held-out validation set $\mathcal{D}_t^{\rm val}$ for every dimension using:
% \vspace{-0.25em}
\begin{equation}
\vspace{-0.5em}
s_i^{(t)} = 
-\mathcal{L}(\mathbf{f}_i^{(t)};\mathcal{D}_t^{\rm val}),
\end{equation}

so that higher scores correspond to lower validation errors. These scores are used to update the experience buffer $\mathcal{E}_t$, which conditions future operator-prior generation. Implementation details are provided in Appendix~\ref{app:hypo_gen}.
\vspace{-0.5em}
\subsection{Hypothesis-Driven Data Acquisition}
\label{sec:acquisition}
\vspace{-0.25em}
We formulate active equation discovery as a coupled optimization process over the candidate hypothesis population and the acquired dataset. Candidate equations identify where additional data (trajectories) would be most informative, while newly acquired trajectories refine both the candidate equations and subsequent hypothesis-space construction.
\vspace{-0.75em}
\paragraph{Predictive-divergence acquisition.} The acquisition committee at round $t$ is $\mathcal Q_t=\mathcal P^{\mathrm{new}}_t$, the candidate systems fitted in that round (at most $K$). For a candidate initial condition $\mathbf u_0$, each $\mathbf f\in\mathcal Q_t$ is rolled out with forward Euler, $\mathbf u_{\ell+1}=\mathbf u_\ell+\Delta t\,\mathbf f(\mathbf u_\ell)$, for $L$ steps, and its states are stacked into $\hat\tau_{\mathbf f}(\mathbf u_0)\in\mathbb R^{Ld}$. Non-finite rollouts are discarded, leaving $\tilde{\mathcal Q}_t(\mathbf u_0)$. With $\bar\tau$ the mean of the retained rollouts,
\begin{equation}
D_{ij}(\mathbf u_0)=\frac{\|\hat\tau_{\mathbf f_i}(\mathbf u_0)-\hat\tau_{\mathbf f_j}(\mathbf u_0)\|_2^2}
                        {Ld\,\|\bar\tau\|_2^2+\varepsilon},
\qquad
A(\mathbf u_0;\mathcal Q_t)=\operatorname*{median}_{i<j} D_{ij}(\mathbf u_0),
\label{eq:active_query}
\end{equation} with $A=0$ if fewer than two rollouts are retained. Normalizing by the committee mean makes $D_{ij}$ symmetric and measures disagreement relative to the magnitude of the predicted trajectories. We approximately maximize $A$ over the feasible box $\mathcal U$ with a Gaussian-process UCB search (Mat\'ern-5/2, $\kappa=2$) over 256 uniformly sampled initial conditions: 3 are scored at random, 7 more are chosen by UCB, and the highest-scoring of the 10 is queried. This GP models the divergence score $A$; it is distinct from the BO baseline (Section~\ref{sec:experiment}), whose GP models PySR's validation error. $A$ is a heuristic: rollouts use forward Euler over $t\in[0,0.9]$ while the oracle uses LSODA, non-finite rollouts are discarded rather than counted as disagreement, and with $K=3$ the median is taken over only three pairs, so one outlying candidate (which enters two of them) still sets the score.

\vspace{-0.75em}
\paragraph{Scoring and experience management.}
LLM-ACES keeps three candidate sets. (i) The acquisition committee $\mathcal{P}^{\mathrm{new}}_t$ contains only the candidates fitted in iteration $t$. The committee is rebuilt from candidates fitted on the current training data, rather than retaining candidates from previous rounds. If fewer than two candidates are fitted in an iteration, no data is acquired in that iteration. (ii) The memory $\mathcal{M}^{(k)}_t$ stores, for each state dimension $k$, every component equation fitted so far. After each acquisition step, all entries are re-scored on the updated validation set. The experience buffer $\mathcal{E}_t$ takes the top-$B$ and bottom-$B$ components per dimension ($B=2$) and is inserted into the next prompt together with all previously proposed operator sets. High-scoring components reinforce effective operator compositions, while low-scoring ones discourage spurious structures. (iii) The final system is assembled per dimension from $\mathcal{M}^{(k)}_T$ (Appendix~\ref{app:llmaces}).

\begin{wrapfigure}{r}{0.52\textwidth}
\vspace{-2.25em}
\begin{minipage}{\linewidth}
\begin{algorithm}[H]
\caption{\modelname}
\label{alg:active-llm-pysr}
\begin{algorithmic}[1]
\Require Initial data $\mathcal{D}_{\rm init}$, oracle $\Omega$, rounds $T$, priors $K$, LLM $\pi_\theta$, metadata $m$, operators $\mathcal{O}$
% \Ensure Best-fit equation $f^\star$

\Statex \textcolor{gray}{$\blacktriangleright $ Initialize}
\State $\mathcal{D}^{0}_{\rm tr}, \mathcal{D}^{0}_{\rm val} \gets \mathcal{D}_{\rm init}[0::2], \mathcal{D}_{\rm init}[1::2]$
\State $\{\mathcal{M}^{(k)}_0\}_k, \mathcal{E}_0 \gets \texttt{Init()}$

\For{$t=0,\ldots,T-1$}
    \Statex \textcolor{gray}{\hspace{1.5em}$\blacktriangleright $ Induce priors}
    \State $\mathcal{C}_t \gets \{c_i^{(t)} \sim \pi_\theta(m, \mathcal{O}, \mathcal{E}_t)\}_{i=1}^{K}$

    \Statex \textcolor{gray}{\hspace{1.5em}$\blacktriangleright $ Fit hypotheses}
    \State $\mathcal{P}^{\rm new}_t \gets \emptyset$
    \ForAll{$c_i^{(t)} \in \mathcal{C}_t$}
        \State $f_i^{(t)} \gets \texttt{FitData}(\mathcal{D}^{t}_{\rm tr}, c_i^{(t)})$
        \State $\mathcal{P}^{\rm new}_t \gets \mathcal{P}^{\rm new}_t \cup \{f_i^{(t)}\}$
    \EndFor
    \State $\mathcal{M}^{(k)}_{t+1} \gets \mathcal{M}^{(k)}_t \cup \{f^{(t)}_{i,k}\}_i$ for all $k$

    \Statex \textcolor{gray}{\hspace{1.5em}$\blacktriangleright $ Acquire data}
    \State $\tau^{(t)} \gets \texttt{AcquireData}(\Omega,\mathcal{P}^{\rm new}_t)$
    \State $\mathcal{D}^{t+1}_{\mathrm{tr}},\, \mathcal{D}^{t+1}_{\mathrm{val}} \leftarrow \texttt{Update}(\mathcal{D}^t_{\mathrm{tr}},\, \mathcal{D}^t_{\mathrm{val}},\, \tau^{(t)})$

    \Statex \textcolor{gray}{\hspace{1.5em}$\blacktriangleright $ Score and update}
    \State $s^{(t+1)} \gets \texttt{Score}(\{\mathcal{M}^{(k)}_{t+1}\}_k,\mathcal{D}^{t+1}_{\rm val})$
    \State $\mathcal{E}_{t+1} \gets \texttt{UpdateMemory}(\{\mathcal{M}^{(k)}_{t+1}\}_k,s^{(t+1)})$
\EndFor
\State $\hat{\mathbf{f}} \gets \bigl(\arg\max_{f \in \mathcal{M}^{(k)}_{T}} s_f^{(T)}\bigr)_{k=1}^{d}$
\State \Return $\hat{\mathbf{f}}$
\end{algorithmic}
\end{algorithm}
\end{minipage}
\vspace{-0.5em}
\end{wrapfigure}

% \vspace{-0.5em}
\subsection{Implementation Details}
% \vspace{-0.25em}

\label{sec:implementation_details}

Algorithm~\ref{alg:active-llm-pysr} summarizes the proposed closed-loop discovery framework. Following~\citep{bideh2026llm}, we omit semantic and physical descriptions of the system from the prompts and label state variables generically ($x_0, x_1, \ldots$), so the LLM relies on \textit{reasoning} and \textit{data-driven feedback} rather than memorized equations. The initial dataset $\mathcal{D}_{\rm init}$ contains 20 oracle samples from the training initial condition, split by alternating indices into 10 training samples ($\mathcal{D}^{0}_{\rm tr}$) and 10 validation samples ($\mathcal{D}^{0}_{\rm val}$), with every successful acquisition adding 10 samples to each split. At each iteration $t$, the LLM $\pi_\theta$ receives task metadata $m$, the experience buffer $\mathcal{E}_{t}$, and all previous operator priors, and proposes up to $K=3$ new priors: one exploiting the best operator patterns in $\mathcal{E}_{t}$, the others exploring operator families not yet represented. Each prior selects, per state dimension, a subset of an operator vocabulary $\mathcal{O}$ of unary (e.g., $\sin$, $\cos$, $\exp$, $\log$) and binary ($+$, $-$, $\times$, $\div$, exponentiation) operators. For each prior, \texttt{PySR}~\citep{cranmer2023interpretable} regresses each derivative component on the full state within that subspace, yielding one candidate system; these form $\mathcal{P}^{\mathrm{new}}_t$, and their component equations are added to the per-dimension memory $\mathcal{M}^{(k)}_t$. $\textsc{AcquireData}(\Omega,\mathcal{P}^{\mathrm{new}}_t)$ then selects the next initial condition via Eq.~\eqref{eq:active_query} (skipped if fewer than two candidates were fitted), and the oracle $\Omega$, implemented with SciPy's~\citep{virtanen2020scipy} \texttt{solve\_ivp} (LSODA), returns a 20-sample trajectory that is split by alternating indices and appended to $\mathcal{D}^{t}_{\rm tr}$ and $\mathcal{D}^{t}_{\rm val}$. All components in memory are then re-scored on the updated validation set, and $\mathcal{E}_{t+1}$ is rebuilt from the two best and two worst components per dimension. After $T$ rounds, the final system is assembled from the best component in each dimension. Further details, prompts, and regression settings are in Appendix~\ref{app:llmaces}.

\vspace{-.5em}
\section{Experiments}
% \vspace{-.5em}
\subsection{Experimental Setup}
% \vspace{-0.25em}
\label{sec:experiment}

\paragraph{Evaluation Metrics.}
\label{sec:metrics}
\vspace{-0.25em}
Following prior work on equation discovery~\citep{shojaeellm, bideh2026mdbench}, we evaluate discovered equations along three complementary axes. (i) \textbf{Data fidelity} quantifies how accurately the learned dynamics reproduce observed trajectories and generalize across regimes and is measured using normalized mean squared error (NMSE). (ii) \textbf{Expression complexity} captures the interpretability and parsimony of symbolic representations, favoring compact and human-interpretable equations. (iii) \textbf{Symbolic accuracy} (SA) measures the mean per-dimension recovery of the underlying functional form and assesses whether the discovered equation is mathematically equivalent to the ground-truth dynamics after replacing numerical coefficients with generic constants. Together, these metrics provide a holistic evaluation, since equations with similar symbolic forms may differ numerically and vice versa.  Appendix~\ref{app:metrics} contains more details on the metrics.

\vspace{-.5em}
\begin{table}[!htbp]
\vspace{-0.05em}
\centering
\setlength{\tabcolsep}{3.5pt}
\caption{\small
\textbf{Performance across ODEBench and ODEBase datasets.}
We report median NMSE (lower is better) for reconstruction, generalization, and extrapolation evaluation, mean expression complexity, and mean symbolic
accuracy (higher is better). Best results are in \textbf{bold} (for complexity, indicates proximity to the benchmark’s mean complexity of $19.3$ for ODEBench and $35.2$ for ODEBase), and second-best results are
\underline{underlined}.}
% \vspace{-0.05em}

\renewcommand{\arraystretch}{1.05}
\resizebox{\textwidth}{!}{%
\begin{tabular}{lccccc|ccccc}
\toprule
& \multicolumn{5}{c|}{\textbf{ODEBench}}
& \multicolumn{5}{c}{\textbf{ODEBase}} \\
\cmidrule(lr){2-6}
\cmidrule(lr){7-11}

\textbf{Method}
& {Recon.} $\downarrow$
& {Gen.} $\downarrow$
& {Ext.} $\downarrow$
& {Comp.}
& {SA (\%)} $\uparrow$
& {Recon.} $\downarrow$
& {Gen.} $\downarrow$
& {Ext.} $\downarrow$
& {Comp.}
& {SA (\%)} $\uparrow$ \\
\midrule

\multicolumn{11}{l}{\textit{\textbf{Passive Symbolic Discovery}}} \\

SINDy
& 5.07e-04 & 5.09e-01 & 1.41e+00 & 11.7 & 20.1
& 1.06e-03 & 1.12e+00 & 3.47e-01 & 12.5 & 10.2 \\

Operon
& 7.95e-05 & 2.57e-01 & 1.84e+00 & 14.0 & 4.0
& 1.49e-05 & 1.46e+00 & 1.41e+00 & 19.0 & 5.4 \\

PySR
& 2.84e-03 & 8.82e-01 & 1.43e+00 & 6.6 & 18.7
& 1.37e-03 & 1.08e+00 & 1.35e+00 & 8.8 & 5.9 \\

E2E
& 3.69e-01 & 1.49e+00 & 2.19e+00 & 52.6 & 0.0
& 1.04e+00 & 1.19e+01 & 1.98e+01 & 83.6 & 0.0 \\

ODEFormer
& 4.61e-03 & 3.83e-01 & 2.04e+00 & 15.9 & 17.1
& 1.64e-02 & 1.15e+01 & 4.09e+00 & 18.5 & 4.8 \\

\midrule
\multicolumn{11}{l}{\textit{\textbf{LLM-guided Symbolic Discovery}}} \\

LLM-only
& 2.20e-08 & 3.45e+00 & 1.79e+00 & 38.9 & 0.0
& 4.52e-08 & 1.06e+01 & 1.11e+00 & 54.9 & 0.0 \\

LLM-ODE
& 4.12e-05 & 4.72e-03 & 2.43e-02 & 25.2 & 9.5
& 7.46e-05 & 8.51e-01 & 6.59e-04 & \underline{32.1} & 3.4 \\

\midrule
\multicolumn{11}{l}{\textit{\textbf{Active Symbolic Discovery}}} \\

Query-by-Committee (QBC)
& 1.81e-09 & 5.07e-09 & 5.69e-08 & 33.7 & 22.6
& 4.97e-09 & 7.47e-06 & 6.38e-06 & 45.2 & 21.5 \\

Bayesian Optimization (BO)
& 1.06e-14 & 9.35e-13 & 3.47e-10 & 22.6 & 62.3
& 8.55e-10 & \underline{4.61e-10} & 5.43e-08
& {31.3} & 65.0 \\

APPS-ODE
& 7.52e-01 & 8.13e-01 & 1.02e+00 & 13.4 & 2.6
& 5.56e-01 & 9.12e-01 & 1.01e+00 & 10.9 & 15.2 \\

\midrule

\rowcolor{blue!10}
\textbf{\modelname \texttt{(GPT)}}
& \textbf{1.33e-17}
& \textbf{8.28e-17}
& \textbf{2.46e-16}
& \underline{17.1}
& \textbf{70.8}
& \underline{2.54e-14}
& 8.50e-10
& \underline{3.05e-12}
& \textbf{33.1}
& \underline{66.1} \\

\rowcolor{blue!10}
\textbf{\modelname \texttt{(Qwen)}}
& \underline{6.30e-16}
& \underline{4.18e-15}
& \underline{1.89e-15}
& \textbf{18.2}
& \underline{69.2}
& \textbf{3.70e-15}
& \textbf{4.09e-15}
& \textbf{3.44e-13}
& 31.2
& \textbf{69.5} \\

\bottomrule
\bottomrule
\end{tabular}}
\vspace{-1.25em}
\label{tab:ode_results}
\end{table}

\vspace{-.75em}

\paragraph{Baselines.}
We compare \modelname against a diverse set of state-of-the-art dynamical system discovery methods spanning \textit{symbolic regression, transformer-based equation generation, LLM-guided discovery}, and \textit{active trajectory acquisition}. Our symbolic regression baselines include \texttt{SINDy}~\citep{brunton2016discovering} and evolutionary approaches such as \texttt{PySR}~\citep{cranmer2023interpretable} and \texttt{Operon}~\citep{burlacu2020operon}. We also consider transformer-based methods, including \texttt{End2End}~(\texttt{E2E})~\cite{kamienny2022end} and \texttt{ODEFormer}~\citep{dascoli2024odeformer}. Among LLM-based approaches, we evaluate \texttt{LLM-ODE}~\citep{bideh2026llm} as well as an \texttt{LLM-only} iterative refinement setting following prior work~\citep{zheng2026newtonbench,kabra2026llm}. For active trajectory acquisition, we compare against \texttt{APPS-ODE}~\citep{jiang2025active}, as well as \texttt{Bayesian optimization (\texttt{BO})} and \texttt{Query-by-Committee (\texttt{QBC})} acquisition strategies~\citep{haut2022active,haut2024active} built on top of \texttt{PySR}. Unlike standard \texttt{PySR}, these active variants iteratively acquire additional trajectories under the same acquisition budget as \modelname. To mitigate dataset recall, all LLM-based methods, including \modelname, are evaluated using anonymized dataset versions (see Section~\ref{app:memory}). All LLM-based baselines use \gpt to generate $1000$ candidate equations. \modelname uses $10$ iterations with up to $3$ priors per round ($30$ LLM calls per dataset) using \gpt as well as \qwen to assess robustness across different LLM backbones. Appendices~\ref{app:baselines} and ~\ref{app:llmaces} contain implementation details for all the baselines and \modelname.
\vspace{-0.75em}

\paragraph{Evaluation Protocol.}
We evaluate all methods on \textbf{ODEBench}~\citep{dascoli2024odeformer} and \textbf{ODEBase}~\citep{luders2022odebase}, comprising a collection of dynamical systems spanning $1$D, $2$D, $3$D, and $4$D state spaces from physics, mathematics, and biology. Following the evaluation protocols of ODEFormer~\citep{dascoli2024odeformer} and MDBench~\citep{bideh2026mdbench}, we assess performance across three trajectory-level settings: reconstruction, generalization, and extrapolation. We report one run per system to limit the computational cost and summarize performance using medians and distributional analyses across benchmark systems. In \emph{reconstruction}, the model is evaluated on trajectories from the training initial condition over $t \in [0,1]$ with $100$ uniformly sampled time steps. In \emph{generalization}, the model trained on the reconstruction dataset is evaluated on the held-out initial condition over the time interval $[0,1]$ with $100$ samples. In \emph{extrapolation} evaluation, we test extrapolation over an extended time range $t \in (1,10]$ with $150$ samples from the training initial condition. These settings jointly evaluate the ability to fit observed dynamics, generalize across initial conditions, and extrapolate beyond the training regime. Additional details on datasets, including dataset names, equations, and stratified subsets used for analyses, are provided in Appendix~\ref{app:dataset}.
% \vspace{-1.em}

% \vspace{-1em}
\vspace{-.5em}
\subsection{Main Results}
\label{sec:main}
\vspace{-0.25em}

Table~\ref{tab:ode_results} summarizes results on ODEBench and ODEBase, containing 63 and 59 dynamical systems, respectively. Reconstruction and extrapolation performance are evaluated using trajectories from a single initial condition, while generalization is assessed on trajectories from a distinct initial condition.
\vspace{-2em}
\paragraph{ODEBench.}
Across all settings, \modelname achieves the strongest overall performance. With the \gpt{} backbone, \modelname attains median reconstruction, generalization, and extrapolation NMSEs of $1.33\times10^{-17}$, $8.28\times10^{-17}$, and $2.46\times10^{-16}$, respectively, outperforming passive and LLM-guided baselines by more than ten orders of magnitude on generalization and extrapolation, and the strongest active baseline, \texttt{BO}, by about six orders on extrapolation ($3.47\times10^{-10}$). Among the baselines, \texttt{BO} is the strongest competitor, followed by \texttt{QBC}. However, these methods recover the correct symbolic structure less reliably, achieving symbolic accuracies of $62.3\%$ and $22.6\%$, respectively. In contrast, \modelname achieves the best symbolic accuracy among all methods ($70.8\%$ with \gpt{} and $69.2\%$ with \qwen{}). The average ground-truth complexity on ODEBench is $19.3$, while \modelname obtains complexities of $17.1$ and $18.2$ with \gpt{} and \qwen{}, respectively. Thus, \modelname recovers equations that are not merely low-error fits but are also close in structural complexity to the underlying governing equations.
\vspace{-1.em}

\paragraph{ODEBase.}
Consistent with ODEBench, \modelname achieves the strongest overall performance. With the \qwen{} backbone, it obtains the best median reconstruction, generalization, and extrapolation NMSEs. Passive symbolic discovery methods often achieve reasonable reconstruction error, but their performance deteriorates substantially on generalization and extrapolation trajectories, suggesting that they fit observed data without reliably identifying the governing structure. Among active baselines, \texttt{BO} is again strongest, followed by \texttt{QBC}, but both lag behind LLM-ACES (Qwen) in predictive accuracy and symbolic recovery. \modelname achieves the best symbolic accuracy, reaching $69.5\%$ with \qwen{} and $66.1\%$ with \gpt{}, compared with $65.0\%$ for \texttt{BO}. The average ground-truth complexity is $35.2$, and \modelname produces equations with complexities of $31.2$ and $33.1$ using \qwen{} and \gpt{}, respectively. In contrast, \texttt{PySR} produces simpler but less faithful expressions, oversimplifying the dynamics~(see Section~\ref{sec:qual_analysis}), while \texttt{QBC} produces more complex expressions without matching \modelname's accuracy. Overall, \modelname provides the strongest balance between predictive accuracy, symbolic fidelity, and ground-truth-aligned expression complexity. Appendix~\ref{app:add_results} includes distributional analysis, robustness to noise and sampling, hyperparameter selection, system recovery, and the role of LLMs in search.
\vspace{-0.5em}

\vspace{-0.25em}
\section{Analysis}
\vspace{-0.35em}
\subsection{Ablation Study}
\vspace{-0.15em}

\begin{figure}[!htbp]
\vspace{-1.em}
    \centering
    \includegraphics[width=0.875\linewidth]{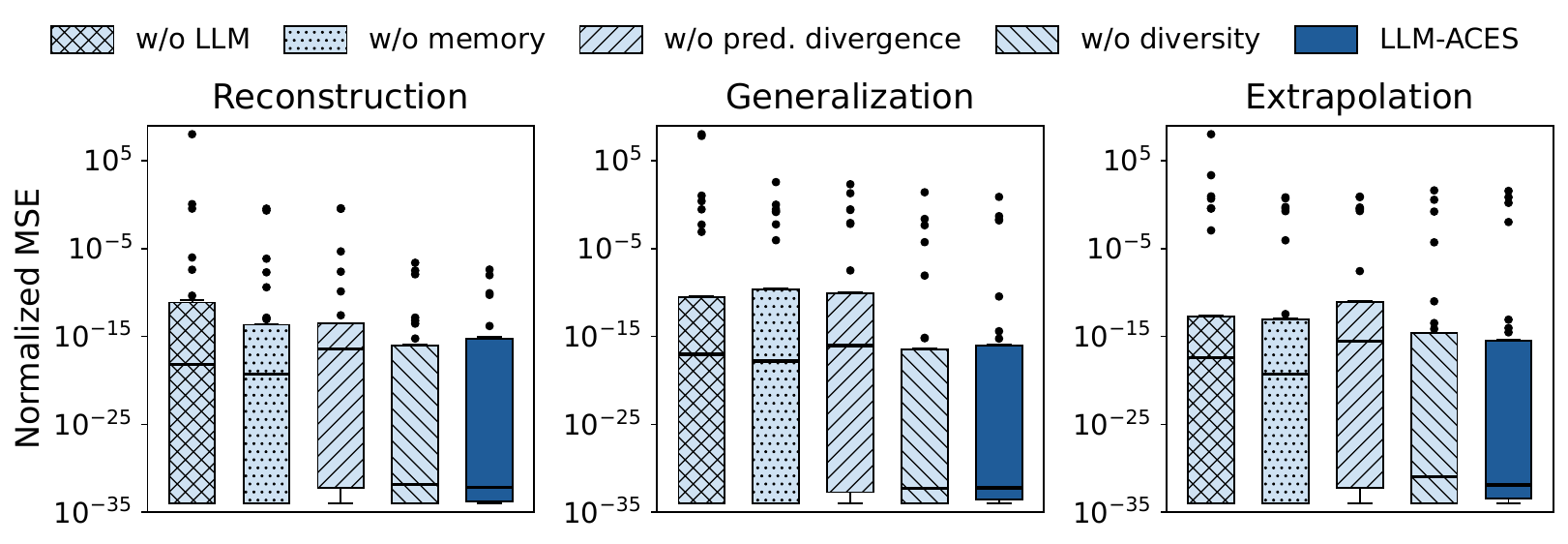}
    \vspace{-0.75em}
    \caption{\small \textbf{Ablation on $30$ ODEBench systems.} Normalized MSE distributions across (i) Reconstruction, (ii) Generalization, and (iii) Extrapolation evaluation settings.}
    \label{fig:ablation}
    \vspace{-1.em}
\end{figure}

We ablate \modelname on $30$ ODEBench systems subset ($10$ each of $1$D, $2$D, and $3$D), with all variants sharing the same query budget and settings (Figure~\ref{fig:ablation}). \modelname attains near machine-precision median NMSE ($\sim\!10^{-32}$) on all three splits and recovers the true equation structure for $26/30$ systems. \textbf{`\textit{w/o} LLM'} replaces the LLM with a UCB bandit over operator subsets; it degrades the most, with the largest outliers and the most failures ($7$ systems with extrapolation NMSE $>10^{-3}$ vs.\ $4$ for the full method), showing that the LLM's semantic priors contribute beyond operator restriction alone. In \textbf{`\textit{w/o} Memory'}, the LLM proposes priors without seeing past equations or tried operator sets; $24\%$ of its priors repeat earlier ones, and accuracy drops on $3$D systems. \textbf{`\textit{w/o} Predictive Divergence'} replaces the acquisition in Eq.~\ref{eq:active_query} with uniform sampling from $\mathcal{U}$; it yields the highest median errors on every split and is the least stable, losing $6$ systems after first solving them ($3$ by the final round), as randomly placed queries fail to discriminate between competing hypotheses. Each of these three variants recovers only $23/30$ systems. \textbf{`\textit{w/o} Diversity'} requests all priors in a single call without asking for distinct ones or filtering repeats; it reaches the same final accuracy ($26/30$), but $20\%$ of its priors repeat previously tried operator sets, so it explores fewer distinct hypothesis spaces than \modelname ($23.8$ vs.\ $27.3$ per run) despite fitting more candidates. Overall, each component targets a distinct failure mode: LLM priors steer the search toward plausible operator families, memory and diversity prevent re-exploring tried hypothesis spaces, and predictive-divergence acquisition gathers the data that separates competing hypotheses; removing any of them either loses systems or wastes search budget. We further analyze memorization and sample efficiency in Sections~\ref{app:memory} and  ~\ref{app:efficiency}.

\subsection{Qualitative Analysis}
\label{sec:qual_analysis}
\begin{figure*}[!htbp]
    % \vspace{-0.25em}
    \centering
    \includegraphics[width=\textwidth]{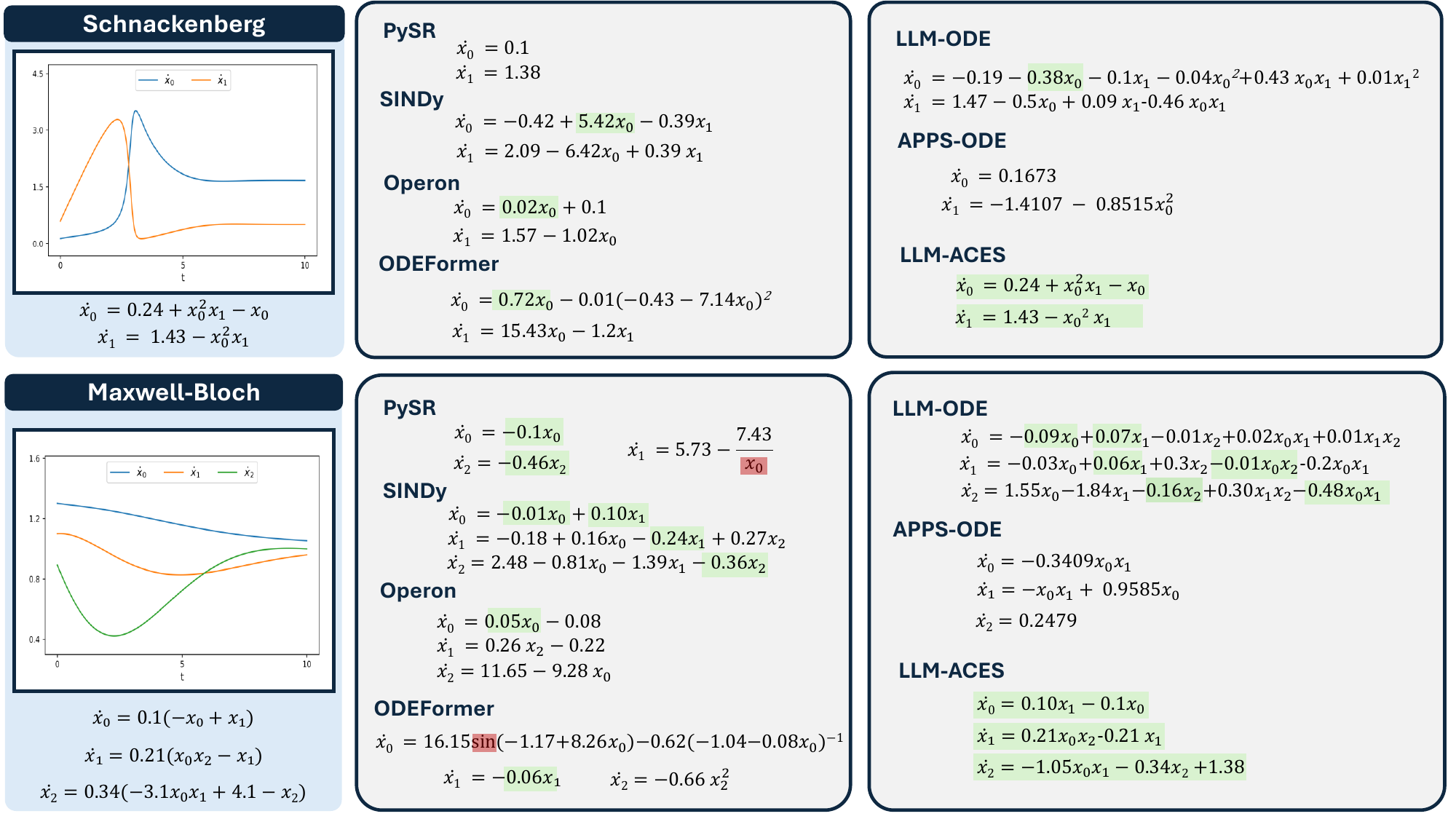}
    \caption{\small \textbf{Discovered equations for the Schnackenberg (top) and Maxwell-Bloch (bottom) systems.} (\textit{Left}): ground-truth governing equations and corresponding phase trajectories. (\textit{Middle}): equations recovered by symbolic and neural ODE discovery baselines. (\textit{Right}): equations recovered by LLM-guided and active learning-based baselines. \colorbox{darkgreen!15}{Green} indicates the correctly recovered symbolic components from ground truth equations.}
    \label{fig:ode_comp}
\end{figure*}

% \vspace{-0.5em}
Figure~\ref{fig:ode_comp} compares the final equations discovered by \modelname{} and representative baselines on the Schnackenberg and Maxwell-Bloch systems. Passive discovery methods that rely solely on fixed datasets often fail to recover the correct governing structure, even when they achieve low error on the observed trajectories. This reflects an identifiability gap: when the available trajectories are not sufficiently informative, structurally distinct equations can remain observationally indistinguishable over the sampled region of state space. We further study this failure mode in Appendix~\ref{app:identifiability}. \texttt{PySR} and \texttt{SINDy} tend to oversimplify the dynamics, returning sparse or nearly linearized expressions that omit essential nonlinear interaction terms. \texttt{Operon} and \texttt{ODEFormer} search over richer hypothesis classes, but frequently introduce spurious polynomial or oscillatory terms that can improve local interpolation while obscuring the underlying mechanism. LLM-guided methods such as \texttt{LLM-ODE} recover some meaningful components but still miss key coupled interactions. Although \texttt{APPS-ODE} performs active data acquisition, its recovered equations are oversimplified, showing that data acquisition alone does not guarantee identifiability, as the queried trajectories must distinguish among competing equations that achieve similar observed-data error. In contrast, \modelname{} selects new trajectories using predictive divergence among candidate equations, directly targeting regions where plausible symbolic hypotheses disagree. Combined with LLM-guided operator priors, this enables \modelname{} to recover compact and correct equations that preserve the underlying dynamics, including the reaction term in Schnackenberg dynamics and the bilinear interaction terms in Maxwell--Bloch dynamics. Appendix~\ref{app:qual_analysis} provides additional trajectory-level comparisons between the dynamics generated by \modelname{} and the corresponding ground-truth systems for both ODEBench and ODEBase datasets.

\vspace{-0.75em}

\subsection{Memorization Analysis}
\label{app:memory}

% \vspace{-.5em}

Many benchmark systems correspond to canonical equations that are likely to have appeared during pretraining, leading to outputs that reproduce verbatim content~\cite{carlini2021extracting, hartmann2023sok}. To disentangle direct dataset recall from genuine discovery, we anonymize ODEBench and ODEBase by replacing all system names, time derivatives, and semantic identifiers with generic labels ($x_0,x_1, \ldots \dot{x}_0, \dot{x}_1, \ldots$), removing any domain-specific terminology from prompts. This eliminates variable-name leakage while preserving the numerical structure of the observations, providing a more faithful test of whether the LLM is reasoning from data or retrieving memorized equations. We study this effect using \texttt{Qwen3-32B} in an iterative refinement setting. At each iteration, the model is provided with the previous best equation along with its corresponding NMSE and asked to generate candidate equations for $100$ iterations.
\begin{wrapfigure}{r}{0.48\linewidth}
    \vspace{1.25em}
    \centering
    \includegraphics[width=\linewidth]{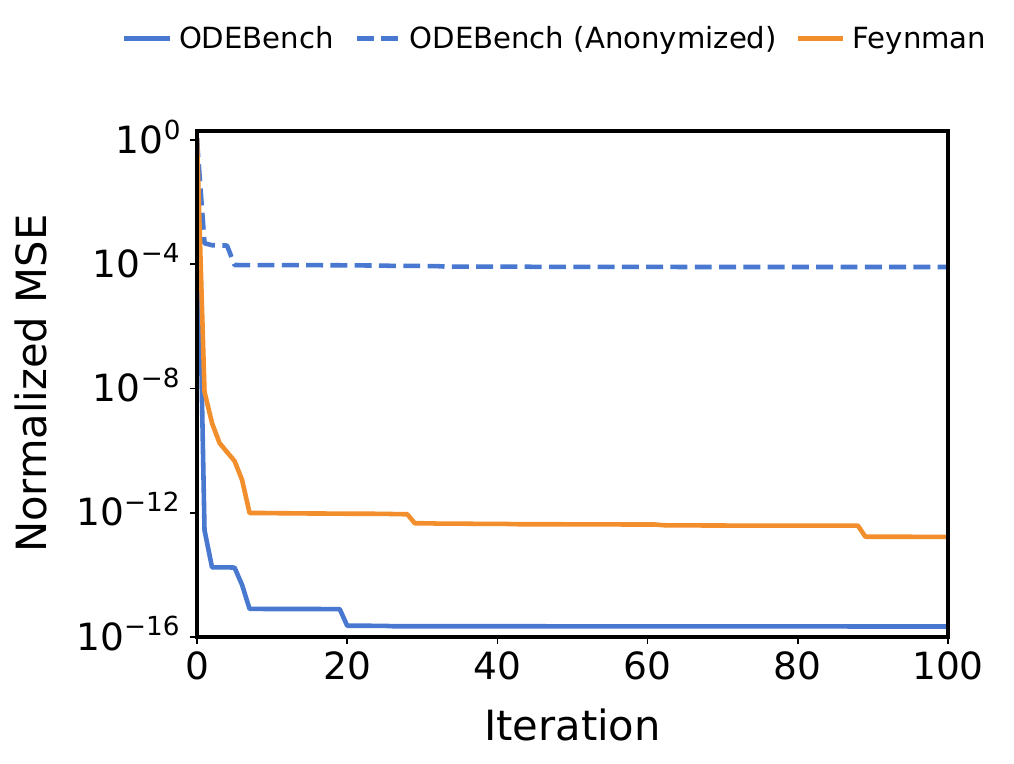}
    \vspace{-1.5em}
    \caption{ \small {Error analysis comparing \qwen on 120 Feynman problems versus 63 ODEBench datasets.}}
    \label{fig:memorization}
    \vspace{-1.50em}
\end{wrapfigure}

Figure~\ref{fig:memorization} compares equation discovery performance across Feynman~\cite{udrescu2020ai}, ODEBench, and an anonymized version of ODEBench. Both Feynman and ODEBench rapidly achieve extremely low NMSE, reaching near-perfect numerical fits after only a few iterations. In contrast, the anonymized version of ODEBench saturates around an NMSE value of $10^{-4}$, indicating that removing semantic cues significantly increases the difficulty of the task. Furthermore, we exactly recovered $31$ of $120$ equations ($25.8\%$) for the Feynman datasets and $17$ out of $63$ for ODEBench, suggesting a high degree of equation recall.  In contrast, anonymizing ODEBench reduces the number of exact recoveries to only $2$ out of $63$ systems. These results indicate that anonymized benchmarks more faithfully assess equation-discovery capability by reducing memorization-based shortcuts and requiring models to infer the underlying functional structure from the available evidence rather than relying on familiarity with known equations.

\vspace{-.5em}
\subsection{Efficiency Analysis}
\label{app:efficiency}
\vspace{-.25em}

\begin{wrapfigure}{r}{0.48\linewidth}
    \vspace{-1.5em}
    \centering
    \includegraphics[width=\linewidth]{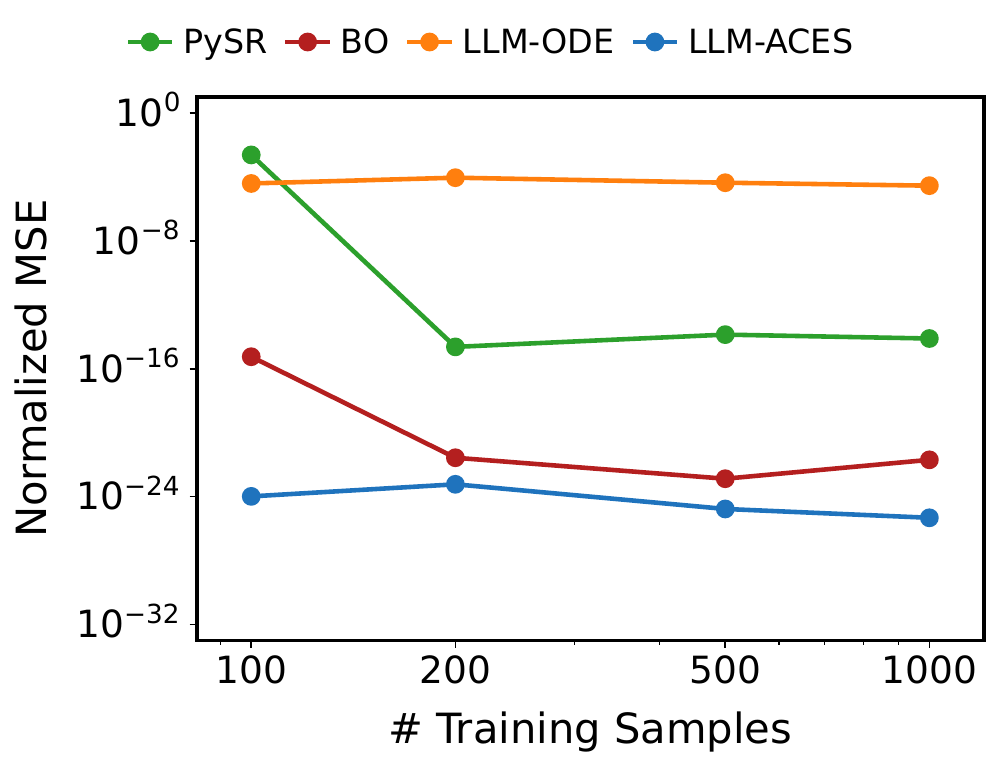}
    \vspace{-1.5em}
    \caption{\small Geometric-mean reconstruction NMSE across the 15-system subset at training budgets of $100, 200, 500$, and $1000$ observations. LLM-ACES achieves the lowest error at each evaluated budget.}
    \label{fig:efficiency}
    \vspace{-.5em}
\end{wrapfigure}

\textcolor{black}{We evaluate the sample efficiency of \modelname on a set of $15$ selected datasets, described in Appendix~\ref{app:dataset}, containing five systems from each of the 1D, 2D, and 3D sets. We compare \modelname against a diverse set of baselines, including symbolic regression (\texttt{PySR}), LLM-based symbolic regression (\texttt{LLM-ODE}), and the Bayesian Optimization variant of \texttt{PySR} (\texttt{BO}), across increasing observation budgets of $100$, $200$, $500$, and $1000$ samples from the respective framework's selected IC(s). Results are aggregated using the geometric mean of NMSE across datasets, which gives each dataset equal weight on the log scale and reduces the influence of large errors relative to arithmetic averaging. As shown in Figure~\ref{fig:efficiency}, additional observations substantially improve the performance of \texttt{PySR} between $100$ and $200$ samples, but its performance quickly saturates around $10^{-14}$ to $10^{-15}$ NMSE thereafter. LLM-ODE exhibits little sensitivity to the number of observations, with reconstruction errors remaining around $10^{-4}$ to $10^{-5}$ across all sample budgets. \texttt{BO} benefits from additional samples, reaching $2.01\times10^{-22}$ at $1000$ observations. Crucially, \modelname outperforms all baselines at \emph{every} observation budget from $100$ samples to $1000$ samples, maintaining the lowest reconstruction error across all evaluated observation budgets beating \texttt{BO} at each level. Even when competing methods are given a significantly larger number of observations ($5-10\times$ data), they cannot surpass \modelname's performance. This shows that \modelname's advantage stems from using observations more effectively by querying the \emph{right data}.} Appendix~\ref{app:add_results} contains additional results and analyses.

\vspace{-0.5em}
\section{Related Work}
\vspace{-.5em}

\paragraph{Discovery in Dynamical Systems.}
Dynamical symbolic regression (SR) seeks governing equations $f$ from trajectories $(t,\mathbf{x}(t))$, often by learning relationships over $(\mathbf{x}(t),\dot{\mathbf{x}}(t))$ pairs despite unavailable or noisy derivatives. Neural and hybrid approaches~\citep{chen2018neural, udrescu2020ai, weilbach2021inferring} provide flexible dynamics models but often sacrifice interpretability or require strong inductive biases. Symbolic methods, including genetic programming~\citep{cranmer2023interpretable, burlacu2020operon} and sparse regression~\citep{brunton2016discovering, messenger2021weak}, instead seek concise interpretable expressions but typically rely on fixed datasets and manually specified operator spaces. Recent neural and transformer-based methods~\citep{valipour2021symbolicgpt, biggio2021neural, li2022transformer, shojaee2023transformer, kamienny2022end} improve scalability by framing equation discovery as sequence prediction, but largely remain passive and do not adaptively acquire data for discovery.

\vspace{-0.5em}
\paragraph{LLMs for Scientific Discovery.}
LLMs have recently been used to guide scientific hypothesis search~\citep{ai4science2023impact, reddy2025towards}, often by proposing candidates evaluated through external oracles in evolutionary or iterative refinement loops~\citep{lehman2023evolution, lange2024large, liu2024large, romera2024mathematical}. In equation discovery, LLM-SR~\citep{shojaee2025llmsr} uses LLMs to guide symbolic regression from problem descriptions, while LLM-ODE~\citep{bideh2026llm} extends this idea to dynamical systems using observed trajectories. LLM-guided search has also been applied to program synthesis, molecular and materials discovery, and scientific optimization~\citep{ma2023eureka, lu2024ai, wang2024efficient, abhyankar2026llema}. Beyond direct candidate generation, \cite{yang2026think} uses LLMs to orchestrate tools and configure downstream models, further illustrating their role in structuring scientific search. These approaches, however, primarily use LLMs to organize or explore the hypothesis space defined by the available evidence; the resulting hypotheses do not themselves determine which observations should be acquired next. \AlgName closes this loop by using LLMs to induce structured symbolic search spaces whose fitted hypotheses directly guide subsequent trajectory acquisition.

\vspace{-0.75em}
\paragraph{Active Learning for Equation Discovery.}

Active learning improves data efficiency by adaptively querying informative samples, including uncertain and diverse examples in deep learning~\citep{settles2009active, Ash2020Deep}. For dynamical systems, related work studies active input design, adaptive trajectory sampling, and exploration strategies for more efficient system identification~\citep{pmlr-v125-wagenmaker20a, JMLR-v23-20-807, pmlr-v190zhao22a, sukhija2023optimistic, haut2022active, haut2023active}. Related symbolic-regression and ODE-discovery methods use active acquisition to select useful observations or phase-space regions~\citep{burbidge2007active,jiang2025active}. Recent work couples LLM-generated mechanisms with hypothesis-guided experiment design~\citep{kabra2026llm}, while related active and PDE discovery methods combine adaptive acquisition or structural search with predefined model classes~\citep{janeiro2025active, du2026llm, huang2026hypothesize}. \AlgName instead couples acquisition with LLM-guided symbolic search-space construction: the LLM proposes operator-level priors, symbolic regression instantiates candidate equations, and their disagreement determines subsequent trajectory acquisition.
%% Conclusion
\vspace{-1.em}
\section{Conclusion}
\vspace{-0.5em}

\label{sec:conclude}
In this work, we introduced \AlgName, a closed-loop framework for dynamical equation discovery that couples LLM-guided symbolic hypothesis construction with active trajectory acquisition. Rather than treating equation discovery as passive inference over a fixed dataset, \AlgName uses disagreement among competing symbolic hypotheses to guide the acquisition of informative trajectories, allowing data collection and equation discovery to refine one another iteratively. Across the ODEBench and ODEBase benchmarks, \AlgName achieves the best performance among the evaluated methods under reconstruction, generalization, and extrapolation evaluation. In particular, \AlgName achieves the lowest median NMSE across all settings on ODEBench and ODEBase with the \gpt{} and \qwen backbones, respectively. \AlgName also attains the highest symbolic accuracy among all methods, reaching $70.8\%$ on ODEBench with \gpt{} and $69.5\%$ on ODEBase with \qwen{}. These improvements are obtained while producing expressions whose complexity is closely aligned with the ground-truth equations, rather than merely minimizing complexity. Ablation studies show that operator-prior induction, memory, and disagreement-based acquisition each help prevent discovery failures, while diversity enforcement improves search efficiency. More broadly, our results suggest that equation discovery should be viewed not as a static regression problem, but as an iterative process in which hypotheses and observations continually inform one another. Symbolic models should serve not only as explanations of existing data, but also as tools for deciding which evidence should be collected next. We hope this perspective will inspire future AI-for-science systems that integrate foundation models, structured priors, and adaptive experimentation to enable more interpretable, data-efficient, and reliable scientific discovery.
\vspace{-0.25em}

\noindent \textbf{Limitations.} 
\AlgName focuses on autonomous ODE systems with query access to an oracle and may not directly generalize to PDEs, stochastic dynamics, or heavily noisy settings without changes to its components. Disagreement-based acquisition weights all candidates equally; implausible or numerically unstable candidates can dominate the score and steer queries toward uninformative regions. The dependence on LLM-induced operator priors and a symbolic-regression backend makes performance sensitive to models, prompts, and search budget.

\vspace{-1.25em}
\section*{Acknowledgements}
\vspace{-.75em}

This research was partially supported by the U.S. National Science Foundation (NSF) under Grant No. 2416728 and Autodesk Research. We thank Modal and Amazon Web Services for providing the computational resources that supported the hosting and implementation of the LLMs in this study.

\bibliographystyle{plain}
\bibliography{neurips_2026}
\newpage
\section*{Impact Statement}
\label{app:impact}
\vspace{-0.75em}
The paper presents \modelname, a closed-loop framework for discovering governing equations of dynamical systems through iterative hypothesis generation, active data acquisition, and feedback. By combining symbolic regression, large language models, and active learning, \modelname aims to accelerate scientific discovery in domains where governing equations are unknown or difficult to derive. The framework could support model discovery in biology, physics, climate science, and engineering, particularly when experiments are costly. However, \modelname relies on LLMs, which may introduce biased or incorrect priors, and the discovered equations may not generalize beyond the observed data regime. The framework also assumes access to a queryable simulator or experimental interface, which may not always be available. We therefore recommend using \modelname as a hypothesis-generation tool with expert oversight and rigorous validation.
\vspace{-1.em}
\section*{Ethics Statement}
\vspace{-0.75em}

This work uses established dynamical-system benchmarks and simulator-generated trajectories and does not involve human subjects, personal data, or interventions on physical systems. LLM-ACES is intended as a research framework rather than an autonomous system for safety-critical deployment; real-world use should incorporate domain-specific feasibility constraints, safety checks, uncertainty assessment, and human oversight. We also mitigate potential benchmark memorization by evaluating LLM-based methods on anonymized versions of the datasets and explicitly analyze its effects.
\vspace{-1em}
\section*{Reproducibility Statement}
\vspace{-0.75em}

We provide the code repository and detailed specifications of the experimental pipeline. Section~\ref{sec:methodology} and Algorithm~\ref{alg:active-llm-pysr} describe LLM-ACES, Appendix~\ref{app:details} documents datasets and evaluation metrics, and Appendix~\ref{app:implementation} reports computational resources, baseline configurations, prompts, operator constraints, symbolic-regression settings, LLM sampling parameters, and the acquisition procedure. Section~\ref{sec:experiment} specifies evaluation protocols and budgets, while Section~\ref{app:memory} describes the benchmark anonymization procedure. As hosted LLM APIs are stochastic and may change over time, exact generations may vary, but we report the model choices, sampling settings, and computational budgets for reproducibility.

\appendix

\vspace{-1.em}
\section*{Appendix}
\vspace{-0.5em}

\section{Experiment Setup}
\label{app:details}

\vspace{-0.5em}
\subsection{Datasets}
\label{app:dataset}
\vspace{-0.5em}

We evaluate all methods on two complementary ODE discovery benchmarks: \textbf{ODEBench}~\cite{dascoli2024odeformer}, which contains predominantly physics-inspired dynamical systems, and \textbf{ODEBase}~\cite{luders2022odebase}, which consists of biologically grounded systems curated from experimental studies. Together, these benchmarks provide a broad evaluation suite spanning nonlinear dynamics and real-world scientific modeling.
% \vspace{-1.5em}
\\
\textbf{ODEBench.}
ODEBench~\citep{dascoli2024odeformer} contains a collection of dynamical systems derived from Steven Strogatz's textbook on nonlinear dynamics~\citep{strogatz2001nonlinear}, together with additional systems sourced from scientific reference materials. The benchmark contains 63 ODE systems covering a diverse range of physical and mathematical phenomena, including population dynamics, oscillatory systems, epidemiological models, chemical reaction networks, chaotic attractors, and classical mechanics. Specifically, it comprises 23 one-dimensional systems, 28 two-dimensional systems, 10 three-dimensional systems, and 2 four-dimensional systems. The benchmark spans a wide range of functional forms, including polynomial, rational, exponential, logarithmic, and trigonometric dynamics, making it a challenging testbed for symbolic equation discovery. The complete list of systems is provided in Tables~\ref{tab:1d_ode}--\ref{tab:3d_ode}. To limit cost, ablations use two fixed ODEBench subsets split by dimension (4D excluded): ODEBench-30 (10 systems each in 1D, 2D and 3D) for the component ablation, and ODEBench-15 (the hardest 5 each from ODEBench-30) for the additional sweeps. Each result is compared with default LLM-ACES on the same systems.
% \vspace{-125em}
\\
\noindent\textbf{ODEBase.}
ODEBase~\cite{luders2022odebase} is a repository of ordinary differential equation systems constructed from curated models in the BioModels database~\cite{le2006biomodels}. The repository was originally developed to facilitate benchmarking and evaluation of symbolic computation methods in systems biology by providing standardized ODE representations derived from SBML models. ODEBase contains mechanistic models originating from a wide range of biological domains, including cancer biology, immunology, virology, pharmacokinetics, metabolic regulation, cell-cycle dynamics, and signaling pathways. From this repository, we select 59 systems for evaluation, comprising 23 two-dimensional and 36 three-dimensional ODEs. Tables~\ref{tab:odebase_2d},~\ref{tab:odebase_3d_part1}, and ~\ref{tab:odebase_3d_part2} report the ODEBase systems.

\vspace{-0.5em}

\begin{table}[!htbp]
\centering
\caption{\small 1-Dimensional ODEBench datasets.}
\vspace{0.25em}
\label{tab:1d_ode}
\scriptsize
\setlength{\tabcolsep}{3pt}
\renewcommand{\arraystretch}{1.15}
\begin{tabular}{r p{4.2cm} p{6.5cm}}
\toprule
\textbf{ID} & \textbf{System} & \textbf{Equation} \\
\midrule
1 & RC-circuit (charging capacitor) & $\dot{x}_{0} = \frac{0.7 - \frac{x_0}{1.2}}{2.31}$ \\ \midrule
2 & Population growth (naive) & $\dot{x}_{0} = 0.23x_0$ \\ \midrule
3 & Population growth with carrying capacity & $\dot{x}_{0} = 0.79x_0\left(1 - \frac{x_0}{74.3}\right)$ \\ \midrule
4 & RC-circuit with nonlinear resistor & $\dot{x}_{0} = -0.5 + \frac{1}{e^{0.5 - x_0/0.96} + 1}$ \\ \midrule
5 & Falling object with air resistance & $\dot{x}_{0} = 9.81 - 0.0021175x_0^2$ \\ \midrule
6 & Autocatalysis & $\dot{x}_{0} = 2.1x_0 - 0.5x_0^2$ \\ \midrule
7 & Gompertz law & $\dot{x}_{0} = 0.032x_0\log(2.29x_0)$ \\ \midrule
8 & Logistic with Allee effect & $\dot{x}_{0} = 0.14x_0\left(-1 + \frac{x_0}{4.4}\right)\left(1 - \frac{x_0}{130}\right)$ \\ \midrule
9 & Language death model & $\dot{x}_{0} = 0.32(1-x_0) - 0.28x_0$ \\ \midrule
10 & Refined language model & $\dot{x}_{0} = 0.2x_0^{1.2}(1-x_0) - 0.8x_0(1-x_0)^{1.2}$ \\ \midrule
11 & Critical slowing down & $\dot{x}_{0} = -x_0^3$ \\ \midrule
12 & Photons in laser & $\dot{x}_{0} = 1.8x_0 - 0.1107x_0^2$ \\ \midrule
13 & Rotating hoop & $\dot{x}_{0} = 0.0981(9.7\cos x_0 - 1)\sin x_0$ \\ \midrule
14 & Budworm model & $\dot{x}_{0} = 0.78x_0\left(1-\frac{x_0}{81}\right) - \frac{0.9x_0^2}{21.2^2+x_0^2}$ \\ \midrule
15 & Budworm (dimensionless) & $\dot{x}_{0} = 0.4x_0\left(1-\frac{x_0}{95}\right) - \frac{x_0^2}{x_0^2+1}$ \\ \midrule
16 & Landau equation & $\dot{x}_{0} = 0.1x_0 - 0.04x_0^3 + 0.001x_0^5$ \\ \midrule
17 & Logistic + harvesting & $\dot{x}_{0} = 0.4x_0\left(1-\frac{x_0}{100}\right) - 0.3$ \\ \midrule
18 & Improved harvesting & $\dot{x}_{0} = 0.4x_0\left(1-\frac{x_0}{100}\right) - \frac{0.24x_0}{50+x_0}$ \\ \midrule
19 & Logistic (dimensionless) & $\dot{x}_{0} = -\frac{0.08x_0}{0.8+x_0} + x_0(1-x_0)$ \\ \midrule
20 & Gene switching & $\dot{x}_{0} = 0.1 - 0.55x_0 + \frac{x_0^2}{x_0^2+1}$ \\ \midrule
21 & Reduced SIR & $\dot{x}_{0} = 1.2 - 0.2x_0 - e^{-x_0}$ \\ \midrule
22 & Protein activation & $\dot{x}_{0} = 1.4 + \frac{0.4x_0^5}{123+x_0^5} - 0.89x_0$ \\ \midrule
23 & Driven pendulum & $\dot{x}_{0} = 0.21 - \sin(x_0)$ \\
\bottomrule \bottomrule
\end{tabular}
\end{table}

\begin{table}[!htbp]
\centering
\caption{\small 2-Dimensional ODEBench datasets.}
\vspace{0.25em}
\label{tab:2d_ode}
\scriptsize
\setlength{\tabcolsep}{3pt}
\fontsize{7}{7.5}\selectfont
\renewcommand{\arraystretch}{1.0}
\begin{tabular}{r p{4.0cm} p{7.0cm}}
\toprule
\textbf{ID} & \textbf{System} & \textbf{Equations} \\
\midrule

\multirow{2}{*}{24} & \multirow{2}{=}{Harmonic oscillator}
& $\dot{x}_{0}=x_{1}$ \\
& & $\dot{x}_{1}=-2.1x_{0}$ \\ \midrule

\multirow{2}{*}{25} & \multirow{2}{=}{Damped oscillator}
& $\dot{x}_{0}=x_{1}$ \\
& & $\dot{x}_{1}=-4.5x_{0}-0.43x_{1}$ \\ \midrule

\multirow{2}{*}{26} & \multirow{2}{=}{Lotka--Volterra competition}
& $\dot{x}_{0}=x_{0}(3-2x_{1}-x_{0})$ \\
& & $\dot{x}_{1}=x_{1}(2-x_{0}-x_{1})$ \\ \midrule

\multirow{2}{*}{27} & \multirow{2}{=}{Lotka--Volterra}
& $\dot{x}_{0}=x_{0}(1.84-1.45x_{1})$ \\
& & $\dot{x}_{1}=-x_{1}(3-1.62x_{0})$ \\ \midrule

\multirow{2}{*}{28} & \multirow{2}{=}{Pendulum}
& $\dot{x}_{0}=x_{1}$ \\
& & $\dot{x}_{1}=-0.9\sin(x_{0})$ \\ \midrule

\multirow{2}{*}{29} & \multirow{2}{=}{Dipole system}
& $\dot{x}_{0}=0.65x_{0}x_{1}$ \\
& & $\dot{x}_{1}=-x_{0}^{2}+x_{1}^{2}$ \\ \midrule

\multirow{2}{*}{30} & \multirow{2}{=}{RNA catalysis}
& $\dot{x}_{0}=x_{0}(-1.61x_{0}x_{1}+x_{1})$ \\
& & $\dot{x}_{1}=x_{1}(-1.61x_{0}x_{1}+x_{0})$ \\ \midrule

\multirow{2}{*}{31} & \multirow{2}{=}{SIR infection}
& $\dot{x}_{0}=-0.4x_{0}x_{1}$ \\
& & $\dot{x}_{1}=0.4x_{0}x_{1}-0.314x_{1}$ \\ \midrule

\multirow{2}{*}{32} & \multirow{2}{=}{Double well oscillator}
& $\dot{x}_{0}=x_{1}$ \\
& & $\dot{x}_{1}=-0.18x_{1}-x_{0}^{3}+x_{0}$ \\ \midrule

\multirow{2}{*}{33} & \multirow{2}{=}{Glider}
& $\dot{x}_{0}=-0.08x_{0}^{2}-\sin(x_{1})$ \\
& & $\dot{x}_{1}=x_{0}-\frac{\cos(x_{1})}{x_{0}}$ \\ \midrule

\multirow{2}{*}{34} & \multirow{2}{=}{Rotating hoop}
& $\dot{x}_{0}=x_{1}$ \\
& & $\dot{x}_{1}=(-0.93+\cos(x_{0}))\sin(x_{0})$ \\ \midrule

\multirow{2}{*}{35} & \multirow{2}{=}{Shear flow dynamics}
& $\dot{x}_{0}=\cos(x_{0})\cot(x_{1})$ \\
& & $\dot{x}_{1}=(4.2\sin^{2}(x_{1})+\cos^{2}(x_{1}))\sin(x_{0})$ \\ \midrule

\multirow{2}{*}{36} & \multirow{2}{=}{Nonlinear damped pendulum}
& $\dot{x}_{0}=x_{1}$ \\
& & $\dot{x}_{1}=-0.07x_{1}\cos(x_{0})-x_{1}-\sin(x_{0})$ \\ \midrule

\multirow{2}{*}{37} & \multirow{2}{=}{Van der Pol}
& $\dot{x}_{0}=x_{1}$ \\
& & $\dot{x}_{1}=-0.43x_{1}(x_{0}^{2}-1)-x_{0}$ \\ \midrule

\multirow{2}{*}{38} & \multirow{2}{=}{Van der Pol (Strogatz)}
& $\dot{x}_{0}=3.37\!\left(-\frac{x_{0}^{3}}{3}+x_{0}+x_{1}\right)$ \\
& & $\dot{x}_{1}=-\frac{x_{0}}{3.37}$ \\ \midrule

\multirow{2}{*}{39} & \multirow{2}{=}{Glycolytic oscillator}
& $\dot{x}_{0}=2.4x_{1}+x_{0}^{2}x_{1}-x_{0}$ \\
& & $\dot{x}_{1}=-2.4x_{0}+0.07-x_{0}^{2}x_{1}$ \\ \midrule

\multirow{2}{*}{40} & \multirow{2}{=}{Duffing}
& $\dot{x}_{0}=x_{1}$ \\
& & $\dot{x}_{1}=0.886x_{1}(1-x_{0}^{2})-x_{0}$ \\ \midrule

\multirow{2}{*}{41} & \multirow{2}{=}{Cell cycle (Tyson)}
& $\dot{x}_{0}=15.3(0.001+x_{0}^{2})(-x_{0}+x_{1})-x_{0}$ \\
& & $\dot{x}_{1}=0.3-x_{0}$ \\ \midrule

\multirow{2}{*}{42} & \multirow{2}{=}{Chemical reaction model}
& $\dot{x}_{0}=8.9-\frac{4.0x_{0}x_{1}}{x_{0}^{2}+1}-x_{0}$ \\
& & $\dot{x}_{1}=1.4x_{0}\left(-\frac{x_{1}}{x_{0}^{2}+1}+1\right)$ \\ \midrule

\multirow{2}{*}{43} & \multirow{2}{=}{Driven pendulum}
& $\dot{x}_{0}=x_{1}$ \\
& & $\dot{x}_{1}=1.67-0.64x_{1}-\sin(x_{0})$ \\ \midrule

\multirow{2}{*}{44} & \multirow{2}{=}{Quadratic damping}
& $\dot{x}_{0}=x_{1}$ \\
& & $\dot{x}_{1}=1.67-0.64x_{1}|x_{1}|-\sin(x_{0})$ \\ \midrule

\multirow{2}{*}{45} & \multirow{2}{=}{Gray--Scott}
& $\dot{x}_{0}=0.5(1-x_{0})-x_{0}x_{1}^{2}$ \\
& & $\dot{x}_{1}=-0.02x_{1}+x_{0}x_{1}^{2}$ \\ \midrule

\multirow{2}{*}{46} & \multirow{2}{=}{Bar magnets}
& $\dot{x}_{0}=0.33\sin(x_{0}-x_{1})-\sin(x_{0})$ \\
& & $\dot{x}_{1}=-0.33\sin(x_{0}-x_{1})-\sin(x_{1})$ \\ \midrule

\multirow{2}{*}{47} & \multirow{2}{=}{Binocular rivalry}
& $\dot{x}_{0}=-x_{0}+\frac{1}{e^{4.89x_{1}-1.4}+1}$ \\
& & $\dot{x}_{1}=-x_{1}+\frac{1}{e^{4.89x_{0}-1.4}+1}$ \\ \midrule

\multirow{2}{*}{48} & \multirow{2}{=}{Bacterial respiration}
& $\dot{x}_{0}=18.3-\frac{x_{0}x_{1}}{0.48x_{0}^{2}+1}-x_{0}$ \\
& & $\dot{x}_{1}=11.23-\frac{x_{0}x_{1}}{0.48x_{0}^{2}+1}$ \\ \midrule

\multirow{2}{*}{49} & \multirow{2}{=}{Brusselator}
& $\dot{x}_{0}=3.1x_{0}^{2}x_{1}-4.03x_{0}+1$ \\
& & $\dot{x}_{1}=3.03x_{0}-3.1x_{0}^{2}x_{1}$ \\ \midrule

\multirow{2}{*}{50} & \multirow{2}{=}{Schnackenberg}
& $\dot{x}_{0}=0.24+x_{0}^{2}x_{1}-x_{0}$ \\
& & $\dot{x}_{1}=1.43-x_{0}^{2}x_{1}$ \\ \midrule

\multirow{2}{*}{51} & \multirow{2}{=}{Oscillator death}
& $\dot{x}_{0}=1.432+\sin(x_{1})\cos(x_{0})$ \\
& & $\dot{x}_{1}=0.972+\sin(x_{1})\cos(x_{0})$ \\

\bottomrule \bottomrule
\end{tabular}
\vspace{-1em}
\end{table}

\begin{table}[!htbp]
\vskip 0.005in
\centering
\caption{\small 3-Dimensional and 4-Dimensional ODEBench datasets.}
\vskip 0.025in
\label{tab:3d_ode}
\scriptsize
\setlength{\tabcolsep}{3pt}
\renewcommand{\arraystretch}{1.15}
\begin{tabular}{r p{4.0cm} p{7.5cm}}
\toprule
\textbf{ID} & \textbf{System} & \textbf{Equations} \\
\midrule

\multirow{3}{*}{52} & \multirow{3}{=}{Maxwell--Bloch}
& $\dot{x}_{0}=0.1(-x_{0}+x_{1})$ \\
& & $\dot{x}_{1}=0.21(x_{0}x_{2}-x_{1})$ \\
& & $\dot{x}_{2}=0.34(-3.1x_{0}x_{1}+4.1-x_{2})$ \\ \midrule

\multirow{3}{*}{53} & \multirow{3}{=}{Apoptosis model}
& $\dot{x}_{0}=0.1-0.05x_{0}-\frac{0.4x_{0}x_{1}}{x_{0}+0.1}$ \\
& & $\dot{x}_{1}=0.4x_{0}x_{1}-0.5x_{1}x_{2}-0.1x_{1}$ \\
& & $\dot{x}_{2}=0.5x_{1}x_{2}-0.5x_{2}$ \\ \midrule

\multirow{3}{*}{54} & \multirow{3}{=}{Lorenz periodic}
& $\dot{x}_{0}=5.1(-x_{0}+x_{1})$ \\
& & $\dot{x}_{1}=12x_{0}-x_{0}x_{2}-x_{1}$ \\
& & $\dot{x}_{2}=-1.67x_{2}+x_{0}x_{1}$ \\ \midrule

\multirow{3}{*}{55} & \multirow{3}{=}{Lorenz complex}
& $\dot{x}_{0}=10(-x_{0}+x_{1})$ \\
& & $\dot{x}_{1}=99.96x_{0}-x_{0}x_{2}-x_{1}$ \\
& & $\dot{x}_{2}=-(8/3)x_{2}+x_{0}x_{1}$ \\ \midrule

\multirow{3}{*}{56} & \multirow{3}{=}{Lorenz chaotic}
& $\dot{x}_{0}=10(-x_{0}+x_{1})$ \\
& & $\dot{x}_{1}=28x_{0}-x_{0}x_{2}-x_{1}$ \\
& & $\dot{x}_{2}=-(8/3)x_{2}+x_{0}x_{1}$ \\ \midrule

\multirow{3}{*}{57} & \multirow{3}{=}{R\"ossler stable}
& $\dot{x}_{0}=5(-x_{1}-x_{2})$ \\
& & $\dot{x}_{1}=5(-0.2x_{1}+x_{0})$ \\
& & $\dot{x}_{2}=5(0.2+x_{2}(-5.7+x_{0}))$ \\ \midrule

\multirow{3}{*}{58} & \multirow{3}{=}{R\"ossler periodic}
& $\dot{x}_{0}=0.1(-x_{1}-x_{2})$ \\
& & $\dot{x}_{1}=0.1(-0.2x_{1}+x_{0})$ \\
& & $\dot{x}_{2}=0.1(0.2+x_{2}(-5.7+x_{0}))$ \\ \midrule

\multirow{3}{*}{59} & \multirow{3}{=}{R\"ossler chaotic}
& $\dot{x}_{0}=0.2(-x_{1}-x_{2})$ \\
& & $\dot{x}_{1}=0.2(-0.2x_{1}+x_{0})$ \\
& & $\dot{x}_{2}=0.2(0.2+x_{2}(-5.7+x_{0}))$ \\ \midrule

\multirow{3}{*}{60} & \multirow{3}{=}{Aizawa attractor}
& $\dot{x}_{0}=-0.65x_{1}+x_{0}(-0.7+x_{2})$ \\
& & $\dot{x}_{1}=0.65x_{0}+x_{1}(-0.7+x_{2})$ \\
& & $\dot{x}_{2}=0.6+0.95x_{2}-\frac{x_{2}^{3}}{3}-x_{0}^{2}-x_{1}^{2}+0.25x_{2}x_{0}^{3}$ \\ \midrule

\multirow{3}{*}{61} & \multirow{3}{=}{Chen--Lee attractor}
& $\dot{x}_{0}=5x_{0}-x_{1}x_{2}$ \\
& & $\dot{x}_{1}=-10x_{1}+x_{0}x_{2}$ \\
& & $\dot{x}_{2}=-3.8x_{2}+\frac{x_{0}x_{1}}{3}$ \\ \midrule

\multirow{4}{*}{62} & \multirow{4}{=}{Binocular rivalry (4D)}
& $\dot{x}_{0}=-x_{0}+\frac{1}{e^{0.89x_{2}+0.4x_{1}-1.4}+1}$ \\
& & $\dot{x}_{1}=x_{0}-x_{1}$ \\
& & $\dot{x}_{2}=-x_{2}+\frac{1}{e^{0.89x_{0}+0.4x_{3}-1.4}+1}$ \\
& & $\dot{x}_{3}=x_{2}-x_{3}$ \\ \midrule

\multirow{4}{*}{63} & \multirow{4}{=}{SEIR model}
& $\dot{x}_{0}=-0.28x_{0}x_{2}$ \\
& & $\dot{x}_{1}=-0.47x_{1}+0.28x_{0}x_{2}$ \\
& & $\dot{x}_{2}=0.47x_{1}-0.3x_{2}$ \\
& & $\dot{x}_{3}=0.3x_{2}$ \\

\bottomrule \bottomrule
\end{tabular}
% \vspace{-0.75em}
\end{table}

\begin{table}[!htbp]
\centering
\caption{\small 2-Dimensional ODEBase datasets. *{Systems 69 and 79 come from different BioModels entries that reduce to the same ODEs in ODEBase.}}
\label{tab:odebase_2d}
\scriptsize
\setlength{\tabcolsep}{3pt}
\renewcommand{\arraystretch}{1.12}
\begin{tabular}{r p{3.4cm} p{9.7cm}}
\toprule
\textbf{ID} & \textbf{System} & \textbf{Equations} \\
\midrule

\multirow{2}{*}{64} & \multirow{2}{=}{MPF and Cyclin Oscillations}
& $\dot{x}_{0} = 1.0x_{0}^2x_{1} - 10.0x_{0}/(x_{0}+1.0) + 3.466x_{1}$ \\
& & $\dot{x}_{1} = 1.2 - 1.0x_{0}$ \\ \midrule

\multirow{2}{*}{65} & \multirow{2}{=}{Cancer--Immune System Competition}
& $\dot{x}_{0} = -0.03125x_{0}^2 - 0.125x_{0}x_{1} + 0.0625x_{0}$ \\
& & $\dot{x}_{1} = -0.08594x_{0}x_{1} - 0.03125x_{1}^2 + 0.03125x_{1}$ \\ \midrule

\multirow{2}{*}{66} & \multirow{2}{=}{FitzHugh--Nagumo Nerve Membrane}
& $\dot{x}_{0} = -1.0x_{0}^3 + 3.0x_{0} + 3.0x_{1} - 1.2$ \\
& & $\dot{x}_{1} = -0.3333x_{0} - 0.2667x_{1} + 0.2333$ \\ \midrule

\multirow{2}{*}{67} & \multirow{2}{=}{One-hit Neuronal Cell Death}
& $\dot{x}_{0} = -0.278x_{0}$ \\
& & $\dot{x}_{1} = -0.223x_{1}$ \\ \midrule

\multirow{2}{*}{68} & \multirow{2}{=}{Stem Cell Simple Model}
& $\dot{x}_{0} = 0.004x_{0} + 0.004x_{1}/(0.01x_{0}^{1.0}+1.0)$ \\
& & $\dot{x}_{1} = 0.006x_{0} - 0.003x_{1} - 0.004x_{1}/(0.01x_{0}^{1.0}+1.0)$ \\ \midrule

\multirow{2}{*}{69} & \multirow{2}{=}{Alzheimer Acetylcholine Positive Feedback*}
& $\dot{x}_{0} = -0.007x_{0}x_{1}$ \\
& & $\dot{x}_{1} = -0.004x_{0} - 0.01x_{1} + 0.33$ \\ \midrule

\multirow{2}{*}{70} & \multirow{2}{=}{Bistable Schl\"{o}gl Model}
& $\dot{x}_{0} = -0.00096x_{0}^{3} + 0.1229x_{0}^{2} - 3.072x_{0} + 12.5$ \\
& & $\dot{x}_{1} = 0.00096x_{0}^{3} - 0.1229x_{0}^{2} + 3.072x_{0} - 12.5$ \\ \midrule

\multirow{2}{*}{71} & \multirow{2}{=}{Lotka--Volterra CAR-T / Tumour}
& $\dot{x}_{0} = 0.002x_{0}x_{1} - 0.16x_{0}$ \\
& & $\dot{x}_{1} = 0.15x_{1}$ \\ \midrule

\multirow{2}{*}{72} & \multirow{2}{=}{Cytokine Inflammation (Rheumatoid Arthritis)}
& $\dot{x}_{0} = -x_{0} + 3.5x_{1}^{2}/(x_{1}^{2}+0.25)$ \\
& & $\dot{x}_{1} = 1.0x_{1}^{2}/(x_{0}^{2}x_{1}^{2}+1.0x_{0}^{2}+1.0x_{1}^{2}+1.0) - 1.25x_{1} + 0.025/(x_{0}^{2}+1.0)$ \\ \midrule

\multirow{2}{*}{73} & \multirow{2}{=}{Calcium Oscillations}
& $\dot{x}_{0} = -5.0x_{0}x_{1}^{4.0}/(x_{1}^{4.0}+81.0) - 0.01x_{0} + 2.0x_{1}$ \\
& & $\dot{x}_{1} = 5.0x_{0}x_{1}^{4.0}/(x_{1}^{4.0}+81.0) + 0.01x_{0} - 3.0x_{1} + 1.0$ \\ \midrule

\multirow{2}{*}{74} & \multirow{2}{=}{Acute Myeloid Leukaemia}
& $\dot{x}_{0} = -0.1x_{0} + 0.3x_{0}/(0.5x_{0}+0.5x_{1}+1.0)$ \\
& & $\dot{x}_{1} = -0.1x_{1} + 0.3x_{1}/(0.5x_{0}+0.5x_{1}+1.0)$ \\ \midrule

\multirow{2}{*}{75} & \multirow{2}{=}{Oncolytic M1 Virus--Sensitive/Normal (SN)}
& $\dot{x}_{0} = -0.2x_{0}x_{1} - 0.02x_{0} + 0.02$ \\
& & $\dot{x}_{1} = 0.16x_{0}x_{1} - 0.03x_{1}$ \\ \midrule

\multirow{2}{*}{76} & \multirow{2}{=}{Bone Marrow Invasion (Absolute)}
& $\dot{x}_{0} = -0.2x_{0}^{2} + 0.1x_{0}$ \\
& & $\dot{x}_{1} = -1.0x_{0}x_{1} - 0.8x_{1}^{2} + 0.7x_{1}$ \\ \midrule

\multirow{2}{*}{77} & \multirow{2}{=}{Reversible Isomerization}
& $\dot{x}_{0} = -0.12x_{0} + 1.0x_{1}$ \\
& & $\dot{x}_{1} = 0.12x_{0} - 1.0x_{1}$ \\ \midrule

\multirow{2}{*}{78} & \multirow{2}{=}{Tumour Under Nonstationary Therapy}
& $\dot{x}_{0} = -1.0x_{0}x_{1} + 2.0x_{0}$ \\
& & $\dot{x}_{1} = 1.0x_{0}x_{1} - 0.2x_{0} - 0.5x_{1} + 0.25$ \\ \midrule

\multirow{2}{*}{79} & \multirow{2}{=}{Alzheimer Choline-Leakage Hypothesis*}
& $\dot{x}_{0} = -0.007x_{0}x_{1}$ \\
& & $\dot{x}_{1} = -0.004x_{0} - 0.01x_{1} + 0.33$ \\ \midrule

\multirow{2}{*}{80} & \multirow{2}{=}{Birth--Death Process}
& $\dot{x}_{0} = 1.0 - 0.025x_{0}$ \\
& & $\dot{x}_{1} = 0.025x_{0} - 1.0$ \\ \midrule

\multirow{2}{*}{81} & \multirow{2}{=}{HIV Latency / Immune Response}
& $\dot{x}_{0} = -0.029x_{0}x_{1} + 0.134x_{0}/(x_{0}+380.0) + 0.001$ \\
& & $\dot{x}_{1} = -0.927x_{0}x_{1} + 0.07$ \\ \midrule

\multirow{2}{*}{82} & \multirow{2}{=}{Bone Marrow Invasion (Relative)}
& $\dot{x}_{0} = -0.8x_{0}^{2} - 0.9x_{0}x_{1} + 0.7x_{0}$ \\
& & $\dot{x}_{1} = -0.1x_{0}x_{1} - 0.2x_{1}^{2} + 0.1x_{1}$ \\ \midrule

\multirow{2}{*}{83} & \multirow{2}{=}{Bioterrorism Panic--Protection}
& $\dot{x}_{0} = -0.6x_{0}^{2} - 2.8x_{0}x_{1} + 6.0x_{0}$ \\
& & $\dot{x}_{1} = 1.0x_{0}x_{1}$ \\ \midrule

\multirow{2}{*}{84} & \multirow{2}{=}{Tumour Immunotherapy}
& $\dot{x}_{0} = 0.004x_{0} - 4.0x_{1}$ \\
& & $\dot{x}_{1} = 0.09x_{0}x_{1} - 0.1x_{1}$ \\ \midrule

\multirow{2}{*}{85} & \multirow{2}{=}{Cancer--Immune Cell Count}
& $\dot{x}_{0} = 0.514x_{0}$ \\
& & $\dot{x}_{1} = 10.0 - 0.02x_{1}$ \\ \midrule

\multirow{2}{*}{86} & \multirow{2}{=}{Tumour--Immune Interaction (Base)}
& $\dot{x}_{0} = -0.006544x_{0}^{2} - 1.0x_{0}x_{1} + 1.636x_{0}$ \\
& & $\dot{x}_{1} = -0.003x_{0}x_{1} + 1.131x_{0}x_{1}/(x_{0}+20.19) - 2.0x_{1} + 0.318$ \\

\bottomrule \bottomrule
\end{tabular}
\end{table}

% Required packages:
% \usepackage{booktabs}
% \usepackage{multirow}
% \usepackage{array}

% -------------------- Table 1: IDs 87--104 --------------------
\begin{table*}[!htbp]
\centering
\caption{\small 3-Dimensional ODEBase datasets.}
\label{tab:odebase_3d_part1}
\scriptsize
\setlength{\tabcolsep}{3pt}
\renewcommand{\arraystretch}{1.1}
\resizebox{\textwidth}{!}{%
\begin{tabular}{r p{3.5cm} p{10.8cm}}
\toprule
\textbf{ID} & \textbf{System} & \textbf{Equations} \\
\midrule
\multirow{3}{*}{87} & \multirow{3}{=}{Human/Mosquito ELP Epidemics} & $\dot{x}_{0}=600.0-0.411x_{0}$ \\
& & $\dot{x}_{1}=0.361x_{0}-0.184x_{1}$ \\
& & $\dot{x}_{2}=0.134x_{1}-0.345x_{2}$ \\ \midrule

\multirow{3}{*}{88} & \multirow{3}{=}{Cancer Virotherapy (Phase I)} & $\dot{x}_{0}=-1.0x_{0}x_{2}$ \\
& & $\dot{x}_{1}=1.0x_{0}x_{2}-1.0x_{1}$ \\
& & $\dot{x}_{2}=-0.02x_{0}x_{2}+1.0x_{1}-0.15x_{2}$ \\ \midrule

\multirow{3}{*}{89} & \multirow{3}{=}{Oncogenesis with Genetic Instability} & $\dot{x}_{0}=0.01-0.01x_{0}$ \\
& & $\dot{x}_{1}=0.03x_{1}$ \\
& & $\dot{x}_{2}=-0.5x_{2}^{2}+0.034x_{2}$ \\ \midrule

\multirow{3}{*}{90} & \multirow{3}{=}{p53--Mdm2 Oscillations (Model 5)} & $\dot{x}_{0}=-3.7x_{0}x_{1}+2.0x_{0}$ \\
& & $\dot{x}_{1}=-0.9x_{1}+1.1x_{2}$ \\
& & $\dot{x}_{2}=1.5x_{0}-1.1x_{2}$ \\ \midrule

\multirow{3}{*}{91} & \multirow{3}{=}{Insulin Kinetics Model A} & $\dot{x}_{0}=-0.1x_{0}x_{2}+0.2x_{1}x_{2}+0.1x_{2}$ \\
& & $\dot{x}_{1}=-0.01x_{0}+0.01+0.01/x_{2}$ \\
& & $\dot{x}_{2}=-0.1x_{1}x_{2}+0.257x_{1}-0.1x_{2}^{2}+0.331x_{2}-0.3187$ \\ \midrule

\multirow{3}{*}{92} & \multirow{3}{=}{p53--Mdm2 Oscillations (Model 1)} & $\dot{x}_{0}=-3.2x_{0}x_{1}+0.3$ \\
& & $\dot{x}_{1}=-0.1x_{1}+0.1x_{2}$ \\
& & $\dot{x}_{2}=0.4x_{0}-0.1x_{2}$ \\ \midrule

\multirow{3}{*}{93} & \multirow{3}{=}{Colon Crypt Cell Cycle (v1)} & $\dot{x}_{0}=-0.002207x_{0}^{2}-0.002207x_{0}x_{1}-0.002207x_{0}x_{2}+0.1648x_{0}$ \\
& & $\dot{x}_{1}=-0.01312x_{0}^{2}-0.0216x_{0}x_{1}-0.01312x_{0}x_{2}+1.574x_{0}-0.008477x_{1}^{2}-0.008477x_{1}x_{2}+0.5972x_{1}$ \\
& & $\dot{x}_{2}=-0.04052x_{0}x_{1}-0.04052x_{1}^{2}-0.04052x_{1}x_{2}+4.863x_{1}-1.101x_{2}$ \\ \midrule

\multirow{3}{*}{94} & \multirow{3}{=}{Prophage Induction} & $\dot{x}_{0}=-0.99x_{0}^{2}/(x_{0}+x_{1})-1.0x_{0}x_{1}/(x_{0}+x_{1})+0.99x_{0}$ \\
& & $\dot{x}_{1}=-0.99x_{0}x_{1}/(x_{0}+x_{1})-1.0x_{1}^{2}/(x_{0}+x_{1})+1.0x_{1}$ \\
& & $\dot{x}_{2}=-0.001x_{2}$ \\ \midrule

\multirow{3}{*}{95} & \multirow{3}{=}{Tumour--Immune with IL-2} & $\dot{x}_{0}=-1.0x_{0}x_{1}/(x_{0}+1.0)+0.18x_{0}$ \\
& & $\dot{x}_{1}=0.05x_{0}+0.124x_{1}x_{2}/(x_{2}+20.0)-0.03x_{1}$ \\
& & $\dot{x}_{2}=5.0x_{0}x_{1}/(x_{0}+10.0)-10.0x_{2}$ \\ \midrule

\multirow{3}{*}{96} & \multirow{3}{=}{Cytotoxic/Helper T Cell--Tumour Interaction} & $\dot{x}_{0}=-10.0x_{0}^{2}-2.075x_{0}x_{2}+10.0x_{0}$ \\
& & $\dot{x}_{1}=0.19x_{0}x_{1}/(x_{0}^{2}+0.0016)-1.0x_{1}+0.5$ \\
& & $\dot{x}_{2}=-2.075x_{0}x_{2}+1.0x_{1}x_{2}-1.0x_{2}+2.0$ \\ \midrule

\multirow{3}{*}{97} & \multirow{3}{=}{Zombie (SIZRC) Epidemic} & $\dot{x}_{0}=-0.009x_{0}x_{1}+0.05$ \\
& & $\dot{x}_{1}=0.004x_{0}x_{1}$ \\
& & $\dot{x}_{2}=0.005x_{0}x_{1}$ \\ \midrule

\multirow{3}{*}{98} & \multirow{3}{=}{Circadian Oscillations / NF-$\kappa$B Signalling} & $\dot{x}_{0}=-954.5x_{0}x_{2}/(x_{0}+0.029)-0.007x_{0}/x_{2}+0.007/x_{2}$ \\
& & $\dot{x}_{1}=1.0x_{0}^{2}-1.0x_{1}$ \\
& & $\dot{x}_{2}=0.035x_{0}+1.0x_{1}-0.035$ \\ \midrule

\multirow{3}{*}{99} & \multirow{3}{=}{Glioma--Immune Interaction} & $\dot{x}_{0}=-0.482x_{0}^{2}-0.07x_{0}x_{1}/(x_{0}+0.903)-2.745x_{0}x_{2}/(x_{0}+0.903)+0.482x_{0}$ \\
& & $\dot{x}_{1}=-0.019x_{0}x_{1}/(x_{0}+0.031)-0.331x_{1}^{2}+0.331x_{1}$ \\
& & $\dot{x}_{2}=0.124x_{0}x_{2}/(x_{0}+2.874)-0.017x_{0}x_{2}/(x_{0}+0.379)-0.007x_{2}$ \\ \midrule

\multirow{3}{*}{100} & \multirow{3}{=}{Hepatitis C Infection Dynamics} & $\dot{x}_{0}=-0.002x_{0}+1.065e{+}4x_{0}/(x_{0}+x_{1})+0.118x_{1}$ \\
& & $\dot{x}_{1}=-0.118x_{1}+342.5x_{1}/(x_{0}+x_{1})$ \\
& & $\dot{x}_{2}=204.0x_{1}-17.91x_{2}$ \\ \midrule

\multirow{3}{*}{101} & \multirow{3}{=}{Tumour--Normal Cell Progression} & $\dot{x}_{0}=-0.931x_{0}x_{1}-0.138x_{0}x_{2}+0.431x_{0}$ \\
& & $\dot{x}_{1}=1.189x_{0}x_{1}-0.1772x_{1}^{2}-0.147x_{1}x_{2}+0.443x_{1}$ \\
& & $\dot{x}_{2}=-0.813x_{0}x_{2}+0.271x_{0}x_{2}/(x_{0}+0.813)-0.363x_{1}x_{2}+0.783x_{1}x_{2}/(x_{1}+0.862)-0.57x_{2}+0.7$ \\ \midrule

\multirow{3}{*}{102} & \multirow{3}{=}{L-Dopa Pharmacokinetics} & $\dot{x}_{0}=-2.11x_{0}$ \\
& & $\dot{x}_{1}=0.889x_{0}-1.659x_{1}$ \\
& & $\dot{x}_{2}=0.4199x_{1}-0.06122x_{2}$ \\ \midrule

\multirow{3}{*}{103} & \multirow{3}{=}{p53--Mdm2 Oscillations (Model 4)} & $\dot{x}_{0}=0.9-1.7x_{1}$ \\
& & $\dot{x}_{1}=-0.8x_{1}+0.8x_{2}$ \\
& & $\dot{x}_{2}=1.1x_{0}-0.8x_{2}$ \\ \midrule

\multirow{3}{*}{104} & \multirow{3}{=}{Tumour Dormancy Equilibrium} & $\dot{x}_{0}=-1.125x_{0}^{2}-0.3x_{0}x_{1}+0.9x_{0}+10.0$ \\
& & $\dot{x}_{1}=0.1x_{1}x_{2}-0.02x_{1}$ \\
& & $\dot{x}_{2}=-0.1x_{1}x_{2}-1.143x_{2}^{2}+0.77x_{2}$ \\
\bottomrule \bottomrule
\end{tabular}%
}
\end{table*}

% -------------------- Table 2: IDs 105--122 --------------------
\begin{table*}[!htbp]
\centering
\caption{\small 3-Dimensional ODEBase datasets (contd.).}
\label{tab:odebase_3d_part2}
\scriptsize
\setlength{\tabcolsep}{3pt}
\renewcommand{\arraystretch}{1.1}
\resizebox{\textwidth}{!}{%
\begin{tabular}{r p{3.5cm} p{10.8cm}}
\toprule
\textbf{ID} & \textbf{System} & \textbf{Equations} \\
\midrule
\multirow{3}{*}{105} & \multirow{3}{=}{Tumour--Immune Noise-assisted Interactions} & $\dot{x}_{0}=-1.0x_{0}^{2}-1.0x_{0}x_{1}+1.748x_{0}+2.73x_{2}$ \\
& & $\dot{x}_{1}=1.0x_{1}x_{2}-0.05x_{1}$ \\
& & $\dot{x}_{2}=1.126x_{0}-15.89x_{2}$ \\ \midrule

\multirow{3}{*}{106} & \multirow{3}{=}{Goldbeter Embryonic Cell Cycle} & $\dot{x}_{0}=-0.25x_{0}-0.25x_{2}+0.25$ \\
& & $\dot{x}_{1}=-6.0x_{0}x_{1}/(-2.0x_{0}x_{1}+2.002x_{0}-1.0x_{1}+1.001)+6.0x_{0}/(-2.0x_{0}x_{1}+2.002x_{0}-1.0x_{1}+1.001)-1.5x_{1}/(x_{1}+0.001)$ \\
& & $\dot{x}_{2}=-1.0x_{1}x_{2}/(1.001-x_{2})+1.0x_{1}/(1.001-x_{2})-0.7x_{2}/(x_{2}+0.001)$ \\ \midrule

\multirow{3}{*}{107} & \multirow{3}{=}{Helper T Cells in Tumour Immune System} & $\dot{x}_{0}=-0.003272x_{0}^{2}-1.0x_{0}x_{1}+1.636x_{0}$ \\
& & $\dot{x}_{1}=0.04x_{0}x_{1}+0.01x_{1}x_{2}-0.374x_{1}$ \\
& & $\dot{x}_{2}=0.002x_{0}x_{2}-0.055x_{2}+0.38$ \\ \midrule

\multirow{3}{*}{108} & \multirow{3}{=}{Cholesterol Biosynthesis (SREBP2)} & $\dot{x}_{0}=-0.001x_{0}$ \\
& & $\dot{x}_{1}=1.0x_{0}-0.002x_{1}$ \\
& & $\dot{x}_{2}=0.462x_{1}-0.004x_{2}$ \\ \midrule

\multirow{3}{*}{109} & \multirow{3}{=}{HIV/CD4 T-cell Interaction} & $\dot{x}_{0}=-0.1x_{0}^{2}x_{2}-0.1x_{0}x_{1}x_{2}+0.8x_{0}x_{2}-0.1x_{0}$ \\
& & $\dot{x}_{1}=-0.1x_{0}x_{1}x_{2}+0.2x_{0}x_{2}-0.1x_{1}^{2}x_{2}+1.0x_{1}x_{2}-0.2x_{1}$ \\
& & $\dot{x}_{2}=1.0x_{1}-0.5x_{2}$ \\ \midrule

\multirow{3}{*}{110} & \multirow{3}{=}{Cyclin-dependent Kinase Oscillations} & $\dot{x}_{0}=-0.25x_{0}x_{2}/(x_{0}+0.001)-0.046x_{0}+0.06$ \\
& & $\dot{x}_{1}=-4.0x_{0}x_{1}/(-x_{0}x_{1}+1.002x_{0}-0.5x_{1}+0.501)+4.0x_{0}/(-x_{0}x_{1}+1.002x_{0}-0.5x_{1}+0.501)-2.0x_{1}/(x_{1}+0.002)$ \\
& & $\dot{x}_{2}=-1.0x_{1}x_{2}/(1.01-x_{2})+1.0x_{1}/(1.01-x_{2})-0.7x_{2}/(x_{2}+0.01)$ \\ \midrule

\multirow{3}{*}{111} & \multirow{3}{=}{Tumour--Normal--Vitamins (TNVM)} & $\dot{x}_{0}=-0.982x_{0}x_{1}+0.222x_{0}x_{2}+0.431x_{0}$ \\
& & $\dot{x}_{1}=0.229x_{0}x_{1}-0.1772x_{1}^{2}-0.497x_{1}x_{2}+0.443x_{1}$ \\
& & $\dot{x}_{2}=0.898-0.961x_{2}$ \\ \midrule

\multirow{3}{*}{112} & \multirow{3}{=}{Colon Crypt Cell Cycle (v0)} & $\dot{x}_{0}=0$ \\
& & $\dot{x}_{1}=1.0x_{0}^{2}/(x_{0}+2.924)+0.218x_{0}-0.024x_{1}$ \\
& & $\dot{x}_{2}=1.0x_{1}^{2}/(x_{1}+29.24)+0.547x_{1}-1.83x_{2}$ \\ \midrule

\multirow{3}{*}{113} & \multirow{3}{=}{Circadian Rhythms (Neurospora)} & $\dot{x}_{0}=-0.505x_{0}/(x_{0}+0.5)+1.6/(x_{2}^{4.0}+1.0)$ \\
& & $\dot{x}_{1}=0.5x_{0}-0.5x_{1}-1.4x_{1}/(x_{1}+0.13)+0.6x_{2}$ \\
& & $\dot{x}_{2}=0.5x_{1}-0.6x_{2}$ \\ \midrule

\multirow{3}{*}{114} & \multirow{3}{=}{Toxicant--Immune System Dynamics} & $\dot{x}_{0}=-0.2x_{0}^{2}-0.05x_{0}x_{1}+0.9x_{0}$ \\
& & $\dot{x}_{1}=0.295x_{0}x_{1}-0.8x_{1}+0.04$ \\
& & $\dot{x}_{2}=2.4x_{0}-0.1x_{2}$ \\ \midrule

\multirow{3}{*}{115} & \multirow{3}{=}{Tumour--CD4+--Cytokine Interactions} & $\dot{x}_{0}=-3.0e{-}5x_{0}^{2}-0.1x_{0}x_{2}/(x_{0}+1.0)+0.03x_{0}$ \\
& & $\dot{x}_{1}=0.02x_{0}x_{1}/(x_{0}+10.0)-0.02x_{1}+10.0$ \\
& & $\dot{x}_{2}=0.1x_{0}x_{1}/(x_{0}+0.1)-47.0x_{2}$ \\ \midrule

\multirow{3}{*}{116} & \multirow{3}{=}{Proteasome Dynamics (Parkinson's)} & $\dot{x}_{0}=-1.0x_{0}x_{1}+25.0/(x_{1}+1.0)$ \\
& & $\dot{x}_{1}=-1.0x_{0}x_{1}-x_{1}+1.0x_{2}+1.0$ \\
& & $\dot{x}_{2}=1.0x_{0}x_{1}-1.0x_{2}$ \\ \midrule

\multirow{3}{*}{117} & \multirow{3}{=}{Weight Cycling Dynamics} & $\dot{x}_{0}=-0.1x_{0}/(x_{0}+0.2)+0.1x_{1}$ \\
& & $\dot{x}_{1}=-1.5x_{1}x_{2}/(x_{1}+0.01)-1.0x_{1}/(1.01-x_{1})+1.0/(1.01-x_{1})$ \\
& & $\dot{x}_{2}=-6.0x_{0}x_{2}/(1.01-x_{2})+6.0x_{0}/(1.01-x_{2})-2.5x_{2}/(x_{2}+0.01)$ \\ \midrule

\multirow{3}{*}{118} & \multirow{3}{=}{Stem Cell + Transit Amplifying Cells} & $\dot{x}_{0}=0.004x_{0}+0.0096x_{1}/(0.01x_{0}^{1.0}+1.0)$ \\
& & $\dot{x}_{1}=0.006x_{0}-0.004x_{1}$ \\
& & $\dot{x}_{2}=0.024x_{1}-0.0096x_{1}/(0.01x_{0}^{1.0}+1.0)-0.003x_{2}$ \\ \midrule

\multirow{3}{*}{119} & \multirow{3}{=}{Tumour--Immune Immunotherapy} & $\dot{x}_{0}=0.044x_{0}x_{2}/(x_{2}+0.02)-0.038x_{0}+1.009x_{1}$ \\
& & $\dot{x}_{1}=-0.018x_{0}-0.123x_{1}^{2}+0.123x_{1}$ \\
& & $\dot{x}_{2}=0.9x_{0}-1.8x_{2}$ \\ \midrule

\multirow{3}{*}{120} & \multirow{3}{=}{Tumour Growth Model} & $\dot{x}_{0}=-17.86x_{0}x_{2}^{2}+0.05x_{0}x_{2}/(x_{0}+x_{1}+x_{2}+1.0)-0.1x_{0}+0.625x_{2}+0.01$ \\
& & $\dot{x}_{1}=-x_{1}+2.0x_{1}/(x_{0}+x_{1}+x_{2}+1.0)$ \\
& & $\dot{x}_{2}=-25.0x_{0}x_{2}^{2}-x_{2}+4.0x_{2}/(x_{0}+x_{1}+x_{2}+1.0)$ \\ \midrule

\multirow{3}{*}{121} & \multirow{3}{=}{Oncolytic M1 Virus--SNT Model} & $\dot{x}_{0}=-0.2x_{0}x_{1}-0.5x_{0}x_{2}-0.02x_{0}+0.02$ \\
& & $\dot{x}_{1}=0.16x_{0}x_{1}-0.03x_{1}$ \\
& & $\dot{x}_{2}=0.4x_{0}x_{2}-0.028x_{2}$ \\ \midrule

\multirow{3}{*}{122} & \multirow{3}{=}{CAR T-cell Therapy in ALL} & $\dot{x}_{0}=-0.07143x_{0}$ \\
& & $\dot{x}_{1}=0.033x_{1}$ \\
& & $\dot{x}_{2}=-0.01667x_{2}$ \\
\bottomrule \bottomrule
\end{tabular}%
}
\end{table*}

\vspace{-1.75em}
\subsection{Evaluation Metrics}
\label{app:metrics}

\vspace{-0.5em}
\paragraph{Data Fidelity.}
We evaluate data fidelity using Normalized Mean Squared Error (NMSE), which measures the relative discrepancy between predicted and ground-truth dynamics, normalized by each system's scale. Given predictions $\hat{y}$ and ground-truth values $y$, we compute:
\begin{align}
\vspace{-1.25em}
\mathrm{NMSE}(\hat{y}, y) =
\frac{\lVert \hat{y} - y \rVert_2^2}{\lVert y \rVert_2^2+\varepsilon},
\vspace{-1.25em}
\end{align}
where $\varepsilon=10^{-10}$ is used for numerical stability. NMSE is nonnegative and ranges from $0$ to $\infty$, with $\mathrm{NMSE}=0$ indicating exact agreement between the predicted and true dynamics. Values closer to zero indicate higher predictive fidelity, while larger values indicate increasing deviation from the ground-truth dynamics. We report NMSE across three evaluation regimes that probe progressively stronger forms of distribution shift. In the reconstruction setting, the discovered model is numerically integrated from the training initial condition $\mathrm{IC}_0$ using the LSODA solver and compared with the ground-truth trajectory over $t\in[0,1]$ sampled at 100 uniformly spaced points. In the generalization setting, the same learned dynamics are evaluated from a held-out initial condition $\mathrm{IC}_1$ over the same time horizon, isolating robustness to unseen initial conditions. In the extrapolation setting, the model is rolled out from $\mathrm{IC}_0$ over the longer horizon $t\in(1,10]$ using 150 sampled points to test long-horizon stability beyond the training window. Together, these regimes assess whether the discovered equations not only fit the observed trajectory but also reproduce the underlying dynamics under new initial conditions and extended temporal extrapolation.

\vspace{-0.75em}
\paragraph{Expression Complexity.}
Expression complexity is measured as the size of the symbolic expression’s tree representation. Each discovered equation is encoded as a rooted tree, where leaves correspond to variables and constants, and internal nodes correspond to unary or binary operators (e.g., $+, -, \times, \sin, \exp$). The complexity is defined as the total number of nodes in the tree:
\[
\mathrm{Complexity}(\mathbf{f}) = |T_\mathbf{f}|,
\]
where $T_\mathbf{f}$ denotes the expression tree of equation $\mathbf{f}$. This quantity is computed recursively by assigning a unit cost to each node in the tree and summing over all sub-expressions. Specifically, each operator contributes one unit to the complexity, and each variable or constant also contributes one unit. Consequently, expressions with greater depth or a larger number of composed operations yield higher complexity values. This provides a structural measure of model size and is commonly used as a parsimony objective in symbolic regression to discourage unnecessarily complex expressions.

\vspace{-0.25em}

\begin{figure}[!htbp]
    \centering
    \includegraphics[width=0.75\linewidth]{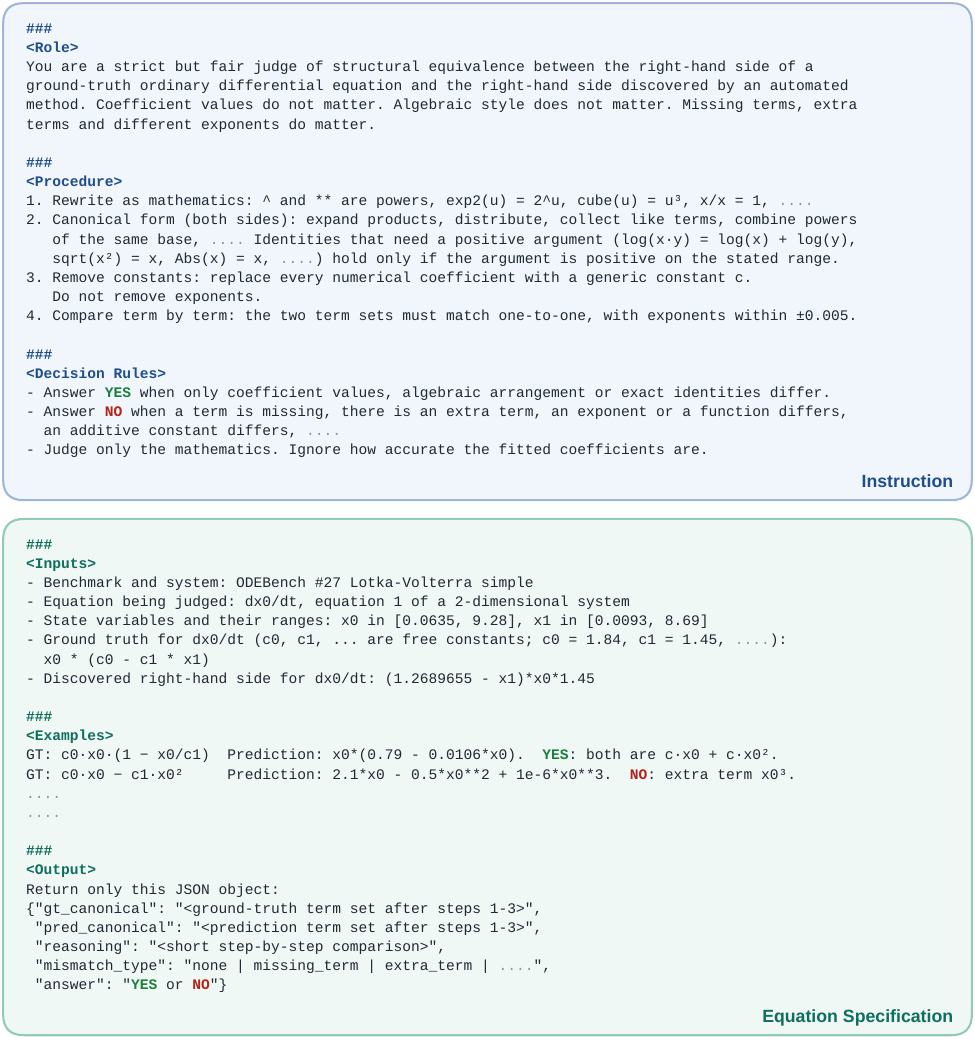}
    \vspace{-0.25em}
    \caption{\small Prompt for LLM-judge to calculate the symbolic accuracy.}
    \label{fig:judge}
\end{figure}
\vspace{-0.5em}

\paragraph{Symbolic Accuracy.} Following LLM-SRBench~\citep{shojaeellm}, we use an LLM judge to decide whether each discovered ODE has the same structure as the ground truth. Exact string match and normalized tree edit distance penalize algebraically equivalent rewrites, so we instead use {\texttt{Claude-Opus-4.6} (temperature 0) as the judge. Each method's output is first mapped to a common $x_0, x_1, \ldots$ variable naming by a method-specific normalizer and given to the judge with its fitted constants intact, together with the ground-truth right-hand side and the range each state variable takes on the benchmark trajectories. The judge follows a fixed procedure: it rewrites both sides in canonical form (expanding products, collecting like terms, and applying identities such as $\log(xy)=\log x+\log y$ only where they hold on the stated range), replaces every numerical coefficient with a generic constant while keeping exponents, and compares the resulting sets of terms. It answers yes only if the terms match one-to-one with exponents agreeing within $\pm 0.005$; a missing or extra term, a different function, or an added or dropped constant term is a mismatch. Each state dimension is judged independently, and a system's score is the fraction of its dimensions recovered (partial credit; e.g., $0.67$ if 2 of 3 are correct), with systems that have no saved equation scoring $0$. Further, to validate the LLM judge, we compared its verdicts against a manual review, where the LLM judge agreed with manual review on 98.4\% of the assessed comparisons.
\vspace{-.25em}
\section{Implementation Details}
\label{app:implementation}
% \vspace{-0.25em}
\subsection{Computational Resources}
\vspace{-0.25em}

All experiments run on a server with two Intel Xeon Gold 5220R processors ($48$ physical cores) and $502$ GB RAM. Non-LLM regression baselines run on this server with a 12-hour timeout per dataset. Furthermore, we used four NVIDIA RTX $8000$ GPUs ($48$ GB each) for setting up the inference for neural network-based or transformer-based methods. For LLM-based methods, we used hosted APIs, for example, GPT through OpenAI and \qwen through Modal and AWS.
\vspace{-0.5em}

\subsection{Baselines}
\vspace{-0.25em}

\label{app:baselines}
We compare our method against established baselines spanning sparse regression, evolutionary symbolic regression, neural sequence modeling, and active data acquisition. For all static methods, we adopt a unified experimental protocol in which each model is trained on the provided set of $200$ data points from the default initial condition. For active regression methods, each method follows its own data acquisition process to acquire equivalent total samples and is ultimately evaluated on the same set of data points. We follow MDBench's implementations\footnote{\url{https://github.com/gryaklab/mdbench}} ~for all baselines but expand the candidate operator sets beyond those used in the original benchmark. MDBench restricts its operator pools to a subset that appears in its equation catalog, introducing a systematic bias as the search spaces match the benchmark's equations. All baselines except APPS-ODE~\cite{jiang2025active}, ODEFormer~\cite{dascoli2024odeformer}, and Operon~\cite{burlacu2020operon} use the expanded set:
\vspace{-0.25em}
\begin{itemize}[leftmargin=*]
\setlength\itemsep{-0.1em}
    \item {\textbf{Unary Operators}}: $\sin, \cos, \tan, \exp, \log, \sqrt{\cdot}, |\cdot|, \tanh, \sinh, \cosh, \cdot^2, \cdot^3, \cdot^{-1}, -\cdot, \sqrt[3]{\cdot}, \log_2, \log_{10}, 2^{\cdot}$ 
    \item {\textbf{Binary Operators}}: \{$+, -, \times, \div, \hat{\;}\}$ 
\end{itemize}

\begin{table*}[!htbp]
\centering
\caption{\small Hyperparameter settings for baselines.}
\vspace{-0.75em}
\fontsize{6}{7}\selectfont
\begin{tabular}{lll}
\toprule
\textbf{Model} & \textbf{Hyperparameter} & \textbf{Values} \\
\midrule
\multirow{6}{*}{SINDy} & threshold & \texttt{np.logspace(-7,0,16)} \\
 & \multirow{2}{*}{basis functions} & [polynomial], [polynomial, sin, cos, tan, exp, log, sqrt, abs, \\
 & & tanh, sinh, cosh, square, cube, inv, neg, cbrt, log2, log10, exp2] \\
 & polynomial order & 1, 2, 3, 4 \\
 & alpha & $10^{-5}, 10^{-4}$ \\
 & optimizer & STLSQ \\
 & max iterations & 200 \\
\midrule

\multirow{6}{*}{PySR} & \#iterations, \#cycles per iteration & 100, 1000 \\
 & \#populations, population size & 20, 100 \\
 & max size, max depth & 40, 20 \\
 & binary operators & [+, -, $\times$, /, \^ ] \\
 & \multirow{2}{*}{unary operators} & sin, cos, tan, exp, log, sqrt, abs, tanh, sinh \\
 & & cosh, square, cube, inv, neg, cbrt, log$_2$, log$_{10}$, exp$^2$ \\
\midrule

\multirow{3}{*}{End2End} & max input points & 200 \\
 & \#trees to refine & 10 \\
 & rescale & True \\
\midrule

\multirow{2}{*}{ODEFormer} & beam temperature & 0.05, 0.1, 0.2, 0.3, 0.5 \\
 & beam size & 50 \\
\midrule

\multirow{8}{*}{Operon} & \multirow{2}{*}{symbols} & add, sub, mul, div, aq, pow, abs, cbrt, cos \\ 
& & cosh, exp, log, sin, sinh, sqrt, tan, tanh, square \\
 & brood size & 10 \\
 & max depth, max length & 10, 50 \\
 & pool size, population size & 1000, 1000 \\
 & tournament size & 3 \\
 & mutation probability & 0.25 \\
 & optimizer & LM (Levenberg--Marquardt) \\
\midrule

\multirow{5}{*}{APPS-ODE} & \#epochs & 50 \\
 & reward threshold & $1/(1+10^{-6})$ \\
 & grammar max length & 10 \\
 & top-K size & 10 \\
 & function set & [add, sub, mul, div, sin, exp, poly, const] \\

\bottomrule
\bottomrule
\end{tabular}
\end{table*}
\vspace{-0.5em}
\subsubsection*{Passive Symbolic Discovery}
\noindent \textbf{SINDy}~\citep{brunton2016discovering} uses the \href{https://github.com/dynamicslab/pysindy}{\texttt{PySINDy}} implementation with STLSQ (sequential thresholding least squares) optimizer. The sparsity threshold is swept over a log-uniform grid from $10^{-7}$ to $1$ (16 values), and $\ell_2$ regularization strength $\alpha \in \{10^{-5}, 10^{-4}\}$. We extend the basis library beyond MDBench's polynomial-only setting to also search over the full expanded nonlinear library above, combined with polynomials up to degree 4.
% \vspace{-0.25em}

\noindent \textbf{Operon}~\citep{burlacu2020operon} uses the operator set for \href{https://github.com/heal-research/operon}{\texttt{operon}} on top of implementation given in~\cite{bideh2026mdbench}. Model selection uses minimum description length on the Pareto front.
% \vspace{-0.25em}

\noindent \textbf{PySR}~\citep{cranmer2023interpretable} uses MDBench's implementation of \href{https://github.com/MilesCranmer/PySR}{\texttt{PySR}} with multi-population evolutionary search over expression trees, with an expanded operator set. Nested constraints prevent pathological compositions (e.g., $\exp(\exp(\cdot))$), and the complexity-fitness tradeoff is scored via a Pareto fitness metric.
% \vspace{-0.25em}

\noindent \textbf{End2End (E2E)}~\citep{kamienny2022end} uses the pretrained Transformer from \href{https://github.com/facebookresearch/symbolicregression}{\texttt{symbolicregression}}, following the implementation in~\cite{bideh2026mdbench}. We cap the input at 200 points, enable input rescaling, and refine the top 10 predicted expression trees using the model's constant-refinement procedure.
% \vspace{-0.25em}

\noindent \textbf{ODEFormer}~\citep{dascoli2024odeformer} is run in inference mode using the pretrained checkpoint, without fine-tuning\footnote{\url{https://github.com/sdascoli/odeformer}}. As a fixed pretrained model, its internal operator vocabulary cannot be modified.
% \vspace{-0.25em}

\subsubsection*{LLM-guided Symbolic Discovery}
% \vspace{-0.25em}
\noindent \textbf{LLM-only} follows the protocol from NewtonBench\footnote{\url{https://github.com/HKUST-KnowComp/NewtonBench}}, prompting \gpt iteratively with feedback from the previous round to guide the future equations. The LLM is implemented using a temperature $1.0$ to maximize generative diversity.
% \vspace{-0.25em}

\noindent \textbf{LLM-ODE}~\citep{bideh2026llm} uses the original implementation from {\href{https://github.com/gryaklab/llm-ode}{\texttt{llm-ode}}} with LLM-proposed structural templates and numerical parameter optimization. The multi-island evolutionary strategy and dynamic experience buffer are used as in the original implementation.
% \vspace{-0.25em}

\subsubsection*{Active Symbolic Discovery}
% \vspace{-0.25em}
\noindent \textbf{Query-By-Committee (QBC)}~\citep{haut2022active, haut2023active, haut2024active} adapts \texttt{QBC} active learning to ODE initial condition selection, using \texttt{PySR} operating on gradient-matched data, where the committee is drawn from the Pareto front. \texttt{QBC} also uses the same oracle as \modelname to query new initial conditions.
% \vspace{-0.25em}

\noindent \textbf{Bayesian Optimization (BO)} combines \texttt{PySR} with a Gaussian process surrogate over initial condition space. Each iteration, the GP predicts the expected value of the NMSE of the \texttt{PySR} fit from a candidate IC and selects the next query using Expected Improvement (EI) from a pool of 256 uniformly sampled ICs. After querying the oracle, \texttt{PySR} is refit on all accumulated data. The GP is updated with the observed NMSE as the reward.
% \vspace{-0.25em}

\noindent \textbf{APPS-ODE}~\citep{jiang2025active} implementation uses the original grammar-RL pipeline with grammar function set \{$+, -, \times, \div, \sin, \exp, \mathrm{poly}, \mathrm{const}$\}. This grammar-constrained setting reduces the symbolic search space and therefore provides a favorable inductive bias when the target dynamics lie within or near the supported grammar. We train APPS-ODE for $50$ policy-gradient epochs, querying initial conditions with $100$ observations as given in their official repository\footnote{\url{https://github.com/jiangnanhugo/APPS-ODE}}. Coefficients are optimized with BFGS; the reward signal is inverse NMSE. For both datasets, trajectories are queried from fresh initial conditions each epoch using the same oracle used for \modelname.

% \vspace{-.5em}

% \noindent \textbf{LLM-only} We follow NewtonBench~\cite{zheng2026newtonbench}, run \gpt with feedback from the previous iteration\footnote{https://github.com/HKUST-KnowComp/NewtonBench}.

% \noindent \textbf{LLM-ODE}~\citep{bideh2026llm} formulates symbolic regression as program synthesis, where candidate equations are represented as executable programs (e.g., Python functions)\footnote{\url{https://github.com/gryaklab/llm-ode}}. A large language model generates structural templates with free parameters, which are subsequently optimized using numerical methods. The approach leverages a dynamic experience buffer and a multi-island evolutionary strategy to maintain diversity and reuse high-quality candidates. By decoupling discrete structure generation from continuous parameter fitting, LLM-ODE efficiently explores the combinatorial search space while mitigating local optima. 

% \noindent \textbf{APPS-ODE}~\citep{jiang2025active} introduces an active learning framework for symbolic ODE discovery\footnote{\url{https://github.com/jiangnanhugo/APPS-ODE}}. It combines a Transformer-based decoder that generates candidate equations via grammar-guided tree construction with a phase portrait–based data acquisition strategy. Rather than sampling individual initial conditions, the method selects informative regions of the state space based on trajectory divergence among candidate models. The decoder is trained using REINFORCE, with rewards inversely proportional to trajectory NMSE. Model coefficients are optimized using BFGS. 
\subsection{LLM-ACES}
% \vspace{-0.25em}
\label{app:llmaces}

\modelname follows a closed-loop  \textit{hypothesis–evaluation–acquisition} framework for discovery of dynamical systems. Instead of directly generating full symbolic equations, the language model operates at the level of \emph{operator priors}, which define structured hypothesis spaces. These priors are instantiated, evaluated, and iteratively refined through feedback and active data acquisition.
% \vspace{-0.5em}
\paragraph{Hypothesis Exploration.}
\label{app:hypo_cons}
Figure~\ref{fig:ode-example} illustrates how \modelname constructs and iteratively refines the hypothesis space for symbolic ODE discovery. At each iteration, the LLM is conditioned on a structured prompt that defines its role and enforces a standardized interface for prior generation. The prompt (Figure~\ref{fig:ode-example}) consists of four key components: (i) Specification, which formalizes the task, system dynamics, and evaluation interface (including the objective defined via negative NMSE and the function signature for candidate equations); (ii) Current Best Equations, which provide the current top-performing solutions along with their scores; (iii) Failure Cases, which capture poorly performing hypotheses to discourage ineffective operator patterns; and (iv) Existing Operator Priors, which maintain a memory of previously discovered operator priors in an experience buffer. Conditioned on these inputs, the LLM generates new priors, defined as operator-level templates that specify the functional structure of candidate equations. The Output Format enforces that these operator priors remain high-level and structured, requiring explicit operator sets per dimension, constrained cardinality, and standardized naming, while explicitly prohibiting direct equation synthesis. This design ensures that the LLM contributes abstract inductive biases instead of explicit solutions, enabling controlled expansion of the hypothesis space across iterations rather than collapsing to a narrow set of forms. The prompt shown in Figure~\ref{fig:ex_diversity} promotes diversity in hypothesis exploration. The model is explicitly instructed to generate operators from previously unused mathematical families, encouraging exploration of new functional regimes while avoiding redundancy with existing operator priors. Each prior defines a constrained symbolic template over a restricted operator vocabulary, including unary transformations such as trigonometric, exponential, polynomial, and logarithmic functions, as well as standard binary operators. To ensure validity and numerical stability, structural constraints are enforced during parsing. These include requiring the presence of basic arithmetic operators (such as addition and multiplication), restricting exponentiation ranges, and filtering out syntactically invalid or duplicate proposals. Together, these constraints produce structured and tractable symbolic search spaces while preventing degenerate or unstable formulations. Success cases guide exploitation, while failure cases and existing operator sets jointly shape exploration by incorporating both negative feedback and historical priors.

\vspace{-1em}
\begin{figure}[!htbp]
    \centering
    \vspace{0.5em}
    \includegraphics[width=0.95\linewidth]{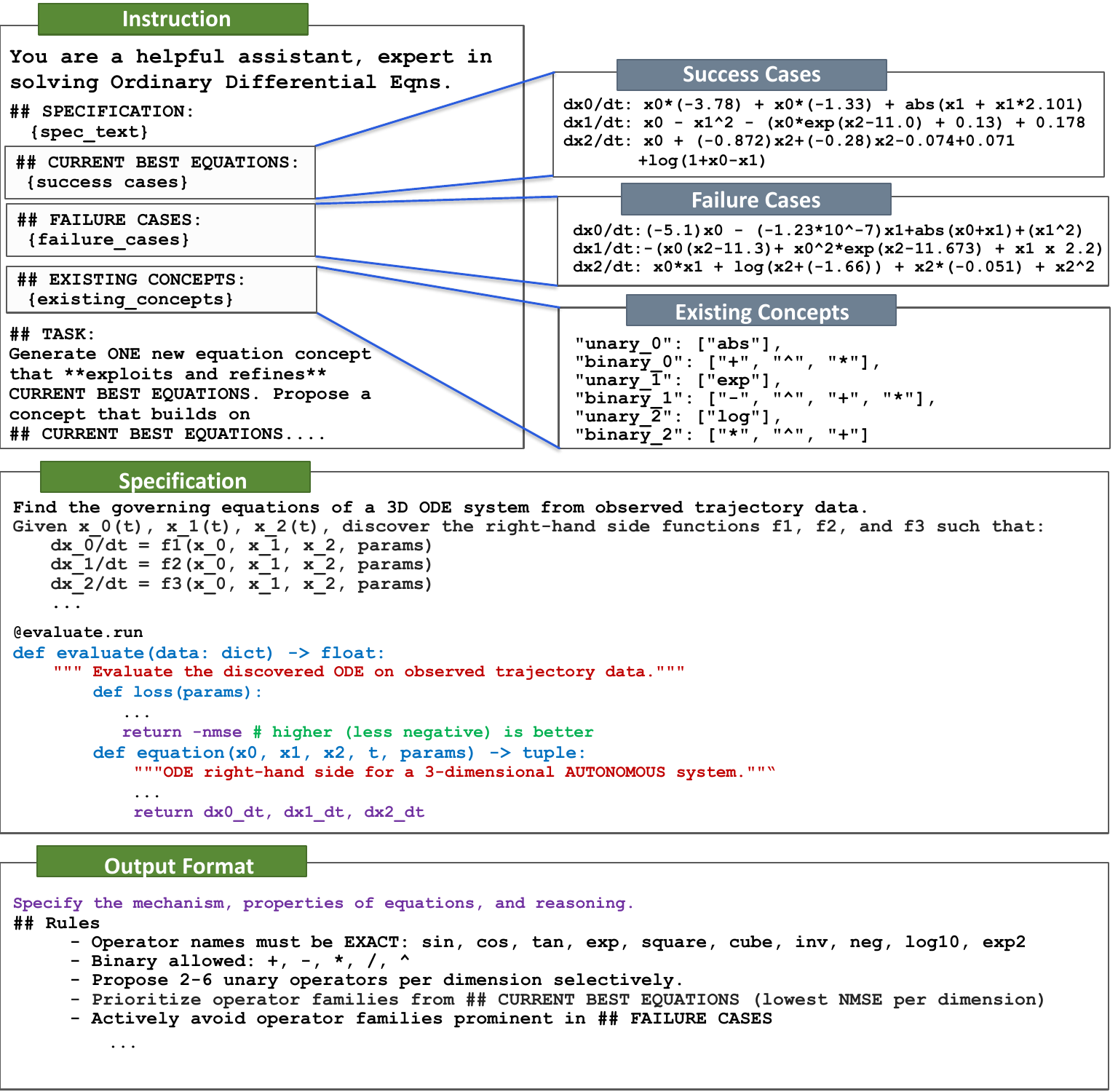}
    % \vspace{-1.5em}
    \caption{\small \textbf{Prompt example of \AlgName for 3D ODE discovery.} The prompt includes instructions, a task specification describing the objective, system characteristics, and data samples, as well as the target function and required structured output format. It also shows intermediate elements such as the current best equations, existing concepts, and failure cases. The prompts refer to operator priors as concepts.}
    \label{fig:ode-example}
    % \vspace{-2em}
\end{figure}

\paragraph{Hypothesis Generation.} 
\label{app:hypo_gen}
Given a set of validated operator priors, \modelname instantiates each prior as a constrained symbolic search space and performs dimension-wise regression using PySR~\citep{cranmer2023interpretable}. For a $d$-dimensional system and $K$ operator priors per iteration, this results in $d \times K$ independent regression problems, each restricted to the operator set defined by its corresponding prior. To ensure fair comparison across priors, all regression tasks are executed under a fixed computational budget, providing uniform search capacity. For symbolic regression, each operator prior set is evaluated using \texttt{PySR} with 20 iterations and 15 populations. Candidate equations are assessed using normalized mean squared error (NMSE) on validation splits. We select the best equation for each state dimension independently by retaining the one with the best validation performance. This decoupled selection strategy lets \modelname preserve partially correct components of the system, avoiding the failure mode in which errors in other dimensions discard accurate sub-dynamics. \modelname is instantiated with \gpt and \qwen, with \qwen served via Modal and DeepInfra, and GPT accessed through the OpenAI API. \modelname runs \texttt{PySR} for 20 iterations for each operator set generated by the LLM. Each run consists of 10 rounds of LLM calls with up to three operator sets generated per iteration at a sampling temperature of 0.8, balancing exploration of novel functional forms with exploitation of prior structural knowledge.

\begin{figure}[!htbp]
    \vspace{-0.75em}
    \centering
    \includegraphics[width=0.9\linewidth]{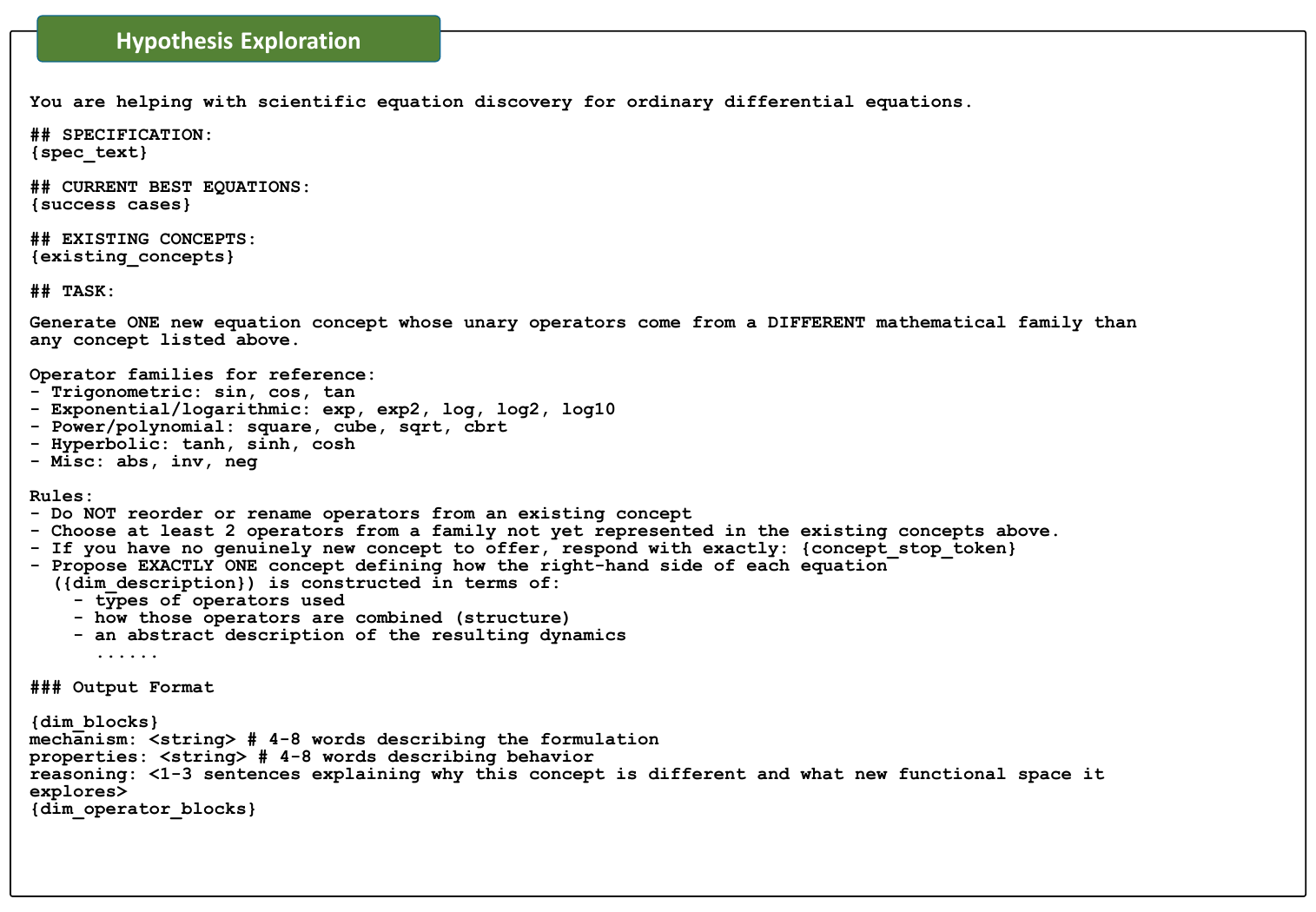}
    \vspace{-0.75em}
    \caption{\small \textbf{Prompt example of \AlgName for hypothesis exploration.} The prompt specifies the task specification, current best equations, existing operator set, and instructs the model to generate a new equation operator prior. This prompt enhances the diversity of operators. }
    \label{fig:ex_diversity}
    % \vspace{-1.5em}
\end{figure}

\vspace{-1em}
\paragraph{Data Acquisition via Active Experiment Selection.}
To address the fundamental limitation of static datasets, \modelname incorporates an active learning mechanism that expands the dataset in regions where current hypotheses disagree. A pool of candidate initial conditions is sampled from predefined bounds, and for each candidate, the learned equation systems are simulated forward using short-horizon numerical integration (e.g., forward Euler with fixed step size). At each round, the acquisition committee consists of the candidate systems fitted in that round. We sample a pool of 256 feasible initial conditions and evaluate the predictive-divergence score at three randomly selected conditions, followed by seven conditions selected using Gaussian-process UCB. The score is the median normalized pairwise discrepancy among finite candidate rollouts, as defined in Eq.~\ref{eq:active_query}. The highest-scoring evaluated initial condition is queried through the oracle. For each selected initial condition, trajectories are generated using the LSODA solver over 20 uniformly spaced time points in the interval $t \in [0,1]$. The selected initial conditions are then queried through a ground-truth ODE oracle, producing new trajectory data. Newly acquired samples are incorporated into the dataset via interleaved splitting, where even-indexed samples are assigned to the training set and odd-indexed samples to the validation set. Starting with the initial 20 data points, each of the ten acquisition steps contributes 10 training and 10 validation samples, ensuring balanced and incremental data augmentation.

\vspace{-1em}
\paragraph{Feedback-Driven Refinement.}
\modelname maintains a feedback-driven memory that aggregates information from prior iterations. Candidate equations are ranked per dimension based on validation performance. High-performing equations are abstracted into positive exemplars (top 2 candidates), while the worst-performing two candidates are recorded as failure cases. These summaries are injected into subsequent prompts, guiding the language model toward effective structural motifs and away from previously unproductive operator combinations. The exploration–exploitation balance is further controlled by varying prompts across iterations, alternating between refining promising operator families and exploring novel structures.
\vspace{-.75em}
\section{Additional Results}
% \vspace{-0.25em}

\label{app:add_results}

\vspace{-0.5em}
\subsection{Distributional Analysis Across Benchmark Systems}
\vspace{-0.5em}

Figures~\ref{fig:odebench_dist} and~\ref{fig:odebase_dist} illustrate the distribution of reconstruction, generalization, and extrapolation NMSE, together with expression complexity, across all systems in ODEBench and ODEBase. Across both benchmarks, \modelname exhibits error distributions that are consistently shifted toward lower values, demonstrating that its improvements are not driven by a few favorable systems but are observed broadly across the benchmarks. In particular, the first and third quartiles (Q1--Q3) of \modelname are substantially lower than those of competing approaches across reconstruction, generalization, and extrapolation settings, indicating that the majority of systems benefit from the proposed framework. Among the baselines, Bayesian optimization is the strongest competitor, although its interquartile ranges are generally wider and shifted toward higher errors, especially under distribution shift. Passive symbolic regression methods exhibit even larger quartiles centered around higher NMSE values, reflecting limited robustness beyond the training trajectories. From the perspective of expression complexity, \modelname maintains relatively compact equations with a comparatively narrow interquartile range, whereas \texttt{QBC} and \texttt{E2E} often produce considerably more complex expressions and exhibit much greater variability. The consistency of these trends across both ODEBench and ODEBase highlights the robustness of \modelname across datasets, system dimensions, and evaluation settings.

\begin{figure}[!htbp]
\vspace{-0.5em}
    \centering
    \includegraphics[width=\linewidth]{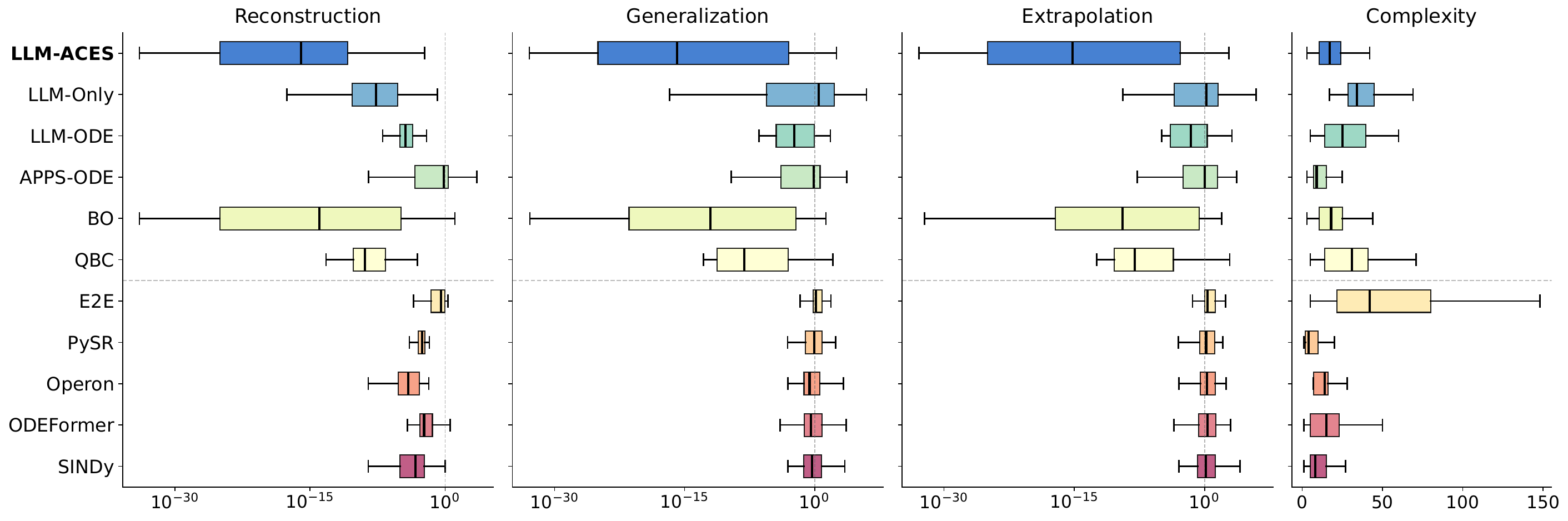}
    \vspace{-1.em}
    \caption{\small Distribution of reconstruction, generalization, extrapolation NMSE, and expression complexity across methods for ODEBench datasets.}
    \label{fig:odebench_dist}
\end{figure}

\begin{figure}[!htbp]
\vspace{-0.5em}
    \centering
    \includegraphics[width=\linewidth]{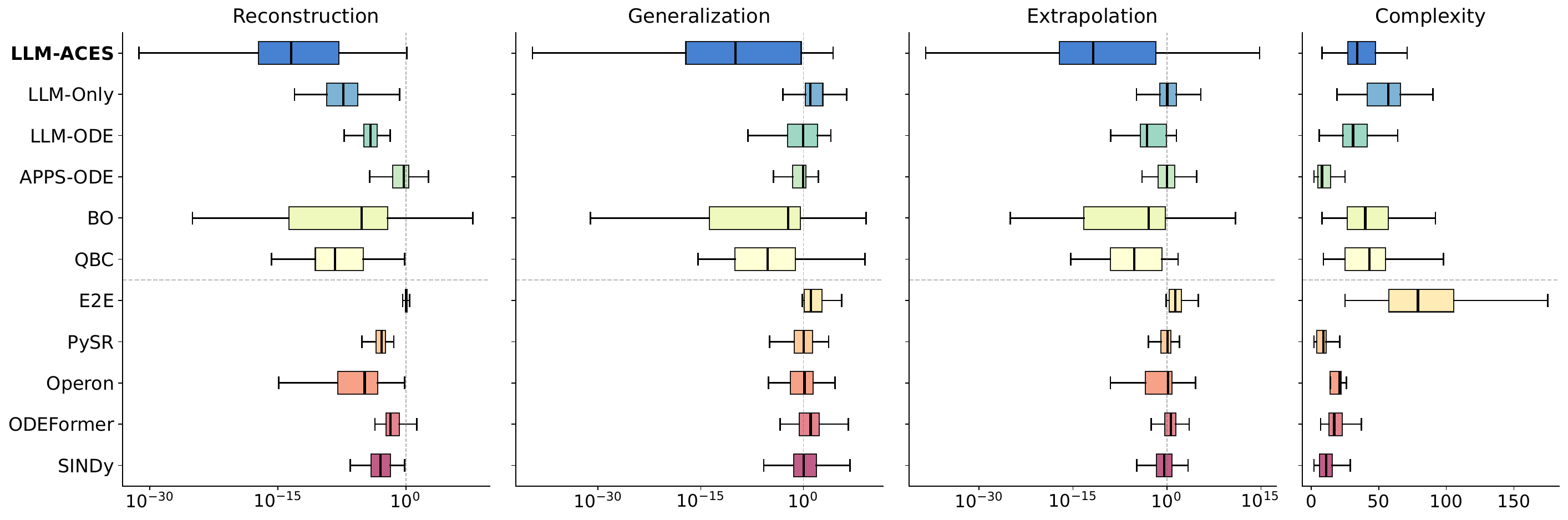}
    \vspace{-1em}
    \caption{\small Distribution of reconstruction, generalization, extrapolation NMSE, and expression complexity across methods for ODEBase datasets.}
    \vspace{-.5em}
    \label{fig:odebase_dist}
\end{figure}

\subsection{Role of LLMs in Search-Space Design}
\label{app:llm-contribution}

At each iteration, \modelname restricts the symbolic regression backend to an operator set proposed by the LLM. We evaluate whether the resulting gains stem from LLM-guided, context-dependent search-space design rather than from search-space restriction alone. Across three controlled variants, we hold the remainder of the pipeline fixed, including candidate fitting, initial-condition acquisition, re-scoring, and the overall query budget, and vary only the mechanism used to construct the operator set. The controls isolate the contribution of LLM guidance from the effects of simple restriction.

\paragraph{Effect of operator-space restriction.}
We constructed a privileged oracle-informed operator library from the union of ground-truth operator families appearing across the benchmark: \texttt{polynomial}, \texttt{sin}, \texttt{cos}, \texttt{exp}, and \texttt{log}. This setting provides substantially stronger prior information than would be available in a realistic discovery scenario and isolates the effect of restricting the symbolic search space. Table~\ref{tab:oracle-library} compares standard symbolic-regression baselines under their default operator sets and the oracle-informed library. The effect of restriction is limited and inconsistent across methods: PySR improves by approximately $18\%$, Operon remains nearly unchanged, and SINDy degrades. Despite access to the ground-truth operator families, all three baselines remain more than ten orders of magnitude above \modelname in reconstruction NMSE. These results indicate that static restriction of the operator space is insufficient to account for the performance of \modelname.

\begin{table}[h]
\centering
\vspace{-0.5em}
\caption{\small Median reconstruction NMSE (lower is better) under the default operator
set versus an oracle-informed library containing the ground-truth operator
families.}
\vspace{0.25em}
\label{tab:oracle-library}
\begin{tabular}{lcc}
\toprule
Method & Full set & Oracle-informed set \\
\midrule
SINDy   & $5.07\times10^{-4}$  & $6.23\times10^{-4}$ \\
PySR    & $2.84\times10^{-3}$  & $2.34\times10^{-3}$ \\
Operon  & $7.95\times10^{-5}$  & $7.73\times10^{-5}$ \\ \midrule
\textbf{\modelname (ours)} & \multicolumn{2}{c}{$\mathbf{1.33\times10^{-17}}$} \\
\bottomrule
\bottomrule
\vspace{-1.5em}
\end{tabular}
\end{table}

\paragraph{Oracle-informed selection within \modelname.}
We further isolated operator selection within the \modelname pipeline by replacing the LLM with the same oracle-informed operator library while leaving all remaining components unchanged, evaluated on a fixed subset of $15$ systems, comprising five systems each from the $1$D, $2$D, and $3$D benchmarks (same as Section~\ref{app:efficiency}). The oracle-informed variant is consistently outperformed by \modelname across reconstruction, generalization, and extrapolation prediction (Table~\ref{tab:oracle-llmaces}). Access to the correct operator families in aggregate therefore does not reproduce the benefit of LLM guidance. The relevant distinction is the selection of an operator subspace conditioned on the current system and search state, rather than the availability of a globally correct operator library.

\begin{table}[h]
\centering
\vspace{-.750em}
\caption{\small \modelname versus an oracle-informed-library variant of the same pipeline, on the 15-system subset. We report the Median NMSE values.}
\vspace{0.25em}

\label{tab:oracle-llmaces}
\begin{tabular}{lccc}
\toprule
Variant & Recon.\ NMSE & Gen.\ NMSE & Ext. NMSE \\
\midrule
Oracle-informed \modelname & $2.53\times10^{-18}$ & $2.12\times10^{-17}$ & $1.02\times10^{-17}$ \\
% NOTE: this row reads 4.01e-19 here but 4.01e-19 in Tables 4-6. Reconcile.
\textbf{\modelname (ours)} & $\mathbf{4.01\times10^{-19}}$ & $\mathbf{7.39\times10^{-18}}$ & $\mathbf{2.29\times10^{-18}}$ \\
\bottomrule
\bottomrule
\end{tabular}
\end{table}

\paragraph{Adaptive non-LLM operator selection.}
We designed an Operator Bandit (OB) using independent UCB1 bandits over the unary and binary operators available to \modelname. At each iteration, the operator set is formed from operators with the highest UCB scores, balancing empirical reward and exploration. PySR performs symbolic search within the selected operator set, and the resulting validation score is propagated to the selected operators through semi-bandit credit assignment. The reward is defined as $1/(1+\mathrm{NMSE}_{\mathrm{val}})$, with zero reward assigned to failed searches. The downstream search procedure and computational budget remain matched to \modelname. Table~\ref{tab:operator-bandit} shows that \modelname outperforms both baselines across all three metrics on the subset of 15 datasets. OB also underperforms the fixed oracle-informed variant, indicating that generic adaptation based on scalar validation feedback does not recover the gains of LLM-guided selection. The distinction is therefore not adaptivity alone, but adaptation conditioned on the current candidate equations and accumulated search context.

\begin{table}[h]
\centering
\vspace{-.5em}
\caption{\small LLM-guided operator selection versus a non-LLM adaptive selector and the fixed oracle-informed library, on the 15-system subset under a matched downstream pipeline and budget.  We report the Median NMSE values.}
\vspace{.25em}
\label{tab:operator-bandit}
\begin{tabular}{lccc}
\toprule
Variant & Recon.\ NMSE & Gen.\ NMSE & Ext. NMSE \\
\midrule
Oracle-informed (fixed, non-adaptive) & $2.53\times10^{-18}$ & $2.12\times10^{-17}$ & $1.02\times10^{-17}$ \\
Operator Bandit (adaptive, non-LLM)   & $7.67\times10^{-17}$ & $2.70\times10^{-17}$ & $6.40\times10^{-17}$ \\
\textbf{\modelname (ours)}              & $\mathbf{4.01\times10^{-19}}$ & $\mathbf{7.39\times10^{-18}}$ & $\mathbf{2.29\times10^{-18}}$ \\
\bottomrule
\bottomrule
\end{tabular}
\end{table}

\subsection{Hyperparameter Sensitivity}
\label{app:hyperparam-sensitivity}

\paragraph{Number of acquisition rounds ($T$).}
The acquisition budget determines the number of trajectories that can be queried from the oracle while maintaining the overall number of samples (see Table~\ref{tab:query-budget}). Since the intended setting assumes costly experiments, performance at small $T$ is particularly relevant. We evaluate $T \in \{1,2,5,10\}$ (each setting is trained on the same number of data points) on the fixed $15$-system subset, with $T{=}10$ used as the default. Performance improves substantially as the acquisition budget increases, with reconstruction NMSE decreasing monotonically across the sweep. Importantly, strong performance is already obtained with a single acquisition round. At $T{=}1$, \modelname achieves errors substantially below the full-budget baselines reported in the main evaluation, demonstrating that its gains persist in the low-query regime. Additional acquisitions further refine the candidate hypotheses, yielding the strongest reconstruction, generalization, and extrapolation performance at the default budget of $T{=}10$.

\begin{table}[h]
\centering
\vspace{-0.5em}
\caption{\small Performance of \modelname as a function of the acquisition budget $T$.  We report the Median NMSE values.}
\vspace{0.25em}

\label{tab:query-budget}
\begin{tabular}{lccc}
\toprule
$T$ & Recon.\ NMSE & Gen.\ NMSE & Ext. NMSE \\
\midrule
1            & $1.79\times10^{-16}$ & $8.07\times10^{-15}$ & $8.18\times10^{-14}$ \\
2            & $3.40\times10^{-17}$ & $1.41\times10^{-15}$ & $6.84\times10^{-16}$ \\
5            & $2.35\times10^{-17}$ & $1.69\times10^{-16}$ & $6.89\times10^{-16}$ \\
10 & $4.01\times10^{-19}$ & $7.39\times10^{-18}$ & $2.29\times10^{-18}$ \\
\bottomrule
\bottomrule
\vspace{-1.25em}

\end{tabular}
\end{table}

\paragraph{Number of priors per round ($K$).}
The parameter $K$ determines the number of operator hypotheses generated and fitted at each round. We evaluate $K \in \{1,3,10\}$ with $T{=}10$ fixed. Increasing $K$ consistently improves performance across all three metrics. The largest qualitative change occurs between $K{=}1$ and $K{=}3$: with a single candidate prior, the population cannot provide a meaningful cross-hypothesis disagreement signal for active acquisition, whereas multiple priors support both broader symbolic search and hypothesis-driven experiment selection. Increasing $K$ to $10$ yields effectively exact recovery on this subset; however, the computational cost increases significantly as compared to the $K=3$ setting.

\begin{table}[h]
\centering
\vspace{-0.5em}
\caption{\small Effect of the number of operator priors per round, $K$.  We report the Median NMSE values.}
\vspace{0.25em}

\label{tab:k-ablation}
\begin{tabular}{lccc}
\toprule
$K$ & Recon.\ NMSE & Gen.\ NMSE & Ext. NMSE \\
\midrule
1           & $7.40\times10^{-16}$ & $7.88\times10^{-13}$ & $1.32\times10^{-10}$ \\
3  & $4.01\times10^{-19}$ & $7.39\times10^{-18}$ & $2.29\times10^{-18}$ \\
10          & $7.12\times10^{-33}$ & $8.03\times10^{-33}$ & $5.54\times10^{-33}$ \\
\bottomrule
\bottomrule
\vspace{-1.25em}
\end{tabular}
\end{table}

\paragraph{Robustness to prompt variation.}
To study the robustness to prompt variations, we removed each of four instructions independently: \texttt{no\_required\_operators} removes the requirement that \{$+$, $*$\} appear in the binary operator set; \texttt{no\_operator\_count\_bound} removes the constraint of 2--6 unary operators; \texttt{no\_family\_diversity} removes the operator-family reference list and the cross-round diversity instruction; and \texttt{no\_best\_equation\_study} removes conditioning on the operator families present in the current best equations. Performance remains strong under all four prompt perturbations, with every variant retaining errors in the low-NMSE regime. As Table~\ref{tab:prompt-ablation} shows, LLM-ACES does not depend critically on any single prompt instruction. All variants stay below $10^{-17}$, indicating a strong performance over passive baselines.

\begin{table}[h]
\centering
\caption{\small Sensitivity to prompt variations. We report the Median NMSE values.}
\label{tab:prompt-ablation}
\fontsize{8}{8}\selectfont
\begin{tabular}{lccc}
\toprule
Variant & Recon.\ NMSE & Gen.\ NMSE & Ext. NMSE \\
\midrule
\texttt{no\_required\_operators}    & $3.62\times10^{-19}$ & $1.44\times10^{-20}$ & $3.38\times10^{-19}$ \\
\texttt{no\_operator\_count\_bound} & $4.45\times10^{-19}$ & $1.64\times10^{-19}$ & $7.84\times10^{-20}$ \\
\texttt{no\_family\_diversity}      & $1.53\times10^{-19}$ & $5.61\times10^{-19}$ & $8.75\times10^{-19}$ \\
\texttt{no\_best\_equation\_study}  & $3.67\times10^{-21}$ & $3.56\times10^{-21}$ & $4.74\times10^{-21}$ \\ \midrule
Full prompt               & $4.01\times10^{-19}$ & $7.39\times10^{-18}$ & $2.29\times10^{-18}$ \\
\bottomrule
\bottomrule
\end{tabular}
\end{table}

\subsection{Robustness to Noise and Irregular Sampling}
\vspace{-0.25em}

To assess robustness under realistic observation noise, we follow the MDBench~\citep{bideh2026mdbench} evaluation protocol by corrupting clean trajectories at multiple SNR levels and evaluating the discovered right-hand side functions using reconstruction NMSE, while also reporting expression complexity and symbolic accuracy. This setup tests whether a method preserves the underlying dynamics rather than merely fitting noisy measurements, since MDBench explicitly evaluates predictive fidelity under varying noise conditions and penalizes overly complex symbolic forms. In addition, we follow the ODEBench~\citep{dascoli2024odeformer} sampling protocol by varying the observation subsampling ratio $\rho$, where a fraction of trajectory points is dropped uniformly at random, thereby measuring performance under both dense and sparse temporal observations. As shown in Figure~\ref{fig:nmse_noise}, across SNR levels and sampling regimes, \modelname achieves consistently stronger reconstruction accuracy than \texttt{SINDy, PySR, Operon}, and \texttt{ODEFormer}, indicating that its generated hypotheses remain stable even when the observed trajectories are noisy or partially sampled. Importantly, this improvement is obtained without a corresponding explosion in expression complexity, suggesting that \modelname discovers more robust equation structures rather than relying on unnecessarily long symbolic expressions. Overall, these results show that \modelname offers a favorable robustness–parsimony trade-off compared to existing baselines, with particularly clear gains in degraded observation settings where conventional symbolic regression and sparse-regression methods become less reliable.

\begin{figure}[!htbp]
    \centering
    \vspace{-4pt}
    \includegraphics[width=0.95\linewidth]{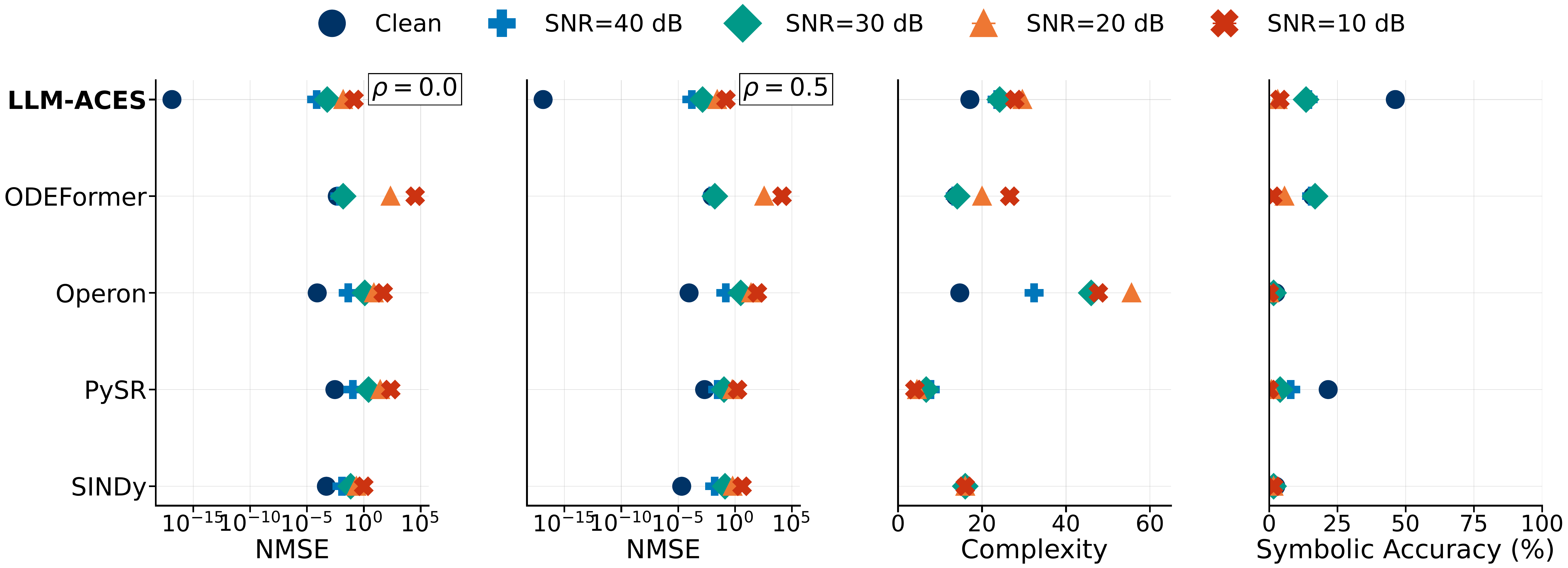}
    \vspace{-4pt}
    \caption{\small Evaluation under noisy observations. The complexity and symbolic accuracy results are for the $\rho=0.0$ setting.}
    \label{fig:nmse_noise}
    \vspace{-10pt}
\end{figure}

\subsection{Full-system Recovery}
\vspace{-.25em}
While Symbolic Accuracy (SA) measures recovery at the level of individual component equations, it can assign partial credit to a multi-dimensional system whose dynamics are only partially recovered. We therefore additionally evaluate {binary full-system recovery}. As shown in Table~\ref{tab:full_system_recovery}, LLM-ACES recovers 38/63 complete systems, compared with 28/63 for the strongest baseline, \texttt{BO}. It leads on 1D (11 vs. 8) and 2D systems (20 vs. 14) and also exceeds  \texttt{BO} on 3D/4D systems (7 vs. 6). Comparing BO with PySR, which share the same regression backend, shows how much actively acquired data contributes; LLM-ACES adds LLM-induced hypothesis spaces and disagreement-driven acquisition on top, giving the highest recovery. Because this metric gives no credit to partial recovery, these results show that LLM-ACES’s lower NMSE is accompanied by more complete recovery of the governing equations, rather than only more accurate trajectory fitting.
\vspace{-1.em}
\begin{table}[htbp]
\centering
\small
\caption{\small \textbf{Binary full-system recovery across ODEBench (63 systems)}. A system counts as recovered only when every component equation matches the ground-truth structure under the symbolic-accuracy protocol.}
\vspace{0.25em}
\fontsize{8}{8}\selectfont
\begin{tabular}{lcccc}
\toprule
Method & 1D & 2D & 3D/4D & Total \\
\midrule
PySR & 1 & 1 & 0 & 2 \\
ODEFormer & 2 & 3 & 0 & 5 \\
Operon & 2 & 0 & 0 & 2 \\
SINDy & 3 & 3 & 0 & 6 \\
LLM-ODE & 4 & 2 & 0 & 6 \\
BO & 8 & 14 & 6 & 28 \\
\midrule
\textbf{LLM-ACES} \texttt{(GPT)} & \textbf{11} & \textbf{20} & \textbf{7} & \textbf{38} \\
\bottomrule
\bottomrule
\label{tab:full_system_recovery}
\end{tabular}
\end{table}

\subsection{Robustness to the internal validation protocol.}
During closed-loop discovery, LLM-ACES uses a validation split only to score and select among candidate equations; the reconstruction, generalization, and extrapolation test sets are fixed and never enter the loop. By default, each queried trajectory is split by interleaving, with alternate time points assigned to training and validation. To test whether this choice affects which equations are selected, we rerun LLM-ACES on all 63 ODEBench systems under two alternative protocols: (i)~\emph{held-out IC validation}, where candidates are validated on oracle trajectories from two fixed initial conditions that are never used for training, and (ii)~\emph{temporal validation}, where the first half of each trajectory is used for training and the second half for validation, so that candidates are selected by how well they extrapolate forward in time. Table~\ref{tab:validation_protocols} reports median NMSE, the number of fully recovered systems, and symbolic accuracy (SA) computed as the fraction of correctly recovered component equations within each system, averaged across systems. The results are stable across protocols: median NMSE remains below $10^{-15}$ on all three test sets, full-system recovery ranges from 37/63 to 43/63, and SA ranges from 66.1\% to 73.7\%. Held-out IC validation, which is independent of the training trajectories, yields the highest recovery and SA, indicating that the interleaved default does not inflate symbolic recovery. Hence, the equations selected by LLM-ACES perform consistently better than the baselines shown in Table~\ref{tab:ode_results} and the resulting NMSE and symbolic-recovery conclusions are not too sensitive to the validation protocol.

\begin{table}[htbp]
  \centering
  \caption{\small LLM-ACES on ODEBench (63 systems) under three internal validation protocols. NMSE is the median over systems on the fixed reconstruction (Recon.), generalization (Gen.), and extrapolation (Ext.) test sets. Recovered counts systems whose every equation is recovered; SA is symbolic accuracy.}
  \vspace{0.25em}
  \label{tab:validation_protocols}
  \small
  \begin{tabular}{lccccc}
    \toprule
    & \multicolumn{3}{c}{Median NMSE $\downarrow$} & & \\
    \cmidrule(lr){2-4}
    Validation protocol & Recon. & Gen. & Ext. & Recovered $\uparrow$ & SA (\%) $\uparrow$ \\
    \midrule
    Interleaved (default)   & $1.33\times10^{-17}$ & $8.28\times10^{-17}$ & $2.46\times10^{-16}$ & 38/63 & 70.8 \\
    Held-out IC             & $1.35\times10^{-16}$ & $7.91\times10^{-17}$ & $2.74\times10^{-16}$ & 43/63 & 73.7 \\
    Temporal (split halves) & $1.52\times10^{-16}$ & $1.50\times10^{-16}$ & $6.27\times10^{-16}$ & 37/63 & 66.1 \\
    \bottomrule
    \bottomrule
  \end{tabular}
\end{table}

% \vspace{-0.5em}
\section{Additional Analyses}
% \vspace{-0.5em}

\label{app:analysis}

% \vspace{-.5em}
\subsection{Data Acquisition for Dynamical Equation Discovery}
% \vspace{-0.5em}

\label{app:identifiability}
Most existing equation discovery methods operate under a static-data paradigm, implicitly assuming that the available observations are sufficient for identifying the underlying dynamics. Yet low fitting error alone provides no guarantee of correct structural recovery, as different equations may explain the same observations equally well. We therefore examine the role of data acquisition in equation discovery and demonstrate that actively selecting experiments leads to both improved predictive performance and more faithful recovery of the true governing equations.

\vspace{-0.5em}
\begin{figure}[!htbp]
    \centering
    \includegraphics[width=0.725\linewidth]{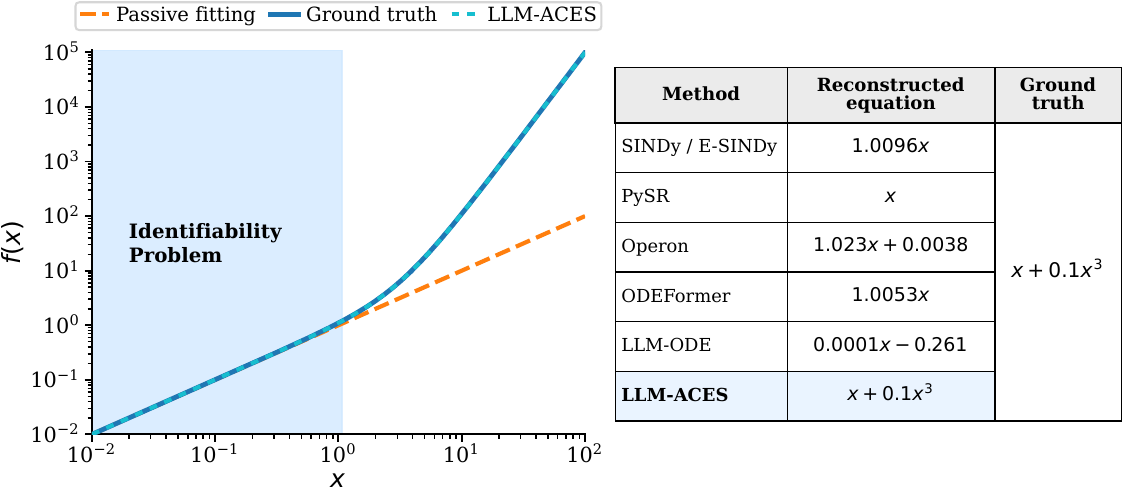}
    \vspace{-0.5em}
\caption{\small \textbf{Identifiability gap in passive equation fitting.} {(\textit{Left})} Comparison of the ground-truth dynamics, \modelname, and passive fitting baselines. The shaded region denotes the low-$x$ regime where passive fitting remains visually close to the ground truth. \textit{(Right)} Reconstructed equations returned by each method.}
\vspace{-1em}
\label{fig:identifiability_gap}    
\end{figure}

\paragraph{Identifiability Gap Analysis.}
When trajectories explore only a limited region of the state space, structurally distinct equations may produce nearly indistinguishable observations. Figure~\ref{fig:identifiability_gap} illustrates this phenomenon using a system with dynamics $\dot{x}=x+0.1x^3$. Training data collected from a low-$x$ regime render the nonlinear contribution effectively invisible, causing a variety of passive methods to recover approximately linear equations despite most of it achieving negligible fitting error. Although these models agree within the observed region, their predictions diverge dramatically outside it. This demonstrates an \emph{identifiability gap}: the inability to distinguish competing hypotheses due to insufficiently informative data. Importantly, the limitation arises from the observations themselves rather than from deficiencies in the symbolic regression algorithms. Further optimization on the same observations may not reliably distinguish competing structures when their predictions differ only below the available observation precision.
\vspace{-0.5em}

\paragraph{Effect of trajectory diversity.}
\begin{wrapfigure}{r}{0.5\linewidth}
    \vspace{-1.25em}
    \centering
    \includegraphics[width=\linewidth]{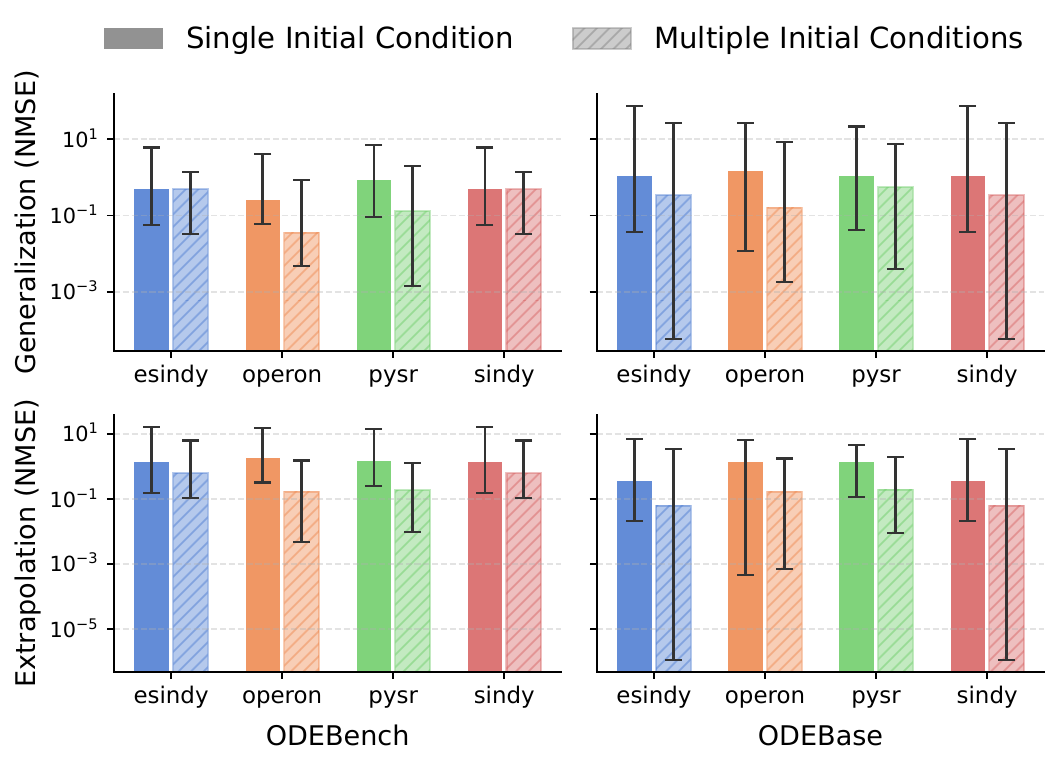}
    \vspace{-1.75em}
    \caption{\small \textbf{Comparison of equation recovery from a single trajectory and multiple trajectories.} Trajectory diversity improves performance across methods.}
    \label{fig:multi_ic}
    \vspace{-1.em}
\end{wrapfigure}
Figure~\ref{fig:multi_ic} examines the effect of increasing the number of initial conditions available for equation discovery while keeping the total number of observations at $100$ samples. We used data from $10$ different randomly sampled initial conditions to test the multiple trajectories setting. The evaluation is done on the second initial condition trajectory, which is not a part of the sampled initial conditions (for generalization), whereas extrapolation is on the extended time range for the first initial condition. Replacing a single trajectory with trajectories generated from multiple randomly sampled initial conditions consistently improves both generalization and extrapolation performance across symbolic regression methods. These results suggest that the diversity of trajectories provides more useful information than densely sampling a single trajectory. Nevertheless, the trajectories are obtained passively and do not explicitly target regions that are informative for distinguishing among candidate equations. As a result, improvements are limited by the coverage achieved through random sampling.
\vspace{-.5em}

\paragraph{Effect of acquisition strategy.}

Figure~\ref{fig:multi_trajectory} shows that the method used to select trajectories has a much larger impact than simply increasing their number. Single-trajectory datasets provide only local information, whereas data from multiple initial conditions improve coverage without explicitly targeting unresolved uncertainties. Approaches such as Bayesian optimization and query-by-committee (QBC) aim to guide experimentation using uncertainty estimates. Rather than seeking merely uncertain regions, \emph{predictive divergence} actively searches for experiments that discriminate between alternative explanations. Across ODEBench datasets, this strategy consistently achieves the lowest generalization error, compared with both passive baselines and existing active acquisition schemes. These results suggest that the central challenge in equation discovery is not obtaining more data, but obtaining the \emph{right} data. Symbolic hypotheses should therefore guide the collection of new trajectories, transforming equation discovery from a passive fitting problem into an iterative process of hypothesis generation and experimental design.
\vspace{-.25em}
\begin{figure}[!htbp]
    \centering
    \includegraphics[width=0.65\linewidth]{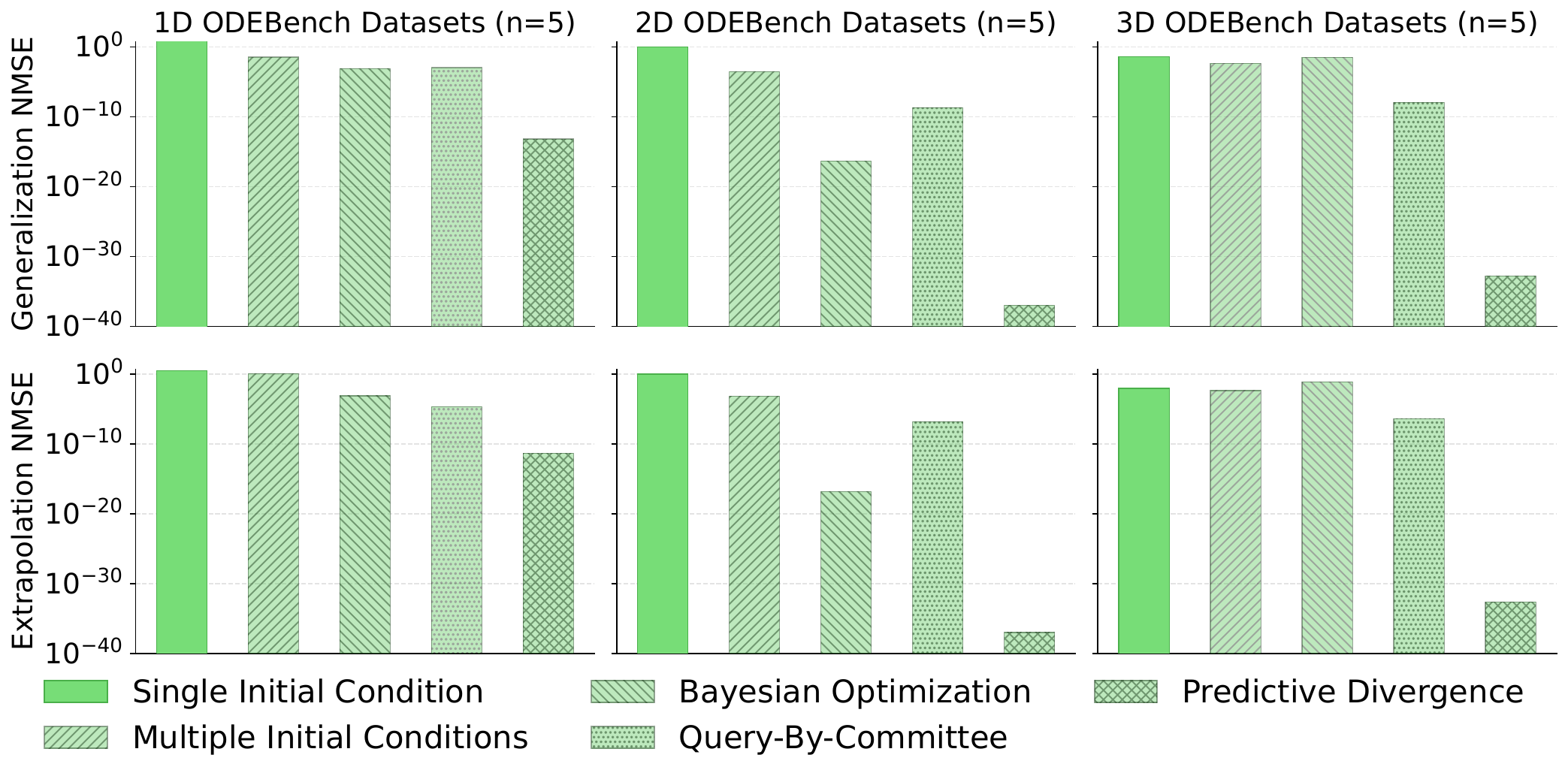}
    \vspace{-.25em}
    \caption{\small \textbf{Comparison of different trajectory acquisition strategies for PySR on 1D, 2D, and 3D ODEBench datasets.} Selecting trajectories based on predictive divergence consistently achieves the lowest NMSE.}
    \label{fig:multi_trajectory}
\end{figure}

\vspace{-1em}
\subsection{Wall-Clock and Compute Analysis}
\label{app:compute}
LLM-ACES incurs additional computation relative to passive symbolic regression
because each acquisition round searches multiple LLM-induced hypothesis spaces and evaluates their disagreement before acquiring new evidence. We therefore evaluate whether its performance gains can be explained simply by allocating more computation to symbolic search. We consider the compute-efficient $T{=}1$ configuration of LLM-ACES, which requires $250.5$ seconds using 60 PySR iterations and only three LLM calls (per dimension), and increase its symbolic-regression backend, PySR, beyond this wall-clock budget. Table~\ref{tab:compute_analysis} shows that additional passive search compute does not close the performance gap. Increasing PySR from its standard search budget to 500 and 750 iterations raises its wall-clock time to $327.3$ and $624.9$ seconds (per dimension), respectively, both exceeding the $250.5$ seconds required by LLM-ACES ($T{=}1$). Nevertheless, more PySR iterations do not guarantee an improvement in performance, as its reconstruction NMSE remains on the order of $10^{-3}$, compared with $1.79{\times}10^{-16}$ for LLM-ACES. Even the 750-iteration PySR run therefore uses approximately $2.5\times$ more wall-clock time while remaining roughly $13$ orders of magnitude worse in NMSE. LLM-ACES also requires substantially less wall-clock time than LLM-ODE that runs for $1000$ calls to generate $1000$ candidates despite achieving considerably lower error.

\begin{table}[htbp]
\centering
\small
\caption{\small Compute-controlled wall-clock comparison.}
\begin{tabular}{lccc}
\toprule
Method & NMSE & Time (s) & Configuration \\
\midrule
PySR & $2.85{\times}10^{-3}$ & 327.3 & 500 iter. \\
PySR & $2.12{\times}10^{-3}$ & 624.9 & 750 iter.\\
LLM-ODE & $3.82{\times}10^{-5}$ & 1683.0 & $\sim$1K LLM calls \\
\midrule
\textbf{LLM-ACES ($T{=}1$)} & $\mathbf{1.79{\times}10^{-16}}$ & \textbf{250.5} & 60 iter. + 3 LLM calls \\
\bottomrule
\bottomrule
\label{tab:compute_analysis}

\end{tabular}
\end{table}

% \vspace{-1.25em}
% \vspace{-1em}
\subsection{Qualitative Analysis}
% \vspace{-0.5em}
We provide qualitative visualizations for a representative trajectory from each ODE in Figure~\ref{fig:qual_analysis_odebench} for ODEBench and Figure~\ref{fig:qual_analysis_odebase} for ODEBase. For each system, we overlay the trajectory generated by \modelname's predicted equation onto the corresponding ground-truth trajectory for comparison.

\label{app:qual_analysis}
\begin{figure}[htbp]
    \vspace{-1em}
    \centering
    \includegraphics[width=\linewidth]{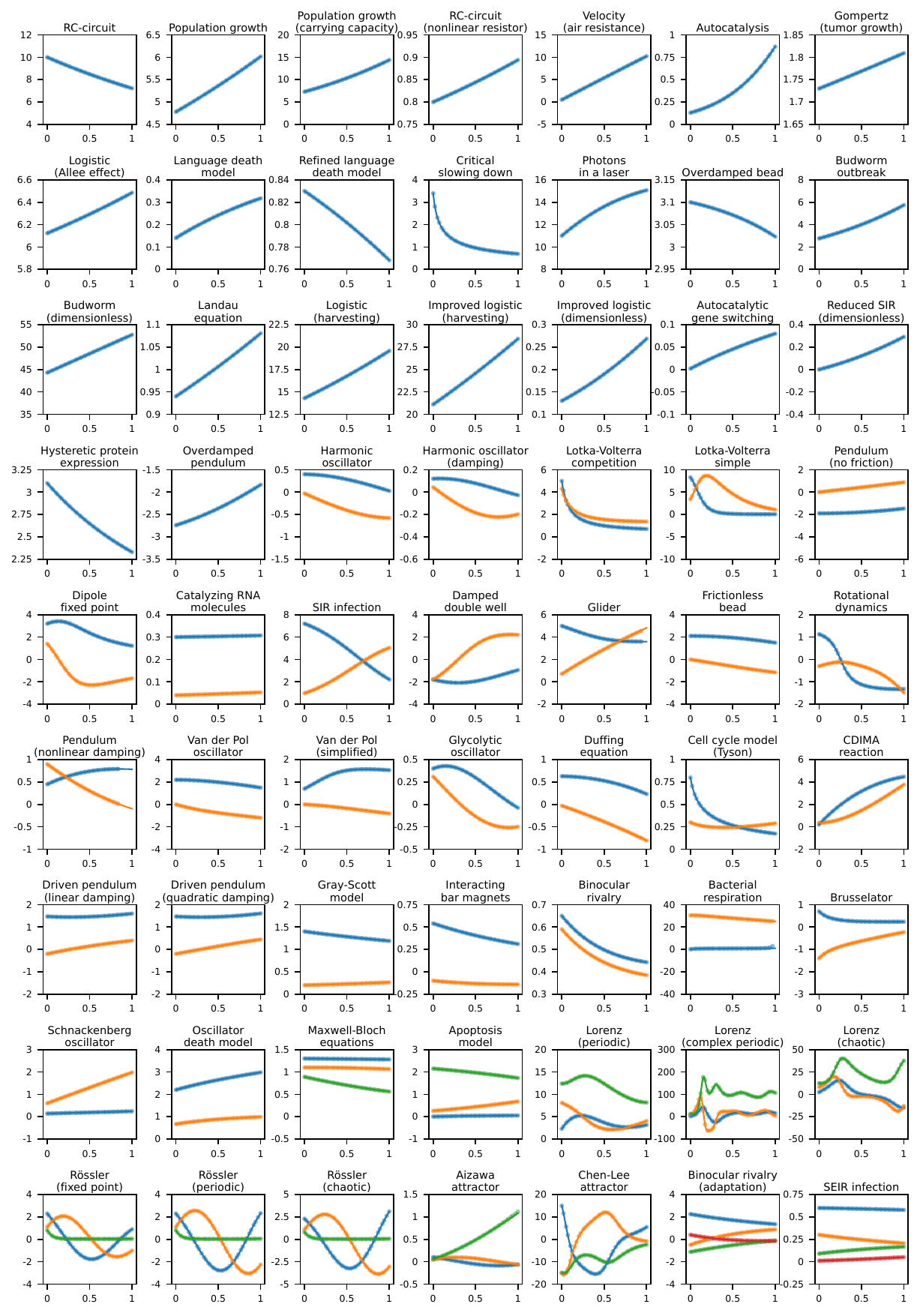}
    \caption{\small Predictions of \modelname (indicated in bold) and the ground-truth equations for all systems in ODEBench for the first set of initial conditions.}
    \label{fig:qual_analysis_odebench}
\end{figure}

\begin{figure}
    \centering
    \includegraphics[width=\linewidth]{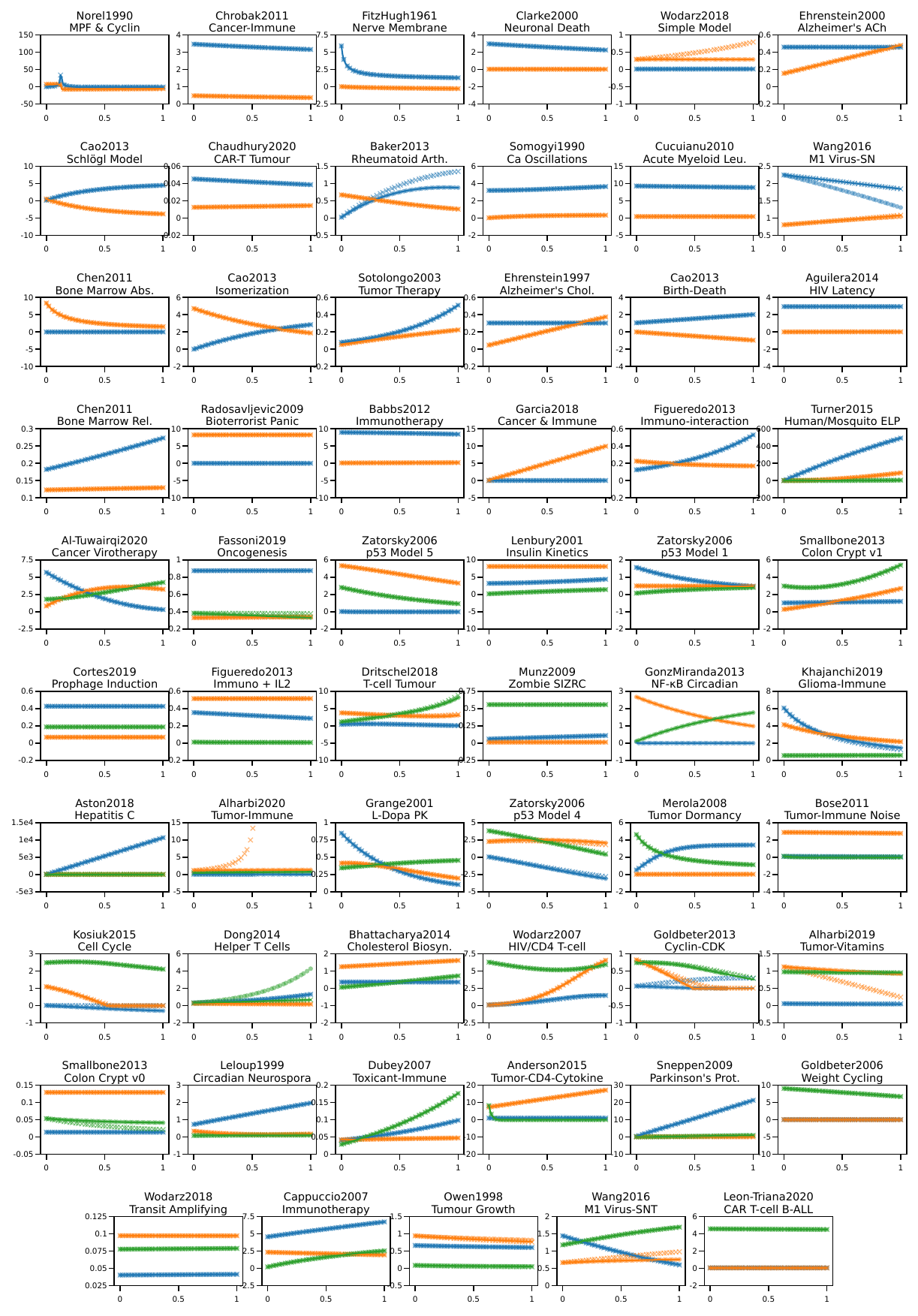}
    \caption{\small Predictions of \modelname (indicated in bold) and the ground-truth equations for all ODEBase systems for the first set of initial conditions.}
    \label{fig:qual_analysis_odebase}
\end{figure}

\newpage
\section*{NeurIPS Paper Checklist}

\begin{enumerate}

\item {\bf Claims}
    \item[] Question: Do the main claims made in the abstract and introduction accurately reflect the paper's contributions and scope?
    \item[] Answer: \answerYes{} % Replace by \answerYes{}, \answerNo{}, or \answerNA{}.
    \item[] Justification: The abstract and introduction outline the key problem addressed, the proposed method, contributions, and a summary of the experimental findings.

\item {\bf Limitations}
    \item[] Question: Does the paper discuss the limitations of the work performed by the authors?
    \item[] Answer: \answerYes{} % Replace by \answerYes{}, \answerNo{}, or \answerNA{}.
    \item[] Justification: We explicitly discuss the limitations of LLM-ACES following the Conclusion. 

\item {\bf Theory assumptions and proofs}
    \item[] Question: For each theoretical result, does the paper provide the full set of assumptions and a complete (and correct) proof?
    \item[] Answer:  \answerNA{} % Replace by \answerYes{}, \answerNo{}, or \answerNA{}.
    \item[] Justification: The paper does not introduce formal theoretical results, theorems, or convergence guarantees requiring proofs. Its contributions are algorithmic and are evaluated empirically through benchmark experiments, ablations, sensitivity analyses, and robustness studies. 

    \item {\bf Experimental result reproducibility}
    \item[] Question: Does the paper fully disclose all the information needed to reproduce the main experimental results of the paper to the extent that it affects the main claims and/or conclusions of the paper (regardless of whether the code and data are provided or not)?
    \item[] Answer: \answerYes{} % Replace by \answerYes{}, \answerNo{}, or \answerNA{}.
    \item[] Justification: The paper specifies the complete LLM-ACES procedure, acquisition objective, evaluation protocol, benchmark setup, baseline configurations, and principal hyperparameters in the main text and Appendix~\ref{app:implementation}. We additionally provide the implementation needed to reproduce the reported experiments through the code repository linked in the paper. 

\item {\bf Open access to data and code}
    \item[] Question: Does the paper provide open access to the data and code, with sufficient instructions to faithfully reproduce the main experimental results, as described in the supplemental material?
    \item[] Answer: \answerYes{} % Replace by \answerYes{}, \answerNo{}, or \answerNA{}.
    \item[] Justification: All experiments use publicly available ODEBench and ODEBase benchmarks, and the implementation of LLM-ACES is released through the repository linked in the paper.

\item {\bf Experimental setting/details}
    \item[] Question: Does the paper specify all the training and test details (e.g., data splits, hyperparameters, how they were chosen, type of optimizer) necessary to understand the results?
    \item[] Answer:\answerYes{} % Replace by \answerYes{}, \answerNo{}, or \answerNA{}.
    \item[] Justification: We provide detailed descriptions of the experimental setup and hyperparameter settings in Section~\ref{sec:experiment} and Appendix~\ref{app:implementation}  to support reproducibility.

\item {\bf Experiment statistical significance}
    \item[] Question: Does the paper report error bars suitably and correctly defined or other appropriate information about the statistical significance of the experiments?
    \item[] Answer: \answerNo{} % Replace by \answerYes{}, \answerNo{}, or \answerNA{}.
    \item[] Justification: Due to the substantial computational cost of symbolic and LLM-based discovery, we perform one run per dynamical system and therefore do not report seed-level confidence intervals or formal significance tests. Instead, we report medians and distributional statistics, including quartiles across benchmark systems, in Appendix~\ref{app:add_results} to characterize the consistency and variability of performance.

\item {\bf Experiments compute resources}
    \item[] Question: For each experiment, does the paper provide sufficient information on the computer resources (type of compute workers, memory, time of execution) needed to reproduce the experiments?
    \item[] Answer: \answerYes{} % Replace by \answerYes{}, \answerNo{}, or \answerNA{}.
    \item[] Justification: Appendix~\ref{app:implementation} reports the computational environment used for the experiments, including CPU and GPU hardware, available memory, per-dataset computational limits, and the hosted services used for LLM inference. Additionally, Appendix~\ref{app:compute} reports wall-clock time for a representative configuration.
    
\item {\bf Code of ethics}
    \item[] Question: Does the research conducted in the paper conform, in every respect, with the NeurIPS Code of Ethics \url{https://neurips.cc/public/EthicsGuidelines}?
    \item[] Answer: \answerYes{}  % Replace by \answerYes{}, \answerNo{}, or \answerNA{}.
    \item[] Justification: The research conforms to the NeurIPS Code of Ethics. The experiments use established scientific benchmarks and computational or simulated dynamical systems, and do not involve private personal data, human-subject experimentation, or other sensitive data collection.

\item {\bf Broader impacts}
    \item[] Question: Does the paper discuss both potential positive societal impacts and negative societal impacts of the work performed?
    \item[] Answer: \answerYes{} % Replace by \answerYes{}, \answerNo{}, or \answerNA{}.
    \item[] Justification: We provide an impact statement discussing the societal impacts of the work in detail.
    
\item {\bf Safeguards}
    \item[] Question: Does the paper describe safeguards that have been put in place for responsible release of data or models that have a high risk for misuse (e.g., pre-trained language models, image generators, or scraped datasets)?
    \item[] Answer: \answerNA{} % Replace by \answerYes{}, \answerNo{}, or \answerNA{}.
    \item[] Justification: The paper poses no such risks.

\item {\bf Licenses for existing assets}
    \item[] Question: Are the creators or original owners of assets (e.g., code, data, models), used in the paper, properly credited and are the license and terms of use explicitly mentioned and properly respected?
    \item[] Answer: \answerYes{} % Replace by \answerYes{}, \answerNo{}, or \answerNA{}.
    \item[] Justification: We have appropriately cited all relevant prior work, including their code repositories, and properly credited the creators of any code, data, and models used.

\item {\bf New assets}
    \item[] Question: Are new assets introduced in the paper well documented and is the documentation provided alongside the assets?
    \item[] Answer:  \answerYes{} \ % Replace by \answerYes{}, \answerNo{}, or \answerNA{}.
    \item[] Justification: We release the LLM-ACES code in the linked repository. 

\item {\bf Crowdsourcing and research with human subjects}
    \item[] Question: For crowdsourcing experiments and research with human subjects, does the paper include the full text of instructions given to participants and screenshots, if applicable, as well as details about compensation (if any)? 
    \item[] Answer: \answerNA{} % Replace by \answerYes{}, \answerNo{}, or \answerNA{}.
    \item[] Justification: This paper does not involve crowdsourcing or research with human subjects. 

\item {\bf Institutional review board (IRB) approvals or equivalent for research with human subjects}
    \item[] Question: Does the paper describe potential risks incurred by study participants, whether such risks were disclosed to the subjects, and whether Institutional Review Board (IRB) approvals (or an equivalent approval/review based on the requirements of your country or institution) were obtained?
    \item[] Answer: \answerNA{} % Replace by \answerYes{}, \answerNo{}, or \answerNA{}.
    \item[] Justification: This paper does not involve any research with human subjects.

\item {\bf Declaration of LLM usage}
    \item[] Question: Does the paper describe the usage of LLMs if it is an important, original, or non-standard component of the core methods in this research? Note that if the LLM is used only for writing, editing, or formatting purposes and does \emph{not} impact the core methodology, scientific rigor, or originality of the research, declaration is not required.
    %this research? 
    \item[] Answer: \answerYes{} % Replace by \answerYes{}, \answerNo{}, or \answerNA{}.
    \item[] Justification: The paper clearly describes the use of LLMs as a core component of the method, specifically for generating structured operator priors and guiding the hypothesis space in the closed-loop discovery process along with using it as a judge for calculating symbolic accuracy.

\end{enumerate}

\end{document}